\documentclass{article} 
\usepackage[preprint]{colm2026_conference}

\usepackage{microtype}
\usepackage{hyperref}
\usepackage{url}
\usepackage{booktabs}
\usepackage{subcaption}
\usepackage{adjustbox}
\usepackage{amsmath}
\usepackage{graphicx}
\usepackage[table]{xcolor}

\usepackage[most]{tcolorbox}
\usepackage{xcolor}
\definecolor{cardgreen}{RGB}{34,139,34}
\usepackage{amssymb}

\usepackage{tabularx}
\usepackage{array}
\usepackage{url}

\usepackage{color-edits}
\addauthor{jn}{blue}
\addauthor{gn}{magenta}
\addauthor{lm}{violet}

\usepackage{lineno}

\definecolor{darkblue}{rgb}{0, 0, 0.5}
\hypersetup{colorlinks=true, citecolor=darkblue, linkcolor=darkblue, urlcolor=darkblue}
\newcommand{\markamplified}{$^{\dagger}$}
\usepackage{caption}

\title{Amplified Does Not Mean Predictive:\\
Reasoning Behaviors in Thinking Models}

\author{%
\textbf{Jean de Dieu Nyandwi}$^{1}$,\;\;
\textbf{Leena Mathur}$^{1}$\\
\textbf{Yonatan Bisk}$^{1}$,\;\;
\textbf{Robert Hawkins}$^{2}$,\;\;
\textbf{Graham Neubig}$^{1}$ \\
$^{1}$Carnegie Mellon University \qquad
$^{2}$Stanford University \\
\texttt{\{jeandedi,gneubig\}@andrew.cmu.edu} \\
\href{https://neulab.github.io/behavioral-lift/}{\texttt{https://neulab.github.io/behavioral-lift/}} \\
\href{https://github.com/neulab/behavioral-lift}{\texttt{https://github.com/neulab/behavioral-lift}} \\
\href{https://huggingface.co/datasets/neulab/behavioral-lift}{\texttt{https://huggingface.co/datasets/neulab/behavioral-lift}}
\vspace{-3mm}
}

\begin{document}

\ifcolmsubmission
\linenumbers
\fi

\maketitle

\begin{abstract}
Which reasoning behaviors are associated with correct answers in reasoning models, and does reasoning-oriented training amplify those behaviors? This distinction is important because reasoning-oriented training can make traces look more deliberative without amplifying the behaviors most tied to model correctness. We quantify this mismatch with Behavioral Lift, a metric that measures how much correctness changes when a behavior is present versus absent in a model's reasoning trace. Across 15 models and 6 benchmarks spanning text-only and vision-language reasoning, we annotate 15,282 traces with a taxonomy whose core behaviors are defined for both LLM and VLM traces. We find evidence for an Amplification-Lift Gap, in which thinking models strongly amplify self-correction, hypothesis testing, and uncertainty acknowledgment, while the highest-lift behaviors are confidence calibration, knowledge alignment, and self-awareness. Confidence calibration is among the strongest positive signals of correctness in both modalities, yet is barely amplified; uncertainty acknowledgment is amplified by 3--7$\times$, yet is weakly or negatively associated with correctness. We find that reasoning-oriented training does not preferentially amplify the highest-Lift behaviors, motivating process-level objectives that reward calibrated and grounded reasoning rather than surface form alone.
\end{abstract}

\section{Introduction}
\label{sec:intro}
Recent progress in reasoning-oriented post-training has produced a wave of ``thinking'' models: language and vision-language models trained to generate extended reasoning traces before answering.
OpenAI's o1~\citep{openai2024o1}, DeepSeek-R1~\citep{guo2025deepseek}, Qwen3~\citep{qwen2025qwen3}, Kimi-k1.5~\citep{kimi2025k15}, and others now routinely produce long chains of reasoning on difficult problems and often outperform their instruction-tuned counterparts.
Longer traces, however, also make it easier to confuse deliberation with reliable reasoning.
Accuracy indicates whether a model gets the answer correct; it does not indicate which behaviors the model used, which failures it encountered, or whether the behaviors more common in thinking traces are the ones most associated with correctness.

In this paper, we study whether thinking models amplify the reasoning behaviors most associated with correct answers.
The distinction is important because a behavior can become more frequent without becoming more diagnostic of successful reasoning.
A model may exhibit more hedging behaviors  because it is confused, self-correct more because it made an earlier mistake, or branch over hypotheses without testing the right ones.
Conversely, less prevalent behaviors, such as calibrating confidence to the strength of the reasoning or applying the appropriate domain framework, may be more predictive of success.

Recent work has begun to analyze reasoning traces through cognitively-inspired taxonomies, trainability signals, and strategy discovery from chain-of-thought traces~\citep{gandhi2025cognitive, kargupta2025cognitive, lee2025cot}.
We go beyond describing which behaviors appear in reasoning traces by comparing thinking and instruct models and separating \emph{behavioral prevalence} from \emph{behavioral lift}. 
To separate prevalence from lift, we annotate reasoning traces with a cross-modal behavioral taxonomy spanning reasoning behaviors, failure modes, reasoning-quality labels, reasoning-type labels, summary metrics, and, for multimodal VLMs, visual grounding. We  formalize two metrics: (1)
\textbf{Behavioral Lift}, which measures how much correctness differs when a behavior is present versus absent in a model's trace; (2)
\textbf{Recovery Rate}, which measures how often a model reaches the correct answer, despite exhibiting at least one reasoning failure. We use our taxonomy to evaluate 15 models from 7 families across 6 benchmarks spanning a task spectrum from pure visual puzzles (VisualPuzzles)~\citep{song2025visualpuzzles} and logical reasoning (LogiQA2)~\citep{liu2020logiqa}, through mathematical reasoning (MathVista, MATH-500)~\citep{lu2023mathvista,hendrycksmath2021}, to knowledge-intensive QA (MMMU, MMLU-Pro)~\citep{yue2024mmmu, wang2024mmlu}.
Using LLM-as-judge annotation under our taxonomy, we produce 15,282 behavioral profiles.

\begin{figure}[t]
\centering
\includegraphics[width=0.94\columnwidth]{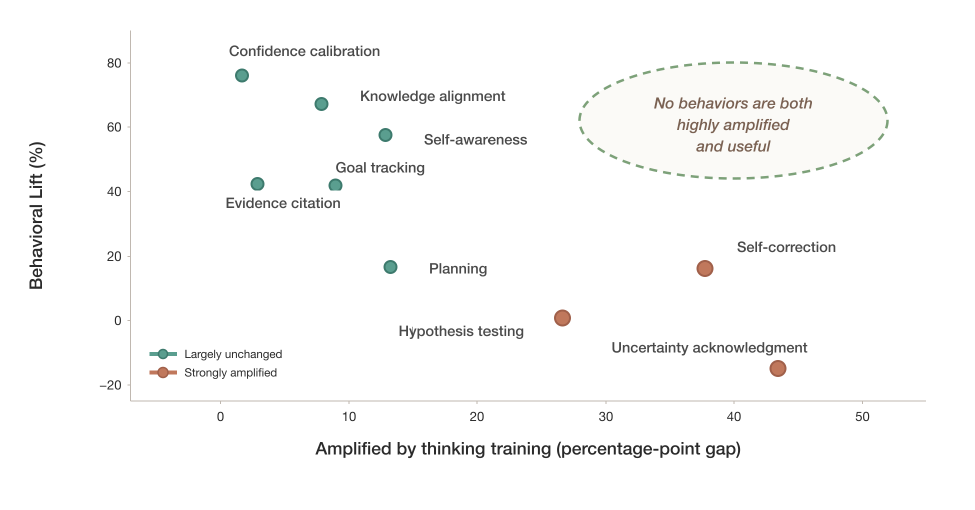}
\caption{The disconnect between what thinking-oriented training amplifies and what predicts success. Each point is one of the nine cross-modal higher-order behaviors, averaged across VLMs and LLMs ($N{=}15{,}282$). The top-right quadrant is empty: the behaviors most amplified by thinking training are not the ones most associated with correctness.}
\label{fig:irony}
\vspace{-10pt}
\end{figure}

Our analysis reveals an \textbf{Amplification-Lift Gap}, in which the behaviors most amplified by thinking training are not the behaviors most associated with correctness.
As shown in Figure~\ref{fig:irony}, \textit{no behavior is both strongly amplified and strongly predictive of success}.
Thinking models selectively amplify self-correction, hypothesis testing, and uncertainty acknowledgment, while leaving six other behaviors, including confidence calibration and knowledge alignment, largely unchanged.
Confidence calibration is one of the strongest positive signals of correctness in both modalities (+72--80\%), while uncertainty acknowledgment is strongly amplified but weakly or negatively associated with correctness.
Thinking models recover better on extended-reasoning tasks and make fewer failures on knowledge-heavy tasks, while instruct models can outperform on tasks where fast recognition of logical form is sufficient for success.
This paper makes four  contributions:
\begin{itemize}
    \item We introduce \textbf{Behavioral Lift}, a metric that separates how often a behavior appears from how strongly it is associated with reasoning  correctness.
    \item We introduce a cross-modal behavioral taxonomy with 9 behaviors and 15,282 annotated traces from 15 LLMs and VLMs across 6 benchmarks.
    \item We contribute empirical evidence for an Amplification-Lift Gap, finding that thinking-oriented models do not exhibit the highest-Lift behaviors.
    \item We establish \textit{recovery} as a mechanism behind thinking-model gains; we find that thinking helps when tasks reward extended computation or recovery from  failures.
\end{itemize}
We release all annotation prompts, metrics code, and behavioral annotations.
\section{Taxonomy and Metrics}
\label{sec:taxonomy}
We develop a behavioral taxonomy that characterizes reasoning traces along multiple dimensions, and we  define two metrics that capture aspects of reasoning quality. The taxonomy includes cross-modal reasoning behaviors, failure modes, reasoning-quality labels, reasoning-type labels, summary metrics, and, for VLMs, visual grounding. 

\subsection{Design Principles}
\label{sec:design_principles}
The following three principles guide the taxonomy: (1) we annotate the reasoning process, in addition to the final answer, distinguishing valid logic from lucky guesses; (2) we define nine higher-order behaviors across modalities using modality-neutral language; (3) we distinguish behavioral presence from Behavioral Lift, since behaviors that appear frequently are not always the ones most associated with success.

\subsection{Taxonomy Structure}
\label{sec:taxonomy_groups}
The taxonomy is organized into six groups, with full definitions in Appendix Tables \ref{tab:llm_taxonomy_1} and \ref{tab:vlm_taxonomy_1}.
Two groups drive the main analyses in this paper: the nine higher-order behaviors, which support cross-modal comparison, and the failure modes, which support the Recovery Rate analysis.
The remaining groups provide modality-specific grounding and descriptive context: reasoning quality distinguishes sound reasoning from lucky guesses, reasoning types characterize the form of reasoning used, summary metrics provide compact trace-level summaries, and visual grounding captures image use in VLM traces. We analyze these VLM-specific grounding labels separately in Appendix~\ref{app:vision_grounding}.

\noindent\textbf{Higher-order behaviors (9, defined for both modalities).} 
Following distinctions in the metacognition literature~\citep{nelson1990metamemory, brown1987metacognition}, we organize the nine behaviors (\autoref{tab:shared_behaviors}) into three functional categories:
\textit{control/regulation} (planning, goal tracking, hypothesis testing, self-correction),
\textit{monitoring/judgment} (uncertainty acknowledgment, confidence calibration, self-awareness), and
\textit{epistemic grounding} (evidence citation, knowledge alignment).\footnote{Our grouping is motivated by the standard metacognitive distinction between monitoring and control/regulation~\citep{nelson1990metamemory, schraw1994assessing}, while treating evidence citation and knowledge alignment as a separate epistemic-grounding family. We define these categories by their observable role in reasoning traces rather than claiming correspondence to specific internal cognitive mechanisms.}
These three categories are descriptive and situate cross-modal behaviors within the literature, and our empirical analyses group behaviors by how strongly thinking-oriented training amplifies them.
All nine behaviors are defined in modality-neutral language (e.g., self-correction refers to the same behavior, whether the model corrects a visual interpretation or a mathematical derivation).
These behaviors are the basis for the amplification and Behavioral Lift analyses.

\noindent\textbf{Failure Modes (7 per modality, 4 defined for both modalities).}
Each failure mode is annotated as present or absent (true means the failure occurred).
Four failure modes use identical definitions in both LLM and VLM traces: logical failure (invalid inferences), post-hoc rationalization (reasoning reverse-engineered from the answer), shortcut (skipping necessary steps), and lucky guess (correct answer with wrong reasoning).
The VLM variant adds failure modes in visual hallucination, visual neglect, and language bias.
The LLM variant adds failure modes in factual error, context misread, and knowledge gap.
Prior behavioral taxonomies~\citep{gandhi2025cognitive, kargupta2025cognitive} characterize what models do well, but do not emphasize failure cases.
Explicit failure annotation enables our Recovery Rate analysis to measure whether models reach correct answers despite exhibiting failures.

\begin{table}[t]
\vspace{-6pt}
\centering
\footnotesize
\setlength{\tabcolsep}{4pt}
\begin{tabular}{@{}p{0.23\textwidth}p{0.52\textwidth}@{}}
\toprule
Behavior & Short definition \\
\midrule
Planning & Breaks the problem into explicit sub-steps. \\
Goal tracking & Tracks progress toward intermediate or final goals. \\
Hypothesis testing & Considers alternatives, cases, or interpretations. \\
Self-correction & Revises or corrects a previous step or claim. \\
Uncertainty ack. & Expresses doubt, ambiguity, or confusion. \\
Confidence calibration & Certainty tracks the strength of reasoning. \\
Self-awareness & Recognizes missing information or limits of support. \\
Evidence citation & Grounds claims in prompt evidence or constraints. \\
Knowledge alignment & Uses the appropriate domain facts or framework. \\
\bottomrule
\end{tabular}
\caption{Nine behaviors used for cross-modal analysis. Full definitions appear in Appendix Tables~\ref{tab:llm_taxonomy_1} and~\ref{tab:vlm_taxonomy_1}.}
\label{tab:shared_behaviors}
\vspace{-8pt}
\end{table}

\subsection{Metrics}
\label{sec:metrics}
We define two metrics to study the questions motivating this paper.

\textbf{Behavioral Lift.}\ 
How much does correctness in model reasoning differ when a behavior is present?
For each behavior $b$, we compute:
\begin{equation}
    \text{Lift}(b) = P(\text{correct} \mid b{=}\texttt{true}) - P(\text{correct} \mid b{=}\texttt{false})
    \label{eq:lift}
\end{equation}
Positive Lift\footnote{We refer to the metric as \textit{Behavioral Lift}, often shortened to \textit{Lift}.} means the behavior is associated with higher accuracy when present; negative lift means it is associated with lower accuracy.
We note that Lift does not imply causation: a behavior may have high Lift as a consequence of the model being on the right track, rather than a cause of correctness. We therefore use Lift as a descriptive measure of reasoning behavior rather than as a causal estimate of a behavior's effect.
We compute Lift on pooled samples across all models per modality for the main analyses and report per-modality values to assess cross-modal consistency. Supplementary analyses compute the same quantity within individual benchmarks and individual models to assess stability.

\textbf{Recovery Rate.}\
Can a model reach correct answers despite exhibiting reasoning failures?
For a set of responses from model $m$:
\begin{equation}
    \text{Recovery}(m) = P(\text{correct} \mid \exists\, f \in \mathcal{F}_{\text{all}}: f{=}\texttt{true})
    \label{eq:recovery}
\end{equation}
where $\mathcal{F}_{\text{all}}$ includes all 7 failure modes for the relevant modality.
A high recovery rate indicates that the model reaches correct answers despite detected failures. 
\section{Experimental Setup}
\label{sec:setup}

We evaluate 15 models across 6 benchmarks, selecting models to enable matched-family comparisons between thinking and instruct variants and benchmarks to span a spectrum from  reasoning to knowledge-intensive tasks.
\subsection{Models}
\label{sec:models}
We select models where both thinking and instruction-tuned variants are publicly available from the same model family, enabling within-family comparison (\autoref{tab:models} in Appendix).
Our evaluation includes 15 open-weight models from 3B to 9B parameters: 4 thinking and 3 non-thinking VLMs, and 4 thinking and 4 non-thinking LLMs, with matched thinking/non-thinking pairs wherever available.
\subsection{Benchmarks}
\label{sec:benchmarks}
We select benchmarks to span three task types per modality. For VLMs: VisualPuzzles (visual reasoning), MathVista (visual mathematical reasoning), and MMMU (multimodal knowledge). For LLMs: LogiQA2 (logical reasoning), MATH-500 (mathematical reasoning), and MMLU-Pro (knowledge-intensive tasks across 14 domains). This design creates a task spectrum within each modality, from structure-based to knowledge-heavy reasoning:
\begin{center}
\small
\begin{tabular}{@{}lccc@{}}
\toprule
& Structure-focused & Mixed & Knowledge-heavy \\
\midrule
VLM & VisualPuzzles (350) & MathVista (300) & MMMU (350) \\
LLM & LogiQA2 (350) & MATH-500 (350) & MMLU-Pro (350) \\
\bottomrule
\end{tabular}
\end{center}

\noindent We target 350 responses per benchmark, except MathVista, where we use 300 from the testmini split. Final counts fall slightly below these targets because a small number of responses fail output parsing and are excluded. In total we annotate 15,282 responses.
\subsection{Evaluation Protocol}
\label{sec:eval_protocol}
\textbf{Inference.}\
All models are evaluated under standardized conditions with full traces retained for annotation; details appear in Appendix~\ref{app:inference}.

\textbf{Behavioral annotation.}\
We annotate each response using LLM-as-judge (GPT-4o).
The judge receives the question, ground-truth answer, and the model's full output, and produces binary annotations for all behaviors in the relevant taxonomy.
For confidence calibration, the judge marks \texttt{true} only when the model's expressed certainty is consistent with the observed strength of its reasoning in the trace, and \texttt{false} when the trace is clearly overconfident or underconfident relative to that reasoning.
The annotation prompt provides explicit definitions and criteria for each behavior (Appendix Pages~\pageref{app:llm_prompt} and~\pageref{app:vlm_prompt}).

\textbf{Judge validation.}\
To assess annotation reliability, we compare GPT-4o\footnote{
GPT-4o has also been shown to effectively monitor stronger reasoning models from CoT traces in a reward-hacking setting~\citep{baker2025monitoring}; our use is taxonomy annotation, which we validate with cross-judge agreement, manual checks, and robustness analyses.
} annotations with three independent judge models: DeepSeek-V3 (DeepSeek), Gemini-2.5-Flash (Google), and Gemini-3-Flash (Google) on 600 stratified MATH-500 samples across all 8 LLMs.
For the behaviors central to the amplification and Behavioral Lift analyses, agreement is moderate to substantial: self-correction ($\kappa = 0.53$--$0.82$), uncertainty acknowledgment ($\kappa = 0.71$--$0.79$), hypothesis testing ($\kappa = 0.51$--$0.59$), and confidence calibration ($\kappa = 0.56$--$0.81$).
Mean agreement across all behaviors ranges from 83.7\% to 88.8\% ($\kappa = 0.46$--$0.56$) across the three judge pairs.
As an additional manual check, we verified 120 stratified LLM traces across six models and three benchmarks; agreement with GPT-4o is 95.1\% over 720 binary decisions for the six focal behaviors ($\kappa = 0.902$).
More ambiguous behaviors, especially self-awareness and some organizational labels, show lower cross-judge agreement.
\section{Results}
\label{sec:results}
\subsection{Thinking-Oriented Models Amplify Correction, Search, and Hesitation}
\label{sec:finding1}
\begin{figure*}[t]
\centering
\includegraphics[width=\textwidth]{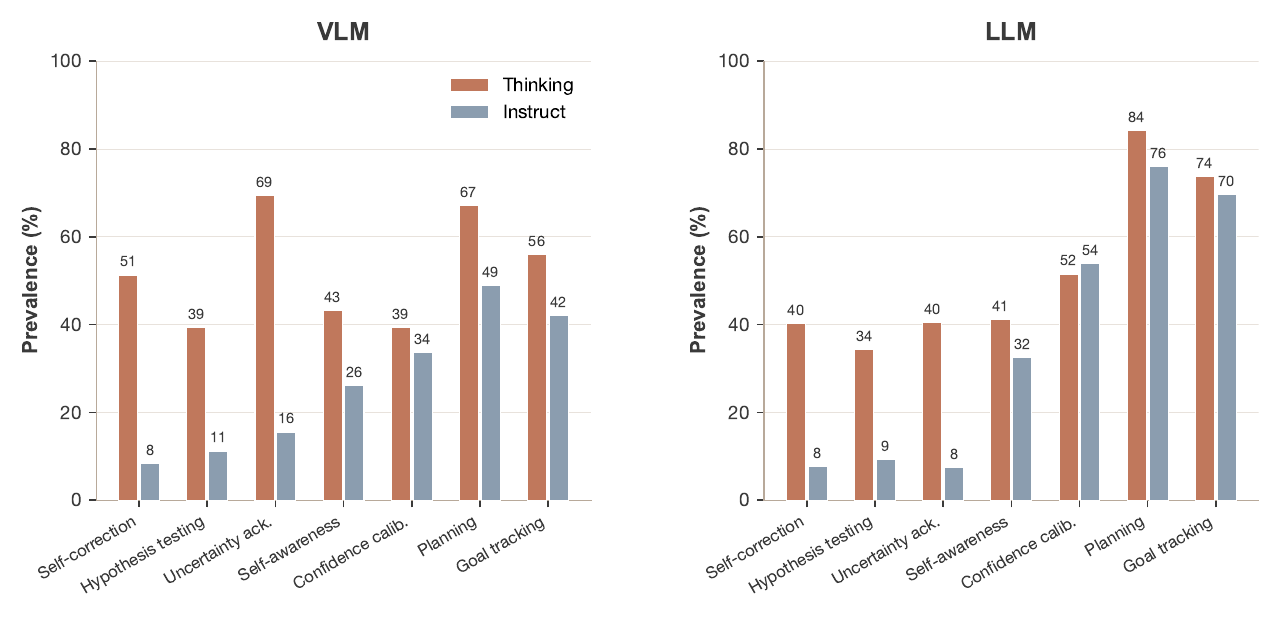}
\caption{Aggregate prevalence of a subset of the nine core higher-order behaviors across all benchmarks per modality. Self-correction, hypothesis testing, and uncertainty acknowledgment show the largest prevalence gaps between thinking and instruct models, while the remaining behaviors are more similar across training types.}
\label{fig:metacognitive_gap}
\end{figure*}

Thinking training selectively amplifies correction, search, and hesitation, while leaving most other core behaviors largely unchanged.

\textbf{Three behaviors are amplified.}\
Self-correction appears in roughly 21–55\% of thinking-model responses versus 3–15\% for comparison models, with consistently large thinking-minus-comparison gaps across benchmarks (\autoref{tab:main_table_appendix}).
Hypothesis testing (22--52\% vs.\ 4--18\%) and uncertainty acknowledgment (25--85\% vs.\ 4--28\%) show comparable gaps.
These gaps hold across all seven model families, both modalities, and all six benchmarks.

\textbf{The remaining behaviors are largely unchanged.}\
Confidence calibration is roughly equal between thinking and instruct models, and on several LLM benchmarks instruct models score slightly higher (66.1\% vs.\ 67.9\% on MATH-500; 40.9\% vs.\ 46.7\% on MMLU-Pro).
Planning and goal tracking are nearly identical on MATH-500 and MMLU-Pro (planning 87--91\% for thinking vs.\ 84--92\% for instruct), with moderate differences on VLM benchmarks.

\subsection{The Most Amplified Behaviors Are Not the Highest-Lift Behaviors}
\label{sec:finding2}
\begin{figure*}[t]
\centering
\includegraphics[width=\textwidth]{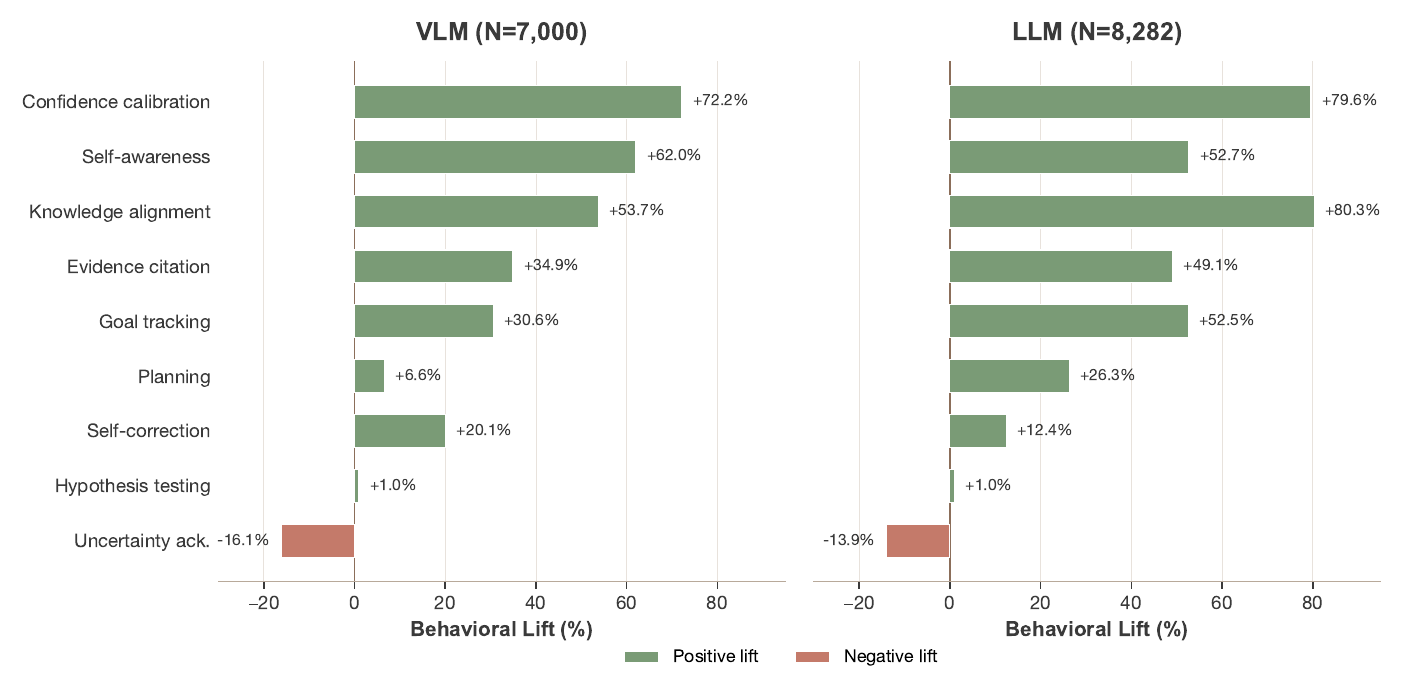}
\caption{Behavioral Lift for the nine cross-modal higher-order behaviors, ordered to highlight the relationship between amplified behaviors and high-Lift behaviors. Bars show Lift only: confidence calibration ranks highest for VLMs and among the highest for LLMs, while the three behaviors identified as amplified in the prevalence analysis cluster near the bottom. Uncertainty acknowledgment has negative Lift in both modalities.}
\label{fig:lift}
\vspace{-10pt}
\end{figure*}

The prevalence analysis identifies three behaviors amplified by thinking training and six that remain largely unchanged. We then compute Behavioral Lift (\autoref{fig:lift}; full statistics for LLMs and VLMs appear in \autoref{tab:lift_detail_llm} and \autoref{tab:lift_detail_vlm}, respectively), which shows that the amplified behaviors are not the behaviors most associated with success.

\textbf{Confidence calibration is one of the strongest positive signals.}\
Across VLM benchmarks, confidence calibration shows +72.2\% Lift (98.8\% accuracy when present vs.\ 26.7\% when absent); across LLM benchmarks, +79.6\% (99.6\% vs.\ 20.0\%).
However, thinking models are no more likely to exhibit this behavior than instruct models.
Confidence calibration is also largely absent when reasoning is weak: it appears in only 7.6\% (VLM) and 3.4\% (LLM) of lucky guesses and 3.4\% (VLM) and 1.6\% (LLM) of post-hoc rationalization traces, versus 94.7\% and 95.9\% on sound traces (Appendix~\autoref{tab:posthoc_prevalence_vlm} and \autoref{tab:posthoc_prevalence_llm}).
Confidence calibration is rare when the answer is correct but the reasoning is weak and common when the reasoning is sound, indicating that it tracks reasoning quality beyond final-answer correctness.

\textbf{The amplified behaviors rank lowest by Behavioral Lift.}\
Uncertainty acknowledgment Lift is $-$16.1\% (VLM) and $-$13.9\% (LLM).
Hypothesis testing Lift is +1.0\% in both modalities.
Self-correction shows modest positive Lift (+20.1\% VLM, +12.4\% LLM), stronger in thinking models than instruct. Appendix~\autoref{tab:posthoc_prevalence_vlm} and \autoref{tab:posthoc_prevalence_llm} show the complementary pattern for the amplified behaviors: uncertainty acknowledgment, and in LLMs also hypothesis testing and self-correction, appear more often on post-hoc traces than on sound traces.

\textbf{The ranking is stable across modalities, benchmarks, and controls.}\
Confidence calibration, self-awareness, and knowledge alignment rank highest in both modalities, while hypothesis testing and uncertainty acknowledgment rank lowest.
This pattern is not an artifact of pooling: the same broad ranking appears within individual benchmarks (\autoref{fig:lift_heatmap}) and within each response-complexity stratum (\autoref{fig:length_control}), and remains visible in per-model analyses (Appendix \autoref{fig:lift_per_llm}, \autoref{fig:lift_per_vlm}). The same ordering appears under reverse conditioning: behaviors enriched in correct traces are the same ones with the highest Lift, while uncertainty acknowledgment is more common in incorrect traces (Appendix ~\autoref{tab:conditional_prevalence}). As an additional validation, linear probes trained on Qwen3-4B hidden states recover several annotated behaviors above chance, and in the thinking model the probe decodability ranking is directionally aligned with Behavioral Lift, especially on incorrect traces (Appendix~\ref{app:probing}).

\begin{table}[t]
\centering
\small
\caption{Within-question analysis across language-only and vision-language models. Values are mean per-question $\Delta$ accuracy between traces where a behavior is present versus absent, computed from 8 samples per question. Confidence calibration remains among the strongest positive signals, while uncertainty acknowledgment is null or negative. Asterisks indicate bootstrap 95\% confidence intervals excluding zero; full results appear in Appendix~\ref{sec:robustness}, \autoref{tab:within_question_llm} (LLMs), and \autoref{tab:within_question_vlm} (VLMs).}
\label{tab:within_question_compact}
\begin{tabular}{lcccc}
\toprule
\textbf{Behavior} & \textbf{LLM-Think} & \textbf{LLM-Inst} & \textbf{VLM-Think} & \textbf{VLM-Inst} \\
\midrule
Confidence calibration      & +0.305$^{*}$ & +0.518$^{*}$ & +0.186$^{*}$ & +0.342$^{*}$ \\
Self-correction             & +0.267$^{*}$ & +0.078$^{*}$ & +0.153$^{*}$ & +0.062$^{*}$ \\
Hypothesis testing          & +0.048$^{*}$ & +0.028        & -0.005        & +0.045        \\
Uncertainty acknowledgment  & +0.006        & -0.094$^{*}$ & -0.017        & -0.004        \\
\bottomrule
\end{tabular}
\end{table}

\textbf{Within-question analysis confirms the pattern.}\
To control for question difficulty, we generate 8 traces per question from Qwen3-4B on 250 MATH-500 problems and compare accuracy between traces where a behavior is present versus absent on the same question.
Confidence calibration shows a large within-question advantage (+0.31 thinking, +0.52 instruct; 95\% CIs exclude zero).
Uncertainty acknowledgment is non-predictive for thinking models (+0.01) and mildly negative for instruct ($-$0.09, CI excludes zero).
Self-correction is stronger in thinking models (+0.27) than instruct (+0.08).
This pattern persists under within-question comparisons (\autoref{tab:within_question_compact}; extended discussion in Appendix~\ref{sec:robustness}).

The same-question and probing analyses suggest that the ranking is not a
question-difficulty artifact or an arbitrary surface-labeling effect.

\subsection{Thinking Models Often Succeed Through Recovery}
\label{sec:finding3}

If the amplified behaviors are not the most predictive, how do thinking models achieve higher accuracy? On extended-reasoning tasks, we find that thinking models often recover after detected failures; on pattern-matching tasks, they gain little or are outperformed. Full failure rates appear in Appendix~\autoref{tab:failures}.

\textbf{Recovery is task-dependent.}\
Recovery Rate, accuracy conditional on at least one failure being detected, varies sharply across benchmarks (\autoref{tab:main_table_appendix}, bottom row).
On benchmarks that reward step-by-step computation, thinking models recover at 2--3$\times$ the rate of instruct models:
VisualPuzzles (23.0\% vs.\ 8.4\%, 2.7$\times$), MATH-500 (40.8\% vs.\ 17.8\%, 2.3$\times$), and MMLU-Pro (16.8\% vs.\ 6.3\%, 2.7$\times$).
On mixed benchmarks, recovery rates are comparable: MathVista (45.9\% vs.\ 47.5\%) and MMMU (32.9\% vs.\ 33.8\%).
On LogiQA2, a pattern-matching task, the pattern reverses: instruct models recover better than thinking models (24.5\% vs. 11.1\%).

\textbf{Some behaviors matter mainly after failures occur.}\
Self-correction has only modest overall Lift, but among traces with at least one detected failure it is strongly associated with recovery (+39pp; ~\autoref{fig:recovery_conditional}).
This helps explain the Behavioral Lift and Recovery Rate results: some behaviors matter most after a trace has entered an error state.

\textbf{Thinking gains come from fewer failures or better recovery.}\
On MathVista and MMMU, thinking models win through \textit{fewer failures}: instruct models show logical failure rates of 53--55\% versus 32--45\% for thinking models, with comparable recovery rates.
On LLM benchmarks and VisualPuzzles, thinking models sometimes exhibit \textit{more} failures (51.1\% vs.\ 41.4\% logical failure on MMLU-Pro), but recover from these failures at much higher rates.
Both mechanisms lead to higher accuracy for thinking models on 4 of 6 benchmarks.

\textbf{LogiQA2 functions as a counterexample to the ``more thinking helps'' assumption.}\
Instruct models outperform thinking models on LogiQA2 (58.4\% vs.\ 54.1\%) by taking more shortcuts (34.2\% vs.\ 20.3\%) that work. This LogiQA2 task rewards recognizing argument structures quickly. These findings suggest that thinking helps when tasks reward computation, but can hurt model performance when the task primarily rewards pattern recognition.
When no failures are detected, both model types reach 96--99\% accuracy, suggesting that the performance difference between instruct models and thinking models in this case is due to model ability to recover from failures.

\subsection{OLMo-3 SFT Show the Amplification–Lift Mismatch Before DPO/RLVR}
The main comparisons in our paper use released model variants, where each checkpoint reflects a full
post-training recipe. To examine this phenomenon at an earlier point in model training, we evaluate
OLMo-3-7B-Think-SFT and OLMo-3-7B-Instruct-SFT on MATH-500 and MMLU-Pro
($N=1{,}443$). These checkpoints share the OLMo-3-7B base and precede the later DPO
and RLVR stages.

The same amplification-Lift gap appears at the SFT checkpoint stage. The Think-SFT branch shows
 higher prevalence of the deliberative behaviors spanning self-correction (34.5\% vs. 1.0\%),
hypothesis testing (22.7\% vs. 1.0\%), and uncertainty acknowledgment (33.1\% vs. 1.0\%).
However, the highest within-checkpoint Lift comes from knowledge alignment and confidence calibration (+81.0\%/+67.4\% in Think-SFT; +81.4\%/+81.3\% in Instruct-SFT), and the  amplified behaviors have lower Lift: self-correction (+16.4\%/+20.6\%) and uncertainty acknowledgment (-16.3\%/-22.7\%). Hypothesis testing is near zero in both checkpoints (-6.1\%/+6.1\%), consistent with its near-zero, least-stable Lift throughout other experiments described in this paper. The gap is therefore visible before the later DPO and RLVR stages, though these comparisons do not isolate the causal contribution of any single training stage.

\subsection{The Amplification-Lift Gap Persists at Scale}
\label{sec:scaling}

We test whether the amplification gap persists at scale by evaluating Qwen3-VL (Think/Instruct) from 2B to 32B on MathVista (\autoref{fig:scaling_analysis}) and Qwen3 (Think) and Qwen2.5 (Instruct) on MATH-500 (\autoref{fig:llm_scaling}).

\textbf{The Amplification-Lift gap persists.}\
At 32B, thinking models  self-correct in 61.3\% of responses versus 11.3\% for instruct, a 50-point gap comparable to the pattern observed for 2B models.
Hypothesis testing and uncertainty acknowledgment show the same pattern (\autoref{fig:scaling_analysis}, left).
We find that the gaps on the less-amplified behaviors narrow or reverse with scale: confidence calibration reaches 78.0\% for the 32B instruct model versus 71.7\% for thinking (\autoref{fig:scaling_analysis}, right). Consistent patterns appear for Qwen3 LLMs on MATH-500 (\autoref{fig:llm_scaling}, \autoref{tab:scaling_llm}).

\textbf{Self-correction Lift diminishes at scale.}\
At 2B, self-correction provides +30.0\% Lift and an 18.3-point advantage.
At 32B, self-correction Lift drops to +5.9\% and the accuracy gap shrinks to 1.7 points. Confidence calibration Lift increases with scale (+57.7\% at 2B to +68.7\% at 32B), reinforcing the mismatch between amplified behaviors and high-lift behaviors.

\textbf{Larger thinking models express less uncertainty.}\
Uncertainty acknowledgment drops from 66.7\% at 2B to 43.7\% at 32B in thinking models, while instruct models remain flat at 8--15\%.
This finding supports the interpretation that uncertainty acknowledgment tracks difficulty, rather than functioning as a measure of calibration.

In a supplementary frontier-model validation on the full GPQA-Diamond benchmark (198 questions across 7 models; $N{=}1{,}386$), seven models from five providers show the same Lift ranking in visible responses (Appendix \autoref{tab:frontier}).

\definecolor{resultboxbg}{HTML}{FAF9F5}
\definecolor{resultboxframe}{HTML}{8B6E5A}
\definecolor{resultboxtitle}{HTML}{3A3A3A}

\begin{tcolorbox}[
    enhanced,
    colframe=resultboxframe,
    colback=resultboxbg,
    boxrule=0.7pt,
    arc=2pt,
    left=8pt,
    right=8pt,
    top=7pt,
    bottom=7pt,
    before skip=8pt,
    after skip=10pt,
    title=\textbf{Summary of Results},
    coltitle=white,
    colbacktitle=resultboxframe,
    titlerule=0pt,
    toptitle=3pt,
    bottomtitle=3pt,
    lefttitle=8pt,
    righttitle=8pt
]
\small
Thinking-oriented models produce more deliberative traces, but this added deliberation is not concentrated in the behaviors most associated with reasoning correctness. The largest lifts are associated with confidence calibration, knowledge alignment, and self-awareness rather than visible search or hesitation. Thinking models help most on tasks that reward recovery from intermediate failures. The amplification–Lift gap persists across benchmarks, model families, and scale.
\end{tcolorbox}

\section{Discussion}
\label{sec:discussion}
Amplified deliberation does not track correctness because accuracy cannot tell a frequent behavior from a useful one. Self-correction is one example: it can mark a successful recovery, but it can equally mark that the reasoning went wrong earlier. Uncertainty is another: it can reflect appropriate caution or simple confusion. Behavioral Lift separates these cases by measuring how much correctness differs when a behavior is present versus absent.

Confidence calibration and uncertainty acknowledgment are easy to conflate, yet only one tracks correctness. Confidence calibration means expressed certainty tracks the strength of the reasoning: the model is cautious when evidence is weak and decisive when the reasoning is strong. Uncertainty acknowledgment is only the presence of explicit doubt or hesitation. Its weak or negative association with correctness suggests that explicit uncertainty is informative only when it tracks the available evidence~\citep{kim2026does}. A model that says ``maybe'' at every step is not calibrated.

This distinction has direct implications for training and evaluation. Current reasoning-oriented training often rewards final-answer correctness, which can produce long traces containing more correction, search, and hesitation; a line of recent work targets this inefficiency directly~\citep{ma2025reasoning, arora2025training}. Visible deliberation, however, is an unreliable proxy for reasoning quality. Our recovery results sharpen this point: thinking helps most when tasks reward extended computation or recovery from intermediate mistakes, and can hurt when fast recognition of logical form is sufficient.
Future evaluations should therefore report both whether thinking helps and how: by preventing failures, recovering from them, improving calibration, or changing the shortcuts models take.

For process supervision, reward behaviors associated with correctness; surface markers of deliberation are insufficient on their own. A process objective should reward evidence-grounded claims, appropriate domain framing, recognition of underspecified information, and confidence that tracks reasoning strength. Longer traces, backtracking, and expressions of uncertainty are not objectives in themselves.
Behavioral Lift can audit whether a process objective rewards behaviors associated with success or only rewards deliberative surface form.

Several limitations remain. Our analysis is limited to visible traces; written reasoning may be incomplete, post hoc~\citep{boppana2026reasoning}, or unfaithful to the model's internal computation~\citep{chen2025reasoning, baker2025monitoring}. Our labels are produced by automated judges, so systematic judge bias remains possible despite multi-judge validation, robustness checks, same-question controls, and probing analyses.
Finally, Behavioral Lift is descriptive: high-lift behaviors may cause better performance, reflect that the model is already on the right track, or co-occur with another useful property.
Prompting results (Appendix~\ref{prompting_test}) are consistent with the ranking, but more controlled training studies are needed.
\section{Related Work}
\label{sec:related_work}

\paragraph{Reasoning-oriented training and long chain-of-thought.}
Recent reasoning models such as DeepSeek-R1~\citep{guo2025deepseek}, Kimi-k1.5~\citep{kimi2025k15}, and Qwen3~\citep{qwen2025qwen3} show that RL or hybrid post-training can elicit long reasoning traces with behaviors such as backtracking and self-correction~\citep{chu2025sft}.
\citet{yeo2025demystifying, yue2025does} study how long CoT reasoning emerges during training, finding that some core abilities are already present in base models but require substantial RL compute to be elicited reliably.
Our work studies the behavioral consequences of this elicitation: thinking-oriented models amplify self-correction, hypothesis testing, and uncertainty acknowledgment~\citep{zhang2025interplay}, but these are not the behaviors most associated with correctness.

\paragraph{Behavioral analysis of reasoning traces.}
Recent work analyzes reasoning traces through cognitive behaviors, taxonomies, strategy discovery, and trace dynamics.
\citet{gandhi2025cognitive} identify cognitive behaviors that predict RL trainability, \citet{kargupta2025cognitive} annotate large-scale traces with a cognitively grounded taxonomy, \citet{lee2025cot} cluster and steer reasoning strategies from CoT traces, \citet{chang2026prism} analyze reasoning through semantic flow and latent computation, and \citet{wang2025thoughts} study instability and underthinking in o1-like reasoning traces.
We extend this line of work by comparing thinking and instruct variants across LLMs and VLMs, and by separating behaviors that are frequent from behaviors that have high Behavioral Lift.

\paragraph{Process supervision and evaluation beyond accuracy.}
Chain-of-thought prompting showed that explicit reasoning can improve final-answer accuracy~\citep{wei2022chain}, while process supervision and process reward models evaluate reasoning at the step level~\citep{lightman2024let}.
Recent process-evaluation benchmarks and verifier-style methods further test whether models can identify reasoning errors or avoid shortcut solutions rather than only produce correct final answers~\citep{zheng2025processbench, zhong2025impossiblebench}.
Our results give a concrete criterion for process supervision: before rewarding a reasoning behavior, one should ask whether it is associated with correctness.
\section{Conclusion}
\label{sec:conclusion}

We quantified the behavioral effects of thinking training across 15,282 traces from 15 models and 6 benchmarks.
Thinking training consistently amplifies self-correction, hypothesis testing, and uncertainty acknowledgment, while leaving several higher-lift behaviors largely unchanged.
The central finding is that the amplified behaviors are not the ones most associated with correctness: confidence calibration is one of the strongest positive signals, whereas uncertainty acknowledgment is often non-predictive or negative.
These results suggest that improving reasoning models requires more than making them think longer.
Future training and evaluation should reward the behaviors that make reasoning reliable: calibrated confidence, grounded use of evidence, appropriate domain knowledge, and recovery when reasoning goes wrong.
\newpage

\section*{Acknowledgments}

We thank Xiang Yue, Jacob Springer, and Seungone Kim for helpful discussions and feedback throughout the development of this work.
\section*{Ethics Statement}
This paper studies reasoning behavior in open-weight language and vision-language models using automated annotation of model outputs. Our analysis is observational and focuses on benchmark responses rather than deployment in real-world decision settings. We will release prompts, metrics code, and annotations to support transparency and reproducibility. We do not claim that visible traces fully reflect internal reasoning, and we discuss this limitation explicitly in the paper.

LLMs were used to assist with grammar and editing, and with implementation and verification of selected analyses. LLM-based annotation is described in the methodology.

\bibliography{colm2026_conference}

\begin{thebibliography}{32}
\providecommand{\natexlab}[1]{#1}
\providecommand{\url}[1]{\texttt{#1}}
\expandafter\ifx\csname urlstyle\endcsname\relax
  \providecommand{\doi}[1]{doi: #1}\else
  \providecommand{\doi}{doi: \begingroup \urlstyle{rm}\Url}\fi

\bibitem[Arora \& Zanette(2025)Arora and Zanette]{arora2025training}
Daman Arora and Andrea Zanette.
\newblock Training language models to reason efficiently, 2025.
\newblock URL \url{https://arxiv.org/abs/2502.04463}.

\bibitem[Baker et~al.(2025)Baker, Huizinga, Gao, Dou, Guan, Madry, Zaremba, Pachocki, and Farhi]{baker2025monitoring}
Bowen Baker, Joost Huizinga, Leo Gao, Zehao Dou, Melody~Y. Guan, Aleksander Madry, Wojciech Zaremba, Jakub Pachocki, and David Farhi.
\newblock Monitoring reasoning models for misbehavior and the risks of promoting obfuscation, 2025.
\newblock URL \url{https://arxiv.org/abs/2503.11926}.

\bibitem[Boppana et~al.(2026)Boppana, Ma, Loeffler, Sarfati, Bigelow, Geiger, Lewis, and Merullo]{boppana2026reasoning}
Siddharth Boppana, Annabel Ma, Max Loeffler, Raphael Sarfati, Eric Bigelow, Atticus Geiger, Owen Lewis, and Jack Merullo.
\newblock Reasoning theater: Disentangling model beliefs from chain-of-thought, 2026.
\newblock URL \url{https://arxiv.org/abs/2603.05488}.

\bibitem[Brown(1987)]{brown1987metacognition}
A.~Brown.
\newblock Metacognition, executive control, self-regulation, and other more mysterious mechanisms.
\newblock 1987.
\newblock URL \url{https://api.semanticscholar.org/CorpusID:147157394}.

\bibitem[Chang et~al.(2026)Chang, Zhou, and Chen]{chang2026prism}
Ruidi Chang, Jiawei Zhou, and Hanjie Chen.
\newblock Prism: A dual view of llm reasoning through semantic flow and latent computation, 2026.
\newblock URL \url{https://arxiv.org/abs/2603.22754}.

\bibitem[Chen et~al.(2025)Chen, Benton, Radhakrishnan, Uesato, Denison, Schulman, Somani, Hase, Wagner, Roger, Mikulik, Bowman, Leike, Kaplan, and Perez]{chen2025reasoning}
Yanda Chen, Joe Benton, Ansh Radhakrishnan, Jonathan Uesato, Carson Denison, John Schulman, Arushi Somani, Peter Hase, Misha Wagner, Fabien Roger, Vlad Mikulik, Samuel~R. Bowman, Jan Leike, Jared Kaplan, and Ethan Perez.
\newblock Reasoning models don't always say what they think, 2025.
\newblock URL \url{https://arxiv.org/abs/2505.05410}.

\bibitem[Chu et~al.(2025)Chu, Zhai, Yang, Tong, Xie, Schuurmans, Le, Levine, and Ma]{chu2025sft}
Tianzhe Chu, Yuexiang Zhai, Jihan Yang, Shengbang Tong, Saining Xie, Dale Schuurmans, Quoc~V. Le, Sergey Levine, and Yi~Ma.
\newblock Sft memorizes, rl generalizes: A comparative study of foundation model post-training, 2025.
\newblock URL \url{https://arxiv.org/abs/2501.17161}.

\bibitem[Gandhi et~al.(2025)Gandhi, Chakravarthy, Singh, Lile, and Goodman]{gandhi2025cognitive}
Kanishk Gandhi, Ayush Chakravarthy, Anikait Singh, Nathan Lile, and Noah~D. Goodman.
\newblock Cognitive behaviors that enable self-improving reasoners, or, four habits of highly effective stars, 2025.
\newblock URL \url{https://arxiv.org/abs/2503.01307}.

\bibitem[Guo et~al.(2025)Guo, Yang, Zhang, Song, Wang, Zhu, Xu, Zhang, Ma, Bi, Zhang, Yu, Wu, Wu, Gou, Shao, Li, Gao, Liu, Xue, Wang, Wu, Feng, Lu, Zhao, Deng, Ruan, Dai, Chen, Ji, Li, Lin, Dai, Luo, Hao, Chen, Li, Zhang, Xu, Ding, Gao, Qu, Li, Guo, Li, Chen, Yuan, Tu, Qiu, Li, Cai, Ni, Liang, Chen, Dong, Hu, You, Gao, Guan, Huang, Yu, Wang, Zhang, Zhao, Wang, Zhang, Xu, Xia, Zhang, Zhang, Tang, Zhou, Li, Wang, Li, Tian, Huang, Zhang, Wang, Chen, Du, Ge, Zhang, Pan, Wang, Chen, Jin, Chen, Lu, Zhou, Chen, Ye, Wang, Yu, Zhou, Pan, Li, Zhou, Wu, Yun, Pei, Sun, Wang, Zeng, Liu, Liang, Gao, Yu, Zhang, Xiao, An, Liu, Wang, Chen, Nie, Cheng, Liu, Xie, Liu, Yang, Li, Su, Lin, Li, Jin, Shen, Chen, Sun, Wang, Song, Zhou, Wang, Shan, Li, Wang, Wei, Zhang, Xu, Li, Zhao, Sun, Wang, Yu, Zhang, Shi, Xiong, He, Piao, Wang, Tan, Ma, Liu, Guo, Ou, Wang, Gong, Zou, He, Xiong, Luo, You, Liu, Zhou, Zhu, Huang, Li, Zheng, Zhu, Ma, Tang, Zha, Yan, Ren, Ren, Sha, Fu, Xu, Xie, Zhang, Hao, Ma, Yan, Wu, Gu, Zhu, Liu, Li, Xie, Song,
  Pan, Huang, Xu, Zhang, and Zhang]{guo2025deepseek}
Daya Guo, Dejian Yang, Haowei Zhang, Junxiao Song, Peiyi Wang, Qihao Zhu, Runxin Xu, Ruoyu Zhang, Shirong Ma, Xiao Bi, Xiaokang Zhang, Xingkai Yu, Yu~Wu, Z.~F. Wu, Zhibin Gou, Zhihong Shao, Zhuoshu Li, Ziyi Gao, Aixin Liu, Bing Xue, Bingxuan Wang, Bochao Wu, Bei Feng, Chengda Lu, Chenggang Zhao, Chengqi Deng, Chong Ruan, Damai Dai, Deli Chen, Dongjie Ji, Erhang Li, Fangyun Lin, Fucong Dai, Fuli Luo, Guangbo Hao, Guanting Chen, Guowei Li, H.~Zhang, Hanwei Xu, Honghui Ding, Huazuo Gao, Hui Qu, Hui Li, Jianzhong Guo, Jiashi Li, Jingchang Chen, Jingyang Yuan, Jinhao Tu, Junjie Qiu, Junlong Li, J.~L. Cai, Jiaqi Ni, Jian Liang, Jin Chen, Kai Dong, Kai Hu, Kaichao You, Kaige Gao, Kang Guan, Kexin Huang, Kuai Yu, Lean Wang, Lecong Zhang, Liang Zhao, Litong Wang, Liyue Zhang, Lei Xu, Leyi Xia, Mingchuan Zhang, Minghua Zhang, Minghui Tang, Mingxu Zhou, Meng Li, Miaojun Wang, Mingming Li, Ning Tian, Panpan Huang, Peng Zhang, Qiancheng Wang, Qinyu Chen, Qiushi Du, Ruiqi Ge, Ruisong Zhang, Ruizhe Pan, Runji Wang, R.~J.
  Chen, R.~L. Jin, Ruyi Chen, Shanghao Lu, Shangyan Zhou, Shanhuang Chen, Shengfeng Ye, Shiyu Wang, Shuiping Yu, Shunfeng Zhou, Shuting Pan, S.~S. Li, Shuang Zhou, Shaoqing Wu, Tao Yun, Tian Pei, Tianyu Sun, T.~Wang, Wangding Zeng, Wen Liu, Wenfeng Liang, Wenjun Gao, Wenqin Yu, Wentao Zhang, W.~L. Xiao, Wei An, Xiaodong Liu, Xiaohan Wang, Xiaokang Chen, Xiaotao Nie, Xin Cheng, Xin Liu, Xin Xie, Xingchao Liu, Xinyu Yang, Xinyuan Li, Xuecheng Su, Xuheng Lin, X.~Q. Li, Xiangyue Jin, Xiaojin Shen, Xiaosha Chen, Xiaowen Sun, Xiaoxiang Wang, Xinnan Song, Xinyi Zhou, Xianzu Wang, Xinxia Shan, Y.~K. Li, Y.~Q. Wang, Y.~X. Wei, Yang Zhang, Yanhong Xu, Yao Li, Yao Zhao, Yaofeng Sun, Yaohui Wang, Yi~Yu, Yichao Zhang, Yifan Shi, Yiliang Xiong, Ying He, Yishi Piao, Yisong Wang, Yixuan Tan, Yiyang Ma, Yiyuan Liu, Yongqiang Guo, Yuan Ou, Yuduan Wang, Yue Gong, Yuheng Zou, Yujia He, Yunfan Xiong, Yuxiang Luo, Yuxiang You, Yuxuan Liu, Yuyang Zhou, Y.~X. Zhu, Yanping Huang, Yaohui Li, Yi~Zheng, Yuchen Zhu, Yunxian Ma, Ying
  Tang, Yukun Zha, Yuting Yan, Z.~Z. Ren, Zehui Ren, Zhangli Sha, Zhe Fu, Zhean Xu, Zhenda Xie, Zhengyan Zhang, Zhewen Hao, Zhicheng Ma, Zhigang Yan, Zhiyu Wu, Zihui Gu, Zijia Zhu, Zijun Liu, Zilin Li, Ziwei Xie, Ziyang Song, Zizheng Pan, Zhen Huang, Zhipeng Xu, Zhongyu Zhang, and Zhen Zhang.
\newblock Deepseek-r1 incentivizes reasoning in llms through reinforcement learning.
\newblock \emph{Nature}, 645\penalty0 (8081):\penalty0 633–638, 2025.
\newblock ISSN 1476-4687.
\newblock \doi{10.1038/s41586-025-09422-z}.
\newblock URL \url{http://dx.doi.org/10.1038/s41586-025-09422-z}.

\bibitem[Hendrycks et~al.(2021)Hendrycks, Burns, Kadavath, Arora, Basart, Tang, Song, and Steinhardt]{hendrycksmath2021}
Dan Hendrycks, Collin Burns, Saurav Kadavath, Akul Arora, Steven Basart, Eric Tang, Dawn Song, and Jacob Steinhardt.
\newblock Measuring mathematical problem solving with the math dataset, 2021.
\newblock URL \url{https://arxiv.org/abs/2103.03874}.

\bibitem[Kargupta et~al.(2025)Kargupta, Li, Wang, Lee, Chen, Ahia, Light, Griffiths, Kleiman-Weiner, Han, Celikyilmaz, and Tsvetkov]{kargupta2025cognitive}
Priyanka Kargupta, Shuyue~Stella Li, Haocheng Wang, Jinu Lee, Shan Chen, Orevaoghene Ahia, Dean Light, Thomas~L. Griffiths, Max Kleiman-Weiner, Jiawei Han, Asli Celikyilmaz, and Yulia Tsvetkov.
\newblock Cognitive foundations for reasoning and their manifestation in llms, 2025.
\newblock URL \url{https://arxiv.org/abs/2511.16660}.

\bibitem[Kim et~al.(2026)Kim, Luo, Kim, Lee, Kim, Jeon, Li, and Yang]{kim2026does}
Jeonghye Kim, Xufang Luo, Minbeom Kim, Sangmook Lee, Dohyung Kim, Jiwon Jeon, Dongsheng Li, and Yuqing Yang.
\newblock Why does self-distillation (sometimes) degrade the reasoning capability of llms?, 2026.
\newblock URL \url{https://arxiv.org/abs/2603.24472}.

\bibitem[Lee et~al.(2025)Lee, Kim, Seo, Jo, Go, Hwang, Park, Yue, Welleck, Neubig, Lee, and Seo]{lee2025cot}
Seongyun Lee, Seungone Kim, Minju Seo, Yongrae Jo, Dongyoung Go, Hyeonbin Hwang, Jinho Park, Xiang Yue, Sean Welleck, Graham Neubig, Moontae Lee, and Minjoon Seo.
\newblock The cot encyclopedia: Analyzing, predicting, and controlling how a reasoning model will think, 2025.
\newblock URL \url{https://arxiv.org/abs/2505.10185}.

\bibitem[Lightman et~al.(2023)Lightman, Kosaraju, Burda, Edwards, Baker, Lee, Leike, Schulman, Sutskever, and Cobbe]{lightman2024let}
Hunter Lightman, Vineet Kosaraju, Yura Burda, Harri Edwards, Bowen Baker, Teddy Lee, Jan Leike, John Schulman, Ilya Sutskever, and Karl Cobbe.
\newblock Let's verify step by step, 2023.
\newblock URL \url{https://arxiv.org/abs/2305.20050}.

\bibitem[Liu et~al.(2020)Liu, Cui, Liu, Huang, Wang, and Zhang]{liu2020logiqa}
Jian Liu, Leyang Cui, Hanmeng Liu, Dandan Huang, Yile Wang, and Yue Zhang.
\newblock Logiqa: A challenge dataset for machine reading comprehension with logical reasoning, 2020.
\newblock URL \url{https://arxiv.org/abs/2007.08124}.

\bibitem[Lu et~al.(2024)Lu, Bansal, Xia, Liu, Li, Hajishirzi, Cheng, Chang, Galley, and Gao]{lu2023mathvista}
Pan Lu, Hritik Bansal, Tony Xia, Jiacheng Liu, Chunyuan Li, Hannaneh Hajishirzi, Hao Cheng, Kai-Wei Chang, Michel Galley, and Jianfeng Gao.
\newblock Mathvista: Evaluating mathematical reasoning of foundation models in visual contexts, 2024.
\newblock URL \url{https://arxiv.org/abs/2310.02255}.

\bibitem[Ma et~al.(2025)Ma, He, Snell, Griggs, Min, and Zaharia]{ma2025reasoning}
Wenjie Ma, Jingxuan He, Charlie Snell, Tyler Griggs, Sewon Min, and Matei Zaharia.
\newblock Reasoning models can be effective without thinking, 2025.
\newblock URL \url{https://arxiv.org/abs/2504.09858}.

\bibitem[Nelson(1990)]{nelson1990metamemory}
Thomas~O. Nelson.
\newblock Metamemory: A theoretical framework and new findings.
\newblock \emph{Psychology of Learning and Motivation}, 26:\penalty0 125--173, 1990.
\newblock URL \url{https://api.semanticscholar.org/CorpusID:39951989}.

\bibitem[{OpenAI}(2024)]{openai2024o1}
{OpenAI}.
\newblock Learning to reason with {LLMs}.
\newblock \url{https://openai.com/index/learning-to-reason-with-llms/}, 2024.
\newblock Technical Blog Post.

\bibitem[Schraw \& Dennison(1994)Schraw and Dennison]{schraw1994assessing}
Gregory Schraw and Rayne~Sperling Dennison.
\newblock Assessing metacognitive awareness.
\newblock \emph{Contemporary Educational Psychology}, 19\penalty0 (4):\penalty0 460--475, 1994.
\newblock ISSN 0361-476X.
\newblock \doi{https://doi.org/10.1006/ceps.1994.1033}.
\newblock URL \url{https://www.sciencedirect.com/science/article/pii/S0361476X84710332}.

\bibitem[Song et~al.(2025)Song, Ou, Kong, Li, Neubig, and Yue]{song2025visualpuzzles}
Yueqi Song, Tianyue Ou, Yibo Kong, Zecheng Li, Graham Neubig, and Xiang Yue.
\newblock Visualpuzzles: Decoupling multimodal reasoning evaluation from domain knowledge, 2025.
\newblock URL \url{https://arxiv.org/abs/2504.10342}.

\bibitem[Team et~al.(2025)Team, Du, Gao, Xing, Jiang, Chen, Li, Xiao, Du, Liao, Tang, Wang, Zhang, Yuan, Lu, Tang, Sung, Wei, Lai, Guo, Zhu, Ding, Hu, Yang, Zhang, Yao, Zhao, Lu, Li, Yu, Gao, Zheng, Yuan, Chen, Guo, Su, Wang, Zhao, Zhang, Liu, Yan, Wu, Shi, Ye, Yu, Dong, Zhang, Ma, Pan, Gong, Liu, Ma, Wei, Cao, Huang, Jiang, Gao, Xiong, He, Huang, Xu, Wu, He, Wei, Jia, Wu, Xu, Zu, Zhou, Pan, Charles, Li, Hu, Liu, Chen, Wang, Liu, Qin, Liu, Yang, Bao, Du, Wu, Wang, Zhou, Wang, Li, Zhu, Zhang, Wang, Yang, Huang, Huang, Xu, Yang, and Lin]{kimi2025k15}
Kimi Team, Angang Du, Bofei Gao, Bowei Xing, Changjiu Jiang, Cheng Chen, Cheng Li, Chenjun Xiao, Chenzhuang Du, Chonghua Liao, Chuning Tang, Congcong Wang, Dehao Zhang, Enming Yuan, Enzhe Lu, Fengxiang Tang, Flood Sung, Guangda Wei, Guokun Lai, Haiqing Guo, Han Zhu, Hao Ding, Hao Hu, Hao Yang, Hao Zhang, Haotian Yao, Haotian Zhao, Haoyu Lu, Haoze Li, Haozhen Yu, Hongcheng Gao, Huabin Zheng, Huan Yuan, Jia Chen, Jianhang Guo, Jianlin Su, Jianzhou Wang, Jie Zhao, Jin Zhang, Jingyuan Liu, Junjie Yan, Junyan Wu, Lidong Shi, Ling Ye, Longhui Yu, Mengnan Dong, Neo Zhang, Ningchen Ma, Qiwei Pan, Qucheng Gong, Shaowei Liu, Shengling Ma, Shupeng Wei, Sihan Cao, Siying Huang, Tao Jiang, Weihao Gao, Weimin Xiong, Weiran He, Weixiao Huang, Weixin Xu, Wenhao Wu, Wenyang He, Xianghui Wei, Xianqing Jia, Xingzhe Wu, Xinran Xu, Xinxing Zu, Xinyu Zhou, Xuehai Pan, Y.~Charles, Yang Li, Yangyang Hu, Yangyang Liu, Yanru Chen, Yejie Wang, Yibo Liu, Yidao Qin, Yifeng Liu, Ying Yang, Yiping Bao, Yulun Du, Yuxin Wu, Yuzhi Wang, Zaida
  Zhou, Zhaoji Wang, Zhaowei Li, Zhen Zhu, Zheng Zhang, Zhexu Wang, Zhilin Yang, Zhiqi Huang, Zihao Huang, Ziyao Xu, Zonghan Yang, and Zongyu Lin.
\newblock Kimi k1.5: Scaling reinforcement learning with llms, 2025.
\newblock URL \url{https://arxiv.org/abs/2501.12599}.

\bibitem[Wang et~al.(2024)Wang, Ma, Zhang, Ni, Chandra, Guo, Ren, Arulraj, He, Jiang, Li, Ku, Wang, Zhuang, Fan, Yue, and Chen]{wang2024mmlu}
Yubo Wang, Xueguang Ma, Ge~Zhang, Yuansheng Ni, Abhranil Chandra, Shiguang Guo, Weiming Ren, Aaran Arulraj, Xuan He, Ziyan Jiang, Tianle Li, Max Ku, Kai Wang, Alex Zhuang, Rongqi Fan, Xiang Yue, and Wenhu Chen.
\newblock Mmlu-pro: A more robust and challenging multi-task language understanding benchmark, 2024.
\newblock URL \url{https://arxiv.org/abs/2406.01574}.

\bibitem[Wang et~al.(2025)Wang, Liu, Xu, Liang, Chen, He, Song, Yu, Li, Zhang, Wang, Tu, Mi, and Yu]{wang2025thoughts}
Yue Wang, Qiuzhi Liu, Jiahao Xu, Tian Liang, Xingyu Chen, Zhiwei He, Linfeng Song, Dian Yu, Juntao Li, Zhuosheng Zhang, Rui Wang, Zhaopeng Tu, Haitao Mi, and Dong Yu.
\newblock Thoughts are all over the place: On the underthinking of o1-like llms, 2025.
\newblock URL \url{https://arxiv.org/abs/2501.18585}.

\bibitem[Wei et~al.(2023)Wei, Wang, Schuurmans, Bosma, Ichter, Xia, Chi, Le, and Zhou]{wei2022chain}
Jason Wei, Xuezhi Wang, Dale Schuurmans, Maarten Bosma, Brian Ichter, Fei Xia, Ed~Chi, Quoc Le, and Denny Zhou.
\newblock Chain-of-thought prompting elicits reasoning in large language models, 2023.
\newblock URL \url{https://arxiv.org/abs/2201.11903}.

\bibitem[Yang et~al.(2025)Yang, Li, Yang, Zhang, Hui, Zheng, Yu, Gao, Huang, Lv, Zheng, Liu, Zhou, Huang, Hu, Ge, Wei, Lin, Tang, Yang, Tu, Zhang, Yang, Yang, Zhou, Zhou, Lin, Dang, Bao, Yang, Yu, Deng, Li, Xue, Li, Zhang, Wang, Zhu, Men, Gao, Liu, Luo, Li, Tang, Yin, Ren, Wang, Zhang, Ren, Fan, Su, Zhang, Zhang, Wan, Liu, Wang, Cui, Zhang, Zhou, and Qiu]{qwen2025qwen3}
An~Yang, Anfeng Li, Baosong Yang, Beichen Zhang, Binyuan Hui, Bo~Zheng, Bowen Yu, Chang Gao, Chengen Huang, Chenxu Lv, Chujie Zheng, Dayiheng Liu, Fan Zhou, Fei Huang, Feng Hu, Hao Ge, Haoran Wei, Huan Lin, Jialong Tang, Jian Yang, Jianhong Tu, Jianwei Zhang, Jianxin Yang, Jiaxi Yang, Jing Zhou, Jingren Zhou, Junyang Lin, Kai Dang, Keqin Bao, Kexin Yang, Le~Yu, Lianghao Deng, Mei Li, Mingfeng Xue, Mingze Li, Pei Zhang, Peng Wang, Qin Zhu, Rui Men, Ruize Gao, Shixuan Liu, Shuang Luo, Tianhao Li, Tianyi Tang, Wenbiao Yin, Xingzhang Ren, Xinyu Wang, Xinyu Zhang, Xuancheng Ren, Yang Fan, Yang Su, Yichang Zhang, Yinger Zhang, Yu~Wan, Yuqiong Liu, Zekun Wang, Zeyu Cui, Zhenru Zhang, Zhipeng Zhou, and Zihan Qiu.
\newblock Qwen3 technical report, 2025.
\newblock URL \url{https://arxiv.org/abs/2505.09388}.

\bibitem[Yeo et~al.(2025)Yeo, Tong, Niu, Neubig, and Yue]{yeo2025demystifying}
Edward Yeo, Yuxuan Tong, Morry Niu, Graham Neubig, and Xiang Yue.
\newblock Demystifying long chain-of-thought reasoning in llms, 2025.
\newblock URL \url{https://arxiv.org/abs/2502.03373}.

\bibitem[Yue et~al.(2024)Yue, Ni, Zhang, Zheng, Liu, Zhang, Stevens, Jiang, Ren, Sun, Wei, Yu, Yuan, Sun, Yin, Zheng, Yang, Liu, Huang, Sun, Su, and Chen]{yue2024mmmu}
Xiang Yue, Yuansheng Ni, Kai Zhang, Tianyu Zheng, Ruoqi Liu, Ge~Zhang, Samuel Stevens, Dongfu Jiang, Weiming Ren, Yuxuan Sun, Cong Wei, Botao Yu, Ruibin Yuan, Renliang Sun, Ming Yin, Boyuan Zheng, Zhenzhu Yang, Yibo Liu, Wenhao Huang, Huan Sun, Yu~Su, and Wenhu Chen.
\newblock Mmmu: A massive multi-discipline multimodal understanding and reasoning benchmark for expert agi, 2024.
\newblock URL \url{https://arxiv.org/abs/2311.16502}.

\bibitem[Yue et~al.(2025)Yue, Chen, Lu, Zhao, Wang, Yue, Song, and Huang]{yue2025does}
Yang Yue, Zhiqi Chen, Rui Lu, Andrew Zhao, Zhaokai Wang, Yang Yue, Shiji Song, and Gao Huang.
\newblock Does reinforcement learning really incentivize reasoning capacity in llms beyond the base model?, 2025.
\newblock URL \url{https://arxiv.org/abs/2504.13837}.

\bibitem[Zhang et~al.(2025)Zhang, Neubig, and Yue]{zhang2025interplay}
Charlie Zhang, Graham Neubig, and Xiang Yue.
\newblock On the interplay of pre-training, mid-training, and rl on reasoning language models, 2025.
\newblock URL \url{https://arxiv.org/abs/2512.07783}.

\bibitem[Zheng et~al.(2025)Zheng, Zhang, Zhang, Lin, Lu, Yu, Liu, Zhou, and Lin]{zheng2025processbench}
Chujie Zheng, Zhenru Zhang, Beichen Zhang, Runji Lin, Keming Lu, Bowen Yu, Dayiheng Liu, Jingren Zhou, and Junyang Lin.
\newblock Processbench: Identifying process errors in mathematical reasoning, 2025.
\newblock URL \url{https://arxiv.org/abs/2412.06559}.

\bibitem[Zhong et~al.(2025)Zhong, Raghunathan, and Carlini]{zhong2025impossiblebench}
Ziqian Zhong, Aditi Raghunathan, and Nicholas Carlini.
\newblock Impossiblebench: Measuring llms' propensity of exploiting test cases, 2025.
\newblock URL \url{https://arxiv.org/abs/2510.20270}.

\end{thebibliography}
\bibliographystyle{colm2026_conference}

\appendix
\section{Inference Details}
\label{app:inference}

\paragraph{Inference and prompting.}
We evaluate LLMs with \texttt{lm-evaluation-harness} and VLMs with \texttt{lmms-eval}, using the benchmark implementations provided by these evaluation suites.
Unless otherwise noted, all runs are zero-shot (\texttt{num\_fewshot=0}), and we use benchmark-native chain-of-thought prompting when a CoT zero-shot variant is available.
For instruct models, prompts therefore follow the benchmark or evaluation-suite templates rather than a paper-specific custom prompt.
Full run scripts, exact subset files, prompt templates, and scoring code will be released with the codebase.

\paragraph{LLM generation settings.}
We run all LLM evaluations with vLLM using family-specific decoding settings chosen to match the recommended regime for each model family.
Reasoning-enabled LLMs---Qwen3 thinking models, OLMo-3-Think, Nemotron-v2 reasoning models, and DeepSeek-R1-Distill---use temperature $0.6$, top-$p$ $0.95$, top-$k$ $20$, and min-$p$ $0$.
Qwen2.5 instruct models use temperature $0.7$, top-$p$ $0.8$, top-$k$ $20$, and min-$p$ $0$.
Nemotron-Base is evaluated with greedy decoding (temperature $0$).
Maximum generation length is 16{,}384 tokens for Qwen scaling runs, 32{,}768 tokens for OLMo-3 and DeepSeek-R1-Distill, and 8{,}192 tokens for Nemotron variants.

\paragraph{VLM generation settings.}
We run all VLM evaluations with vLLM-backend using family-specific decoding settings.
For Qwen3-VL-8B and GLM-4.1V-9B, thinking variants use temperature $1.0$, top-$p$ $0.95$, top-$k$ $20$, repetition penalty $1.0$, presence penalty $0.0$, and a maximum generation length of 40{,}960 tokens; instruct variants use temperature $0.7$, top-$p$ $0.8$, top-$k$ $20$, repetition penalty $1.0$, presence penalty $1.5$, and a maximum generation length of 16{,}384 tokens.
For Kimi-VL-A3B, the thinking variant uses temperature $0.8$ and top-$p$ $0.8$ with sampling enabled, while the instruct variant uses temperature $0.2$ and top-$p$ $0.8$ without sampling; both use a maximum generation length of 16{,}000 tokens.
For InternVL3.5-8B, the thinking variant uses temperature $0.6$, top-$p$ $0.8$, top-$k$ $20$, min-$p$ $0$, and sampling enabled, while the instruct variant uses temperature $0.8$, top-$p$ $0.8$, top-$k$ $20$, min-$p$ $0$, and sampling disabled; both use a maximum generation length of 16{,}000 tokens.

We otherwise retain the model-specific defaults and benchmark wrappers provided by the evaluation frameworks, and we do not add extra prompting intended to elicit particular behaviors.~\autoref{tab:generation_settings} lists the decoding settings used for each model family.
\begin{table*}[t]
\centering
\small
\setlength{\tabcolsep}{5pt}
\caption{Generation settings used for the main evaluations. We evaluate LLMs with \texttt{lm-evaluation-harness} and VLMs with \texttt{lmms-eval}, using benchmark-native evaluation wrappers and zero-shot prompting. When a CoT zero-shot benchmark variant is available, we use that variant. We otherwise retain the model-specific defaults and benchmark wrappers provided by the evaluation frameworks, and we do not add extra prompting intended to elicit particular behaviors. Exact run scripts, subset files, prompt templates, and scoring code will be released with the codebase.}
\label{tab:generation_settings}

\textbf{(a) LLMs}

\begin{tabular}{@{}lccccc@{}}
\toprule
Model & Temp. & Top-$p$ & Top-$k$ & Min-$p$ & Max gen toks \\
\midrule
Qwen3                         & 0.6 & 0.95 & 20 & 0  & 16,384 \\
Qwen2.5-Instruct             & 0.7 & 0.8  & 20 & 0  & 16,384 \\
OLMo-3-7B-Think              & 0.6 & 0.95 & 20 & 0  & 32,768 \\
OLMo-3-7B-Instruct           & 0.6 & 0.95 & 20 & 0  & 32,768 \\
Nemotron-v2                  & 0.6 & 0.95 & 20 & 0  & 8,192 \\
Nemotron-v2-Base                & 0.0 & --   & -- & -- & 8,192 \\
DeepSeek-R1-Distill-Qwen-7B  & 0.6 & 0.95 & 20 & 0  & 32,768 \\
\bottomrule
\end{tabular}

\vspace{0.8em}

\textbf{(b) VLMs}

\begin{tabular}{@{}lccccc@{}}
\toprule
Model & Temp. & Top-$p$ & Top-$k$ & Min-$p$ & Max gen toks \\
\midrule
Qwen3-VL-8B-Thinking     & 1.0 & 0.95 & 20 & -- & 40,960 \\
Qwen3-VL-8B-Instruct     & 0.7 & 0.8  & 20 & -- & 16,384 \\
GLM-4.1V-9B-Thinking     & 1.0 & 0.95 & 20 & -- & 40,960 \\
Kimi-VL-A3B-Thinking     & 0.8 & 0.8  & 20 & 0  & 16,384 \\
Kimi-VL-A3B-Instruct     & 0.2 & 0.8  & 20 & 0  & 16,384 \\
InternVL3.5-8B           & 0.6 & 0.8  & 20 & 0  & 16,384 \\
InternVL3.5-8B-Instruct  & 0.8 & 0.8  & 20 & 0  & 16,384 \\
\bottomrule
\end{tabular}
\end{table*}

\section{Qualitative samples and surface markers of annotated behaviors}

We include behavior-positive traces to make the annotation labels auditable in concrete model outputs. The examples show distinct surface forms, with overlap at label boundaries. In particular, uncertainty acknowledgment is often marked by explicit hedging and self-doubt language such as ``I’m not sure'' or ``I’m confused,'' while self-awareness is more often expressed as an information audit, for example when the model states that the passage does not mention or does not specify the information needed to justify a conclusion. Self-correction tends to appear as an explicit break in the reasoning path, with phrases like ``wait,'' ``actually,'' or ``let’s start over,'' and hypothesis testing is most clearly marked by branching language such as ``alternatively,'' ``suppose,'' or ``case 1 / case 2.''

Confidence calibration is less tied to any single phrase. Instead, it often appears as a confidence arc in which the model becomes more or less certain as the reasoning weakens or stabilizes. Knowledge alignment is often visible when the model converts the problem into the appropriate domain framework rather than relying on generic technical language, and goal tracking appears most clearly when the model explicitly states what has been established and what remains to be solved. Table~\ref{tab:surface_markers} summarizes the most common recurring markers and qualitative patterns for the annotated behaviors. Pages~\pageref{card:first}--\pageref{card:last} provide selected qualitative samples for the behaviors central to our findings.

\section{Robustness to annotation noise and question difficulty}
\label{sec:robustness}

\paragraph{Within-question analysis.}
To reduce confounding from question difficulty, we compare traces from the same model on the same question. For each MATH-500 problem, we generate 8 traces at temperature 0.6 and compute the per-question accuracy difference between traces where a behavior is present versus absent. The same broad ranking appears across Qwen3-4B-Think, Qwen3-4B-Instruct, and Qwen3-4B-VL on MathVista: confidence calibration remains strongly positive, self-correction is positive but generally smaller, hypothesis testing is weak, and uncertainty acknowledgment is null or negative. The core ranking therefore does not appear to be driven only by easier questions eliciting different behaviors. These comparisons control for question difficulty. They do not control for the latent quality of the reasoning path: a trace already on a productive path may both express calibrated confidence and reach the correct answer. We therefore read the within-question results as constraining a difficulty-based explanation, not as resolving causal direction.

\paragraph{Sensitivity to random annotation noise.}
We also test whether the Behavioral Lift ranking is fragile to random annotation errors. For each behavior, we flip 5\%, 10\%, 15\%, and 20\% of labels at random and recompute Lift over 1000 trials. The main ranking is stable under this perturbation. For LLMs, confidence calibration and knowledge alignment remain the top behaviors, while uncertainty acknowledgment remains the lowest-ranked behavior in every trial. For VLMs, confidence calibration remains the top-ranked behavior and uncertainty acknowledgment remains the lowest-ranked behavior in every trial, with the rest of the ordering also largely unchanged. The only unstable behavior is hypothesis testing, whose clean Lift is already near zero.

\paragraph{Lucky-guess analysis.}
To test whether confidence calibration is simply a downstream signal of correctness, we compare its prevalence on lucky guesses, where the final answer is correct despite flawed reasoning, versus non-lucky responses. In both LLMs and VLMs, and for both thinking and instruct models, calibration is far less common on lucky guesses than on non-lucky responses. Calibration therefore tracks sound reasoning more closely than final-answer correctness alone.

\paragraph{Temporal position of behaviors.}
We analyze when each behavior first appears relative to answer commitment~\citep{chang2026prism} across sampled thinking-model traces. All four behaviors typically appear before the answer (91--99\% of traces), with mean behavior positions of 27--35\% through the trace versus 79--94\% for the answer. This pattern is shared across behaviors and reflects the general structure of reasoning traces rather than a property unique to calibration.

\section{Length-Controlled Prevalence Analysis}
\label{app:length_control}

One concern is that thinking-oriented traces are longer, giving more opportunities for behaviors such as self-correction, hypothesis testing, and uncertainty acknowledgment to appear. To test whether length alone explains the amplification pattern, we bin responses into shared word-count quintiles within each modality and recompute thinking-minus-comparison prevalence gaps inside each bin. Length accounts for part of the pattern, especially among the shortest traces, but does not eliminate it: for LLMs, self-correction, hypothesis testing, and uncertainty acknowledgment remain more prevalent in thinking-oriented traces in each of the four non-shortest bins. For VLMs, self-correction remains amplified in 4/5 bins, uncertainty acknowledgment in 5/5, and hypothesis testing in 3/5. Under this trace-level definition of amplification, response length alone does not explain the main prevalence pattern.


\section{Linear Probing of Behavioral Representations}
\label{app:probing}

As a supporting validation, we train linear probes on hidden states from
Qwen3-4B-Thinking and Qwen3-4B-Instruct to test whether the annotated
behaviors correspond to internally decodable structure rather than arbitrary
surface labels. For each behavior, we perform teacher-forced replay of the
model's saved traces, extract hidden states from the generated portion only
(layer 35 of 36), and train logistic regression probes with group-aware
cross-validation splitting by question to prevent leakage across repeated
traces.

\autoref{tab:probe_auc} shows that several annotated behaviors are linearly decodable from Qwen hidden states.
Two patterns are notable. First, in the thinking model, probe decodability is
directionally aligned with Behavioral Lift and is most strongly aligned on
incorrect-only traces, where the rank correlation reaches $\rho = +0.86$
($p = 0.014$). Among incorrect thinking traces, knowledge alignment and goal
tracking are among the most linearly separable evaluated behaviors, while
hypothesis testing and uncertainty acknowledgment are among the least
separable. Second, in both models, confidence calibration remains strongly decodable on correct-only traces,
indicating that it is not a trivial proxy for final correctness.

The instruct model shows a more mixed pattern. On incorrect traces, the rank
correlation is negative ($\rho = -0.71$, $p = 0.071$), with hypothesis testing
and uncertainty acknowledgment among the most decodable behaviors. This is
consistent with these behaviors being more distinctive in instruct traces,
especially when they are relatively rare. By contrast, confidence calibration
is nearly absent from incorrect traces in both models ($1.7\%$ thinking,
$0.7\%$ instruct), preventing reliable probe evaluation in that subset.

The probing results are consistent with the annotation labels: several behaviors, including confidence calibration and knowledge alignment, are decodable from hidden states, and in the thinking model this decodability is directionally aligned with Behavioral Lift, especially on incorrect traces.

\begin{table}[h]
\centering
\caption{Linear probe AUC for higher-order behaviors in Qwen3-4B-Thinking and
Qwen3-4B-Instruct hidden states. Probes use layer 35 representations from the
generated portion of teacher-forced replays, with the best AUC over five trace
positions (25\%, 50\%, 75\%, last token, mean pool) reported for each behavior.
Dashes indicate insufficient minority-class samples ($<30$). Behaviors are
sorted by Behavioral Lift rank from \autoref{fig:lift}.
$^\dagger$~marks behaviors amplified at least $3\times$ by thinking training.}
\label{tab:probe_auc}
\small
\begin{tabular}{l c ccc ccc}
\toprule
& & \multicolumn{3}{c}{\textbf{Thinking}} & \multicolumn{3}{c}{\textbf{Instruct}} \\
\cmidrule(lr){3-5} \cmidrule(lr){6-8}
\textbf{Behavior} & \textbf{Lift} & All & Correct & Incorrect & All & Correct & Incorrect \\
\midrule
Confidence calibration     & 1 & .775 & .771 & --   & .849 & .889 & --   \\
Self-awareness             & 2 & .672 & .628 & --   & .609 & .599 & --   \\
Knowledge alignment        & 3 & .798 & .795 & .704 & .807 & .886 & .624 \\
Evidence citation          & 4 & .798 & .802 & .651 & .821 & .832 & .652 \\
Goal tracking              & 5 & .793 & .744 & .690 & .800 & .774 & .645 \\
Planning present           & 6 & .684 & .664 & .608 & .704 & .695 & .605 \\
\midrule
Self-correction$^\dagger$           & 7 & .727 & .755 & .582 & .807 & .842 & .693 \\
Hypothesis testing$^\dagger$        & 8 & .641 & .655 & .561 & .750 & .748 & .759 \\
Uncertainty ack.$^\dagger$ & 9 & .709 & .723 & .598 & .873 & .844 & .731 \\
\midrule
Spearman $\rho$            &   & +.40 & +.32 & \textbf{+.86}$^*$ & $-.12$ & +.15 & $-.71$ \\
$N$                        &   & 1989 & 1751 & 238  & 1999 & 1694 & 305  \\
\bottomrule
\end{tabular}

\vspace{2pt}
{\footnotesize $^*$~$p = 0.014$. All other $\rho$ values have $p > 0.05$.}
\end{table}

\section{Prompting Behavioral Lift at Inference Time}
\label{prompting_test}

To test whether the Behavioral Lift ranking points to behaviors worth trying at
inference time, we prompt Qwen3-4B-Thinking on three benchmarks: MATH-500
(500 questions), MMLU-Pro Math (550 questions), and GPQA (448 questions).
We compare three conditions under identical decoding settings:

\begin{itemize}
\item \textbf{High-Lift prompt}: encourages confidence calibration, knowledge alignment, and self-awareness. These are the top-ranked Lift behaviors and are not strongly amplified in thinking traces. The prompt instructs the model to let confidence
track reasoning strength, identify the applicable domain framework, and
recognize when information is insufficient.

\item \textbf{Low-Lift prompt}: encourages pervasive uncertainty
acknowledgment, the most amplified behavior with consistently negative
Lift. The prompt instructs the model to express doubt at every step, note
confusion before continuing, and add caveats even when fairly confident.

\item \textbf{Baseline}: no behavioral prompt.
\end{itemize}

High-Lift prompting improves accuracy on MATH-500 and MMLU-Pro and remains best on GPQA, while Low-Lift uncertainty prompting degrades accuracy on every benchmark (Table~\ref{tab:prompt_intervention}).
The ordering High-Lift $>$ Baseline $>$ Low-Lift holds throughout.
The effect scales with baseline model competence: on MATH-500 (baseline
84.8\%), the high-Lift prompt improves accuracy by 5.8pp ($p < 0.001$) and
recovery from 57.1\% to 68.9\%, while reducing logical failure, context
misread, and knowledge gap rates. On MMLU-Pro Math (baseline 58.4\%), the
gap between conditions reaches 15.7pp ($p < 0.001$), with the low-Lift prompt
degrading accuracy by 12.1pp ($p < 0.001$). On GPQA, where the 4B model
operates near chance (baseline 49.1\%), the high-Lift prompt produces only a
marginal gain (+0.7pp, n.s.), but the low-Lift prompt still degrades accuracy
($-$3.5pp, $p = 0.14$), with the high-vs-low gap approaching significance
($p = 0.086$).

The prompts shifted behaviors as intended. Across benchmarks, the high-Lift
prompt increased confidence calibration (+0.4 to +4.4pp), knowledge alignment
(+3.8 to +8.0pp), and reduced uncertainty acknowledgment ($-$18.1 to
$-$21.2pp). The low-Lift prompt drove uncertainty acknowledgment to near
saturation (94.4--99.3\%), while reducing confidence calibration by up to
24.3pp.

The prompting results are consistent with the Behavioral Lift ranking: encouraging high-lift behaviors improved accuracy in two of three benchmarks, whereas pervasive uncertainty prompting reduced accuracy on all three. The effect is clearest on tasks dominated by executable reasoning chains and fades as baseline competence drops, consistent with the
task-dependent recovery mechanisms identified in our Recovery Rate analysis.

This experiment is preliminary. The prompts change several properties of a trace at once, so the results do not isolate the effect of any single behavior, and the GPQA differences are not statistically significant. We report it as evidence that the Lift ranking carries usable signal at generation time. Testing whether these behaviors improve reasoning directly calls for controlled training interventions, which we see as the natural next step.

\begin{table*}[t]
\centering
\caption{Prompting results on Qwen3-4B-Thinking across three benchmarks. High-Lift encourages confidence calibration, knowledge alignment,
and self-awareness. Low-Lift encourages pervasive uncertainty acknowledgment.
Deltas are relative to baseline; significance from paired bootstrap
(10,000 resamples, matched by question).}
\label{tab:prompt_intervention}
\footnotesize
\setlength{\tabcolsep}{2pt}
\resizebox{\textwidth}{!}{%
\begin{tabular}{@{}lccccccccc@{}}
\toprule
& \multicolumn{3}{c}{\textbf{MATH-500}} & \multicolumn{3}{c}{\textbf{MMLU-Pro Math}} & \multicolumn{3}{c}{\textbf{GPQA}} \\
\cmidrule(lr){2-4} \cmidrule(lr){5-7} \cmidrule(lr){8-10}
& Hi & Base & Lo & Hi & Base & Lo & Hi & Base & Lo \\
\midrule
\multicolumn{10}{@{}l}{\textit{Accuracy \& Recovery (\%)}} \\
Accuracy & \textbf{90.6} & 84.8 & 80.2 & \textbf{62.0} & 58.4 & 46.3 & \textbf{49.8} & 49.1 & 45.6 \\
\quad $\Delta$ & +5.8$^{***}$ & -- & $-$4.6$^{*}$ & +3.6$^{*}$ & -- & $-$12.1$^{***}$ & +0.7 & -- & $-$3.5 \\
\quad 95\% CI & [+2.6,+9.0] & & [$-$9.0,$-$0.2] & [+0.5,+6.7] & & [$-$16.4,$-$7.7] & [$-$3.8,+5.1] & & [$-$8.3,+0.9] \\
Recovery & 68.9 & 57.1 & 46.2 & 20.2 & 18.5 & 13.4 & 34.8 & 36.1 & 28.2 \\
\midrule
\multicolumn{10}{@{}l}{\textit{Targeted behavior prevalence (\%)}} \\
Conf.\ calib. & 62.0 & 58.6 & 34.3 & 49.1 & 44.7 & 21.7 & 16.1 & 15.6 & 23.3 \\
Knowl.\ align. & 88.8 & 80.8 & 78.6 & 75.3 & 71.5 & 71.2 & 39.5 & 33.7 & 35.3 \\
Self-awareness & 45.6 & 49.0 & 55.5 & 46.2 & 40.5 & 36.8 & 23.2 & 21.4 & 28.9 \\
Uncert.\ ack. & 33.8 & 55.0 & 97.6 & 62.2 & 60.2 & 94.4 & 55.6 & 73.7 & 99.3 \\
\midrule
\multicolumn{10}{@{}l}{\textit{Failure rates (\%)}} \\
Logical fail. & 25.2 & 31.0 & 29.3 & 43.5 & 46.2 & 50.6 & 75.2 & 77.2 & 73.6 \\
Post-hoc rat. & 25.0 & 29.2 & 26.1 & 35.3 & 38.7 & 43.5 & 72.5 & 75.9 & 68.5 \\
Context misr. & 11.8 & 18.0 & 20.0 & 19.6 & 24.5 & 24.2 & 59.6 & 66.3 & 61.7 \\
\bottomrule
\end{tabular}%
}

\vspace{2pt}
\parbox{\textwidth}{\footnotesize
Hi = High-Lift prompt; Lo = Low-Lift prompt; Base = no behavioral prompt.
$^{***}p<0.001$, $^{**}p<0.01$, $^{*}p<0.05$ (paired bootstrap, 10,000 resamples).
95\% CIs are for the accuracy delta vs.\ baseline. High-Lift vs.\ Low-Lift gap:
MATH-500 +10.4pp ($p<0.001$), MMLU-Pro +15.7pp ($p<0.001$), GPQA +4.3pp ($p=0.086$).
}
\end{table*}

\clearpage
\section{Visual Claim Accuracy Matters More Than Image References}
\label{app:vision_grounding}

For VLM traces, we additionally annotate two modality-specific grounding labels:
\textit{visual references present}, which records whether the trace explicitly mentions
concrete image content, and \textit{visual claims accurate}, which records whether those
visual statements are factually correct. These labels separate image mention from correct image use.

The distinction is sharp. Across all VLM traces, accurate visual claims have
\textbf{+60.9} Behavioral Lift: responses with accurate visual claims are correct
87.8\% of the time, compared with 26.9\% when their visual claims are inaccurate.
By contrast, simply mentioning visual content has essentially no positive association
with correctness (\textbf{-3.3} Lift). The same pattern holds within each benchmark:
visual claim accuracy is strongly positive on VisualPuzzles (+71.7), MathVista (+54.3),
and MMMU (+45.1), whereas visual reference presence is near zero except for a modest
positive effect on MathVista (+12.1). This supports the taxonomy design choice to
distinguish behavioral \emph{presence} from behavioral \emph{quality}.

Thinking models also shift the profile of visual failures. They show less
\textit{visual neglect} than instruct models overall (36.0\% vs.\ 47.4\%), with especially
large reductions on MathVista and MMMU, but they exhibit more
\textit{visual hallucination} (24.7\% vs.\ 18.0\%). Longer reasoning traces appear to attend to the image more actively while also creating more opportunities to state incorrect visual claims. Recovery from visual failures is asymmetric:
thinking models recover better from hallucination (20.8\% vs.\ 12.1\%), but worse from
visual neglect (15.3\% vs.\ 20.4\%). In other words, neglect is often avoided upstream
rather than repaired downstream.

Confidence calibration remains strongly associated with correctness even after conditioning
on visual grounding quality. Among traces with accurate visual claims, calibration has
+43.9 Lift; among traces with inaccurate visual claims, its Lift rises to +73.5. Thus, visual
grounding quality and metacognitive quality capture different aspects of successful VLM
reasoning: accurate perception is highly predictive overall, while calibration is especially
informative when the trace contains imperfect visual interpretation.

\begin{figure*}[t]
  \centering
  \includegraphics[width=\textwidth]{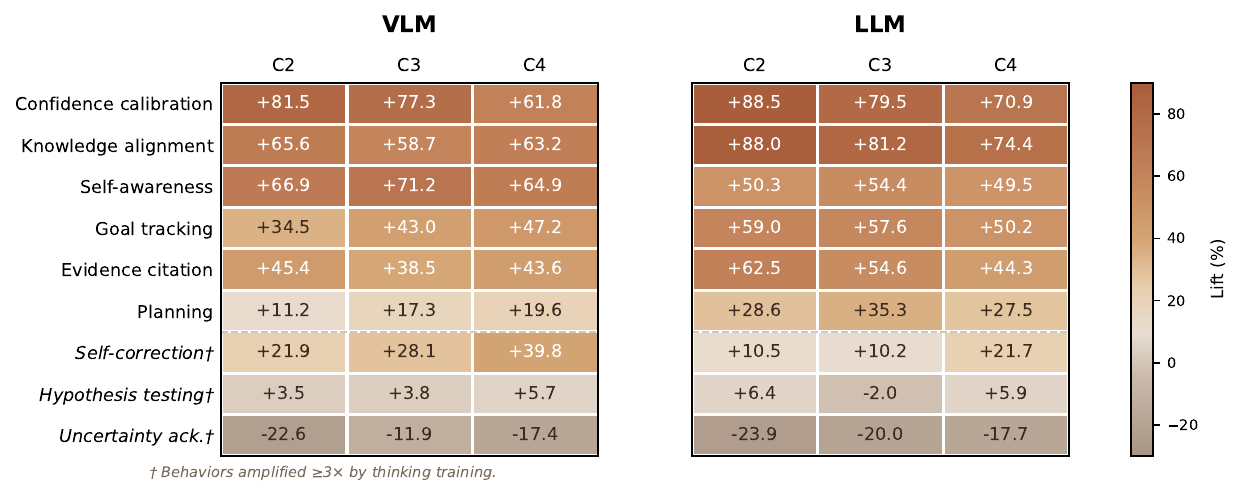}
  \caption{Behavioral Lift (\%) within response complexity bins (C2--C4), pooled across all models and benchmarks per modality. The ranking from Figure~\ref{fig:lift} holds within each bin: confidence calibration remains among the strongest positive associations with correctness and uncertainty acknowledgment shows negative Lift in every bin in both modalities, confirming that the disconnect between amplification and Behavioral Lift is not an artifact of response length or problem difficulty. Word-count quintile analysis in~\autoref{fig:wordcount} yields consistent results.}
  \label{fig:length_control}
\end{figure*}

\begin{figure}[t]
  \centering
  \includegraphics[width=\textwidth]{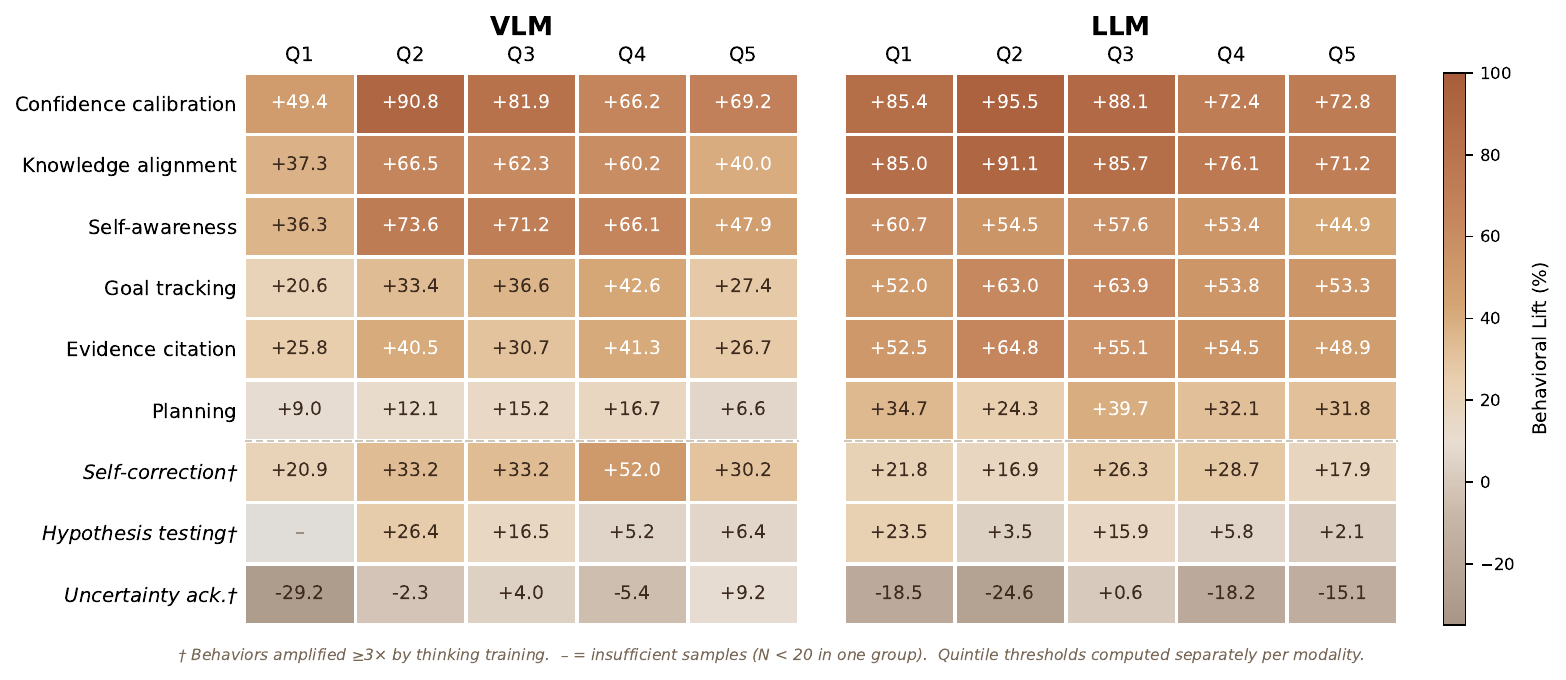}
    \caption{Behavioral Lift (\%) within word-count quintiles, pooled across all models and benchmarks per modality (VLM: $N{=}7{,}000$; LLM: $N{=}8{,}282$). Quintile thresholds are computed separately per modality. Dashes indicate insufficient samples ($N{<}20$ in one group). The top-level pattern from \autoref{fig:lift} holds: confidence calibration and knowledge alignment dominate, while the behaviors identified as amplified in the prevalence analysis (\dag) cluster lower. This complements \autoref{fig:length_control} using a purely mechanical binning strategy with no annotator involvement. Appendix~\ref{app:length_control} shows that the main prevalence-amplification pattern persists within shared word-count bins, so it is not explained by response length alone.}
  \label{fig:wordcount}
\end{figure}

\begin{figure*}[t]
\centering
\includegraphics[width=\textwidth]{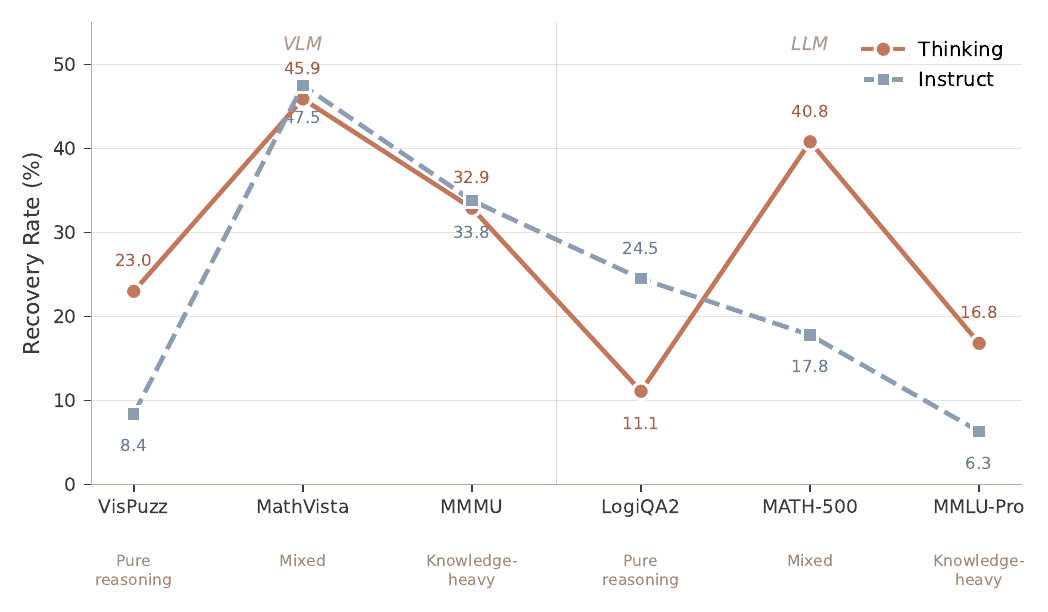}
\caption{Recovery Rate across the task spectrum. Thinking models recover from detected failures at 2.3--2.7$\times$ the rate of instruct models on benchmarks that reward step-by-step computation (VisualPuzzles, MATH-500, MMLU-Pro). On LogiQA2, instruct models recover better. Recovery advantage tracks task type, not modality.}
\label{fig:recovery}
\end{figure*}

\begin{figure}[t]
\centering
\includegraphics[width=\columnwidth]{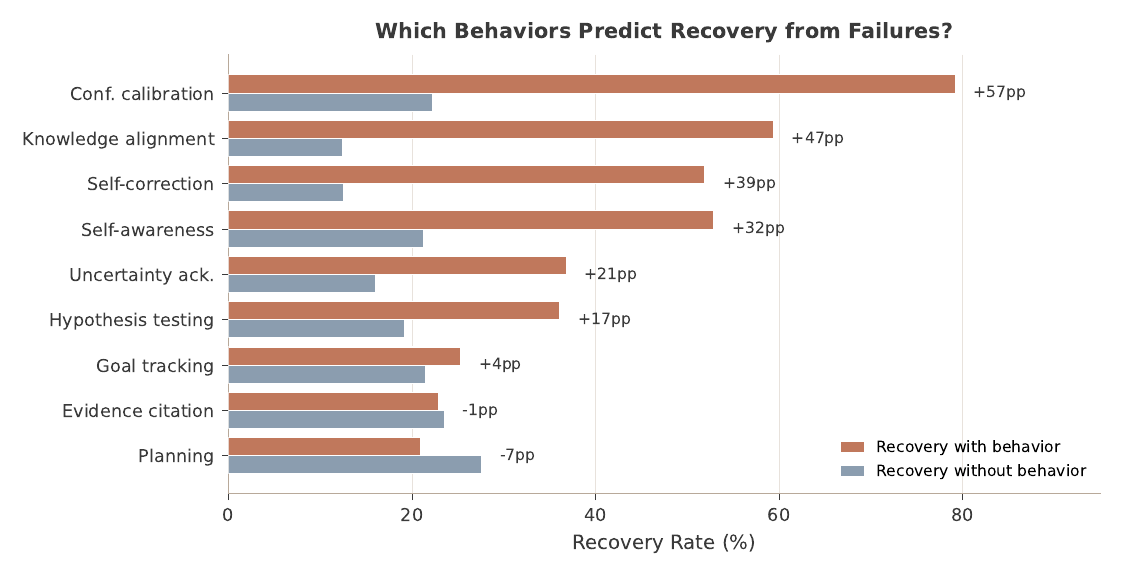}
\caption{Recovery rate conditional on behavior presence among traces with at least one detected failure ($N{=}4{,}778$). Confidence calibration, knowledge alignment, and self-correction are the strongest positive predictors of recovery. This links the Behavioral Lift and Recovery Rate analyses: self-correction has modest overall Lift but strongly predicts recovery from failures.}
\label{fig:recovery_conditional}
\end{figure}

\begin{figure*}[t]
\centering
\includegraphics[width=\textwidth]{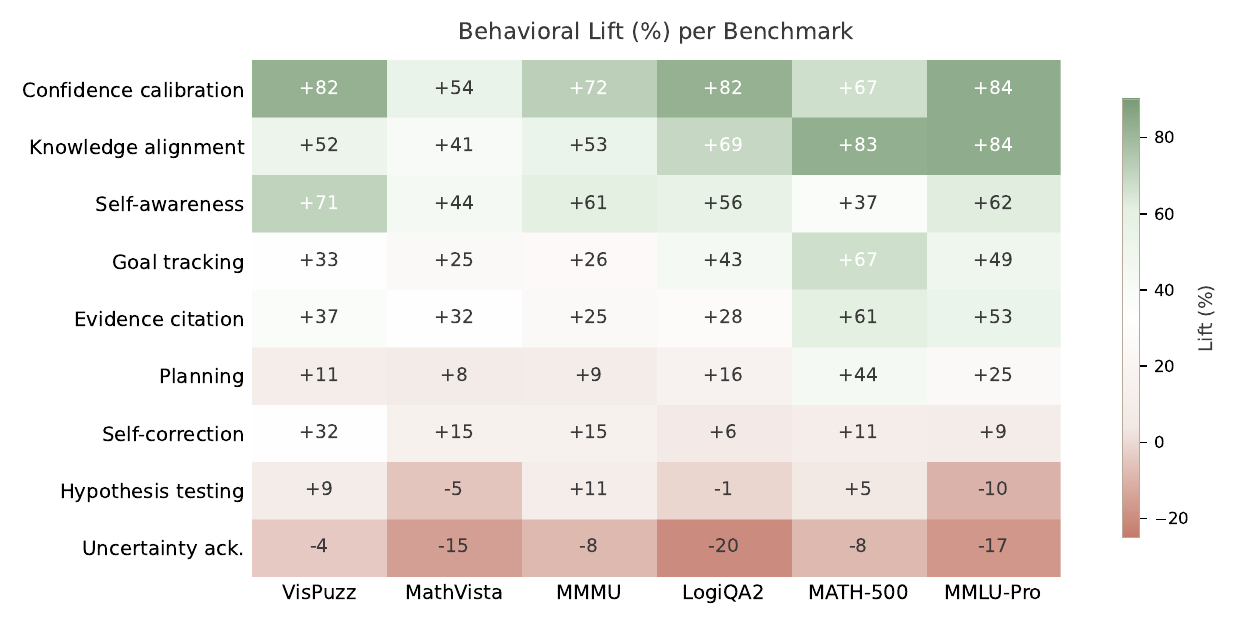}
\caption{Behavioral Lift (\%) by benchmark for VLM and LLM tasks. Each cell computes Lift within a single benchmark. The broad ranking is stable: confidence calibration and knowledge alignment show the strongest positive associations with correctness, while the behaviors identified as amplified in the prevalence analysis cluster near the bottom.}
\label{fig:lift_heatmap}
\end{figure*}

\begin{figure*}[t]
\centering
\includegraphics[width=\textwidth]{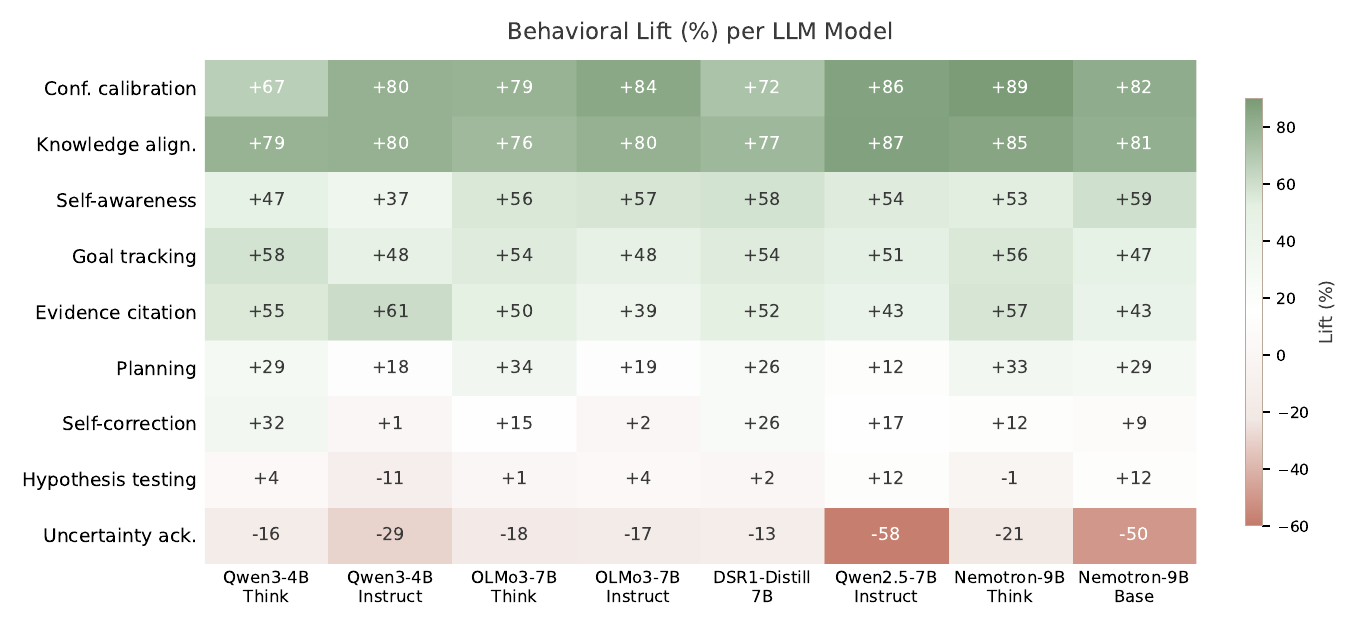}
\caption{Behavioral Lift (\%) by model for all LLMs. Each column computes Lift from a single model's responses. The Behavioral Lift ordering holds across models: confidence calibration and knowledge alignment typically show the strongest positive associations with correctness, while the behaviors identified as amplified in the prevalence analysis tend to cluster lower.}
\label{fig:lift_per_llm}
\end{figure*}

\begin{figure*}[t]
\centering
\includegraphics[width=\textwidth]{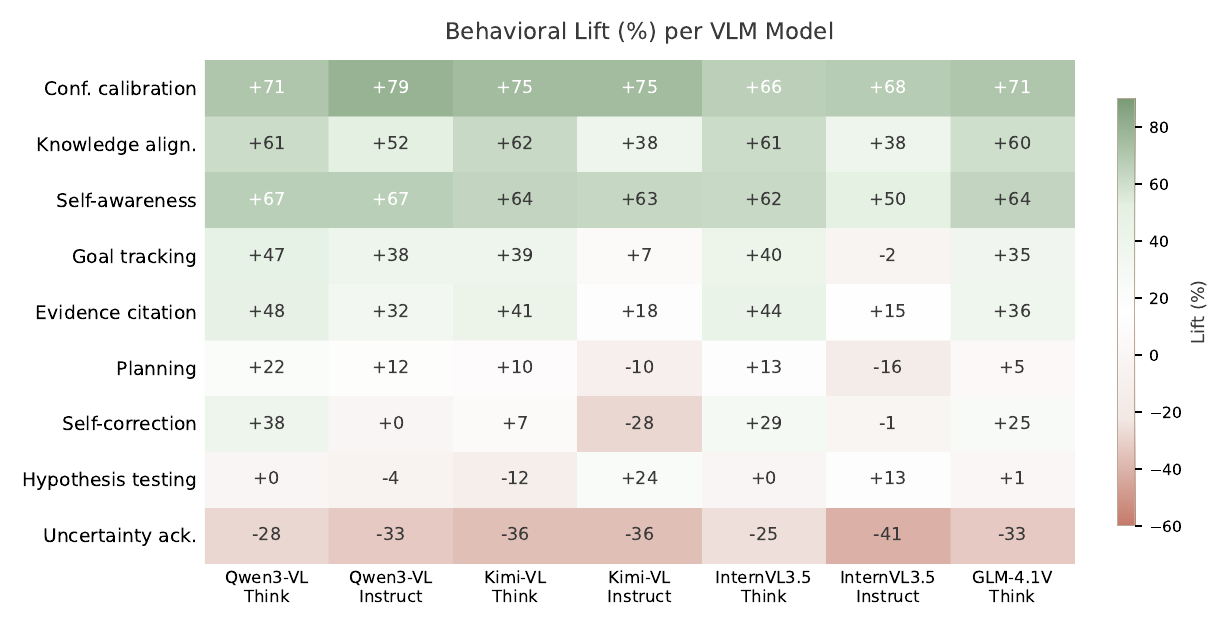}
\caption{Behavioral Lift (\%) by model for all VLMs. Each column computes Lift from a single model's responses. The Behavioral Lift ordering holds across models: confidence calibration and knowledge alignment typically show the strongest positive associations with correctness, while the behaviors identified as amplified in the prevalence analysis tend to cluster lower.}
\label{fig:lift_per_vlm}
\end{figure*}

\begin{figure}[t]
\centering
\includegraphics[width=\columnwidth]{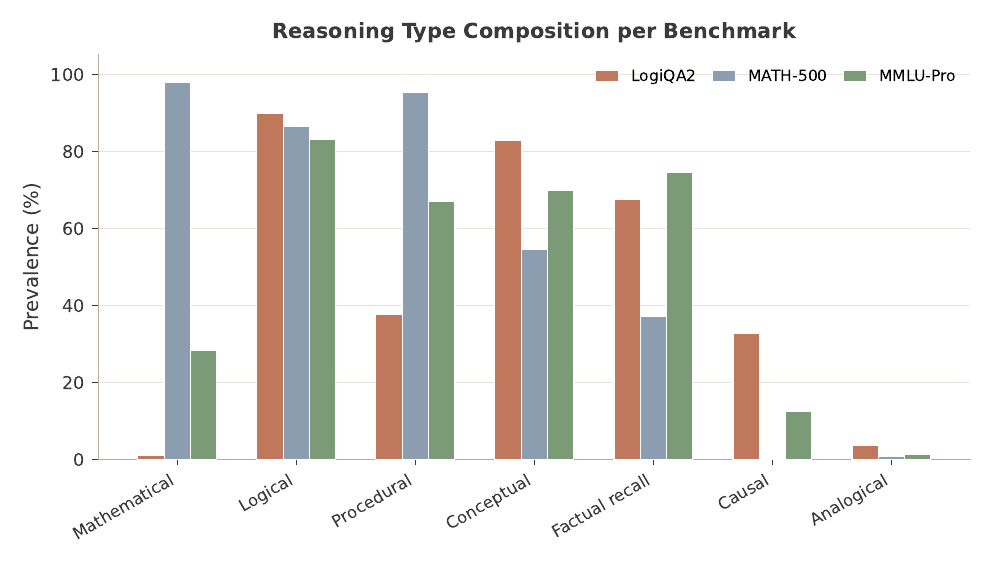}
\caption{Reasoning type composition per benchmark. Each benchmark emphasizes different reasoning demands: MATH-500 is dominated by mathematical and procedural reasoning, LogiQA2 by logical and conceptual reasoning. This variation underlies the task-dependent recovery mechanisms observed across benchmarks.}
\label{fig:reasoning_types}
\end{figure}

\begin{figure}[t]
\centering
\includegraphics[width=\columnwidth]{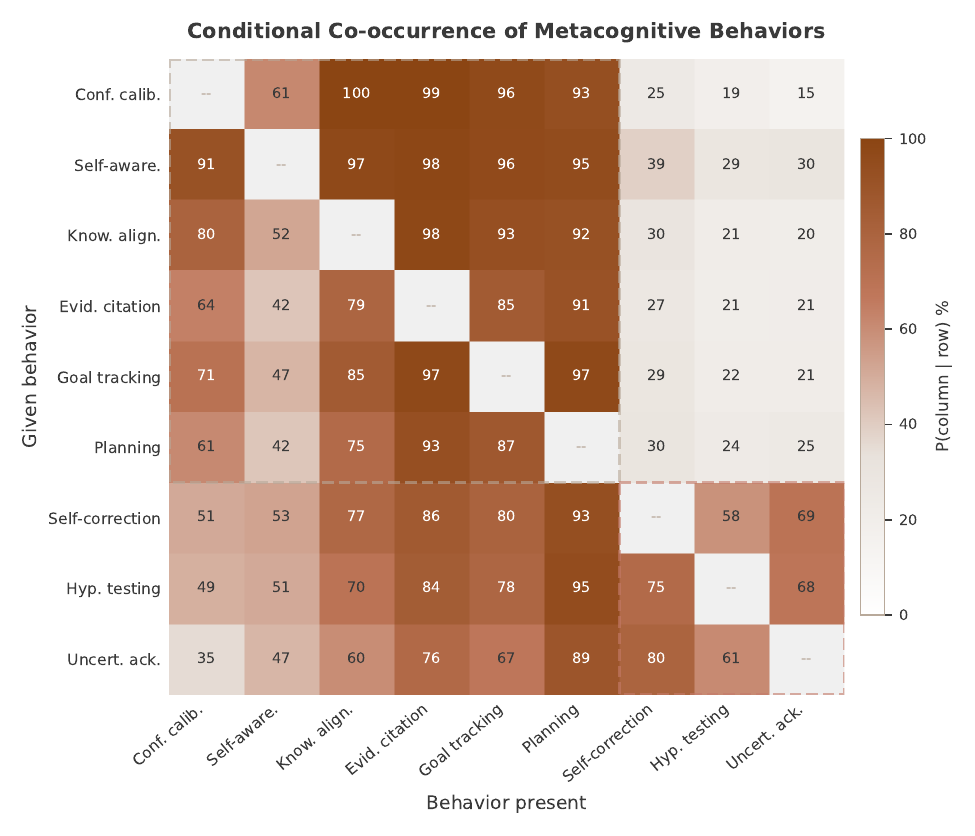}
\caption{Conditional co-occurrence of cross-modal higher-order behaviors: $P(\text{column behavior present} \mid \text{row behavior present})$. Behaviors in the top-left dashed block co-occur at high rates (60--100\%), while the three behaviors most amplified by thinking training in the bottom-right dashed block also co-occur (58--80\%) but show lower cross-block co-occurrence with the others (15--30\%). This suggests the amplified behaviors form a relatively distinct cluster.}
\label{fig:cooccurrence}
\end{figure}

\begin{figure}[t]
\centering
\includegraphics[width=\columnwidth]{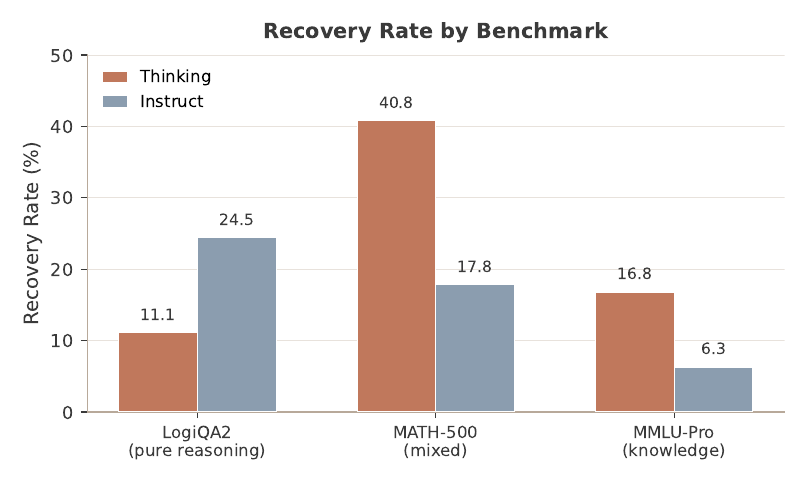}
\caption{Recovery Rate by benchmark for LLM models. Thinking models recover at 2--3$\times$ the rate of instruct models on MATH-500 and MMLU-Pro, but instruct models recover better on LogiQA2. The pattern tracks task structure: extended reasoning tasks favor thinking models, while pattern-matching tasks do not.}
\label{fig:recovery_benchmark}
\end{figure}

\begin{figure*}[t]
\centering
\includegraphics[width=\textwidth]{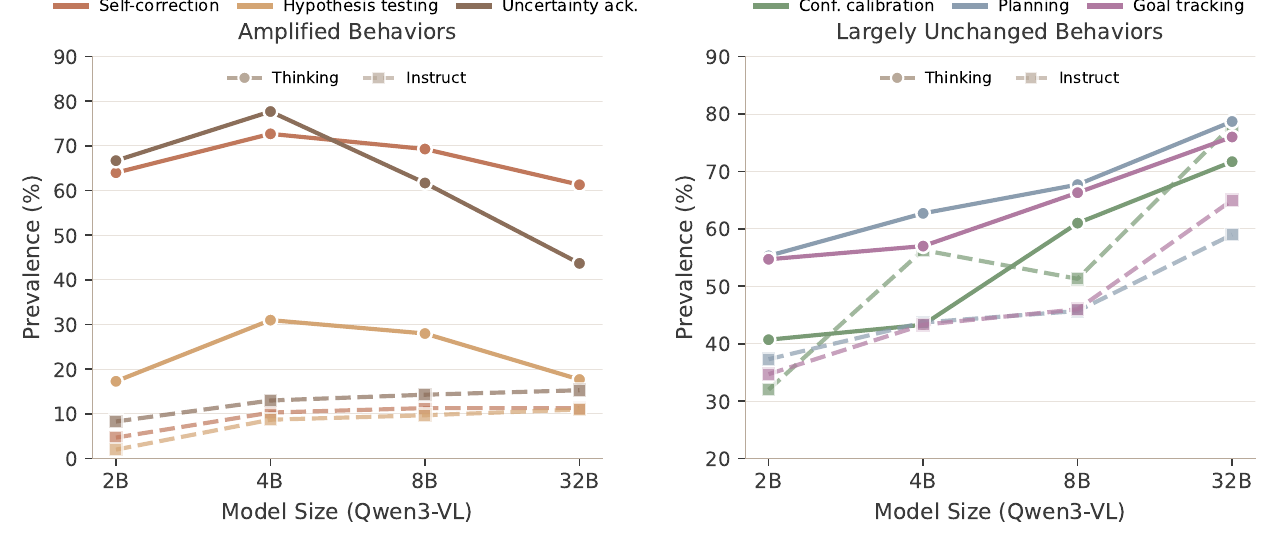}
\caption{Behavioral prevalence across model scales for Qwen3-VL on MathVista. The gap in amplified behaviors (left) persists from 2B to 32B parameters. Largely unchanged behaviors (right) converge as model size increases, with instruct models matching or exceeding thinking models at 32B.}
\label{fig:scaling_analysis}
\end{figure*}

\label{sec:scaling}
\begin{figure*}[t]
\centering
\includegraphics[width=\textwidth]{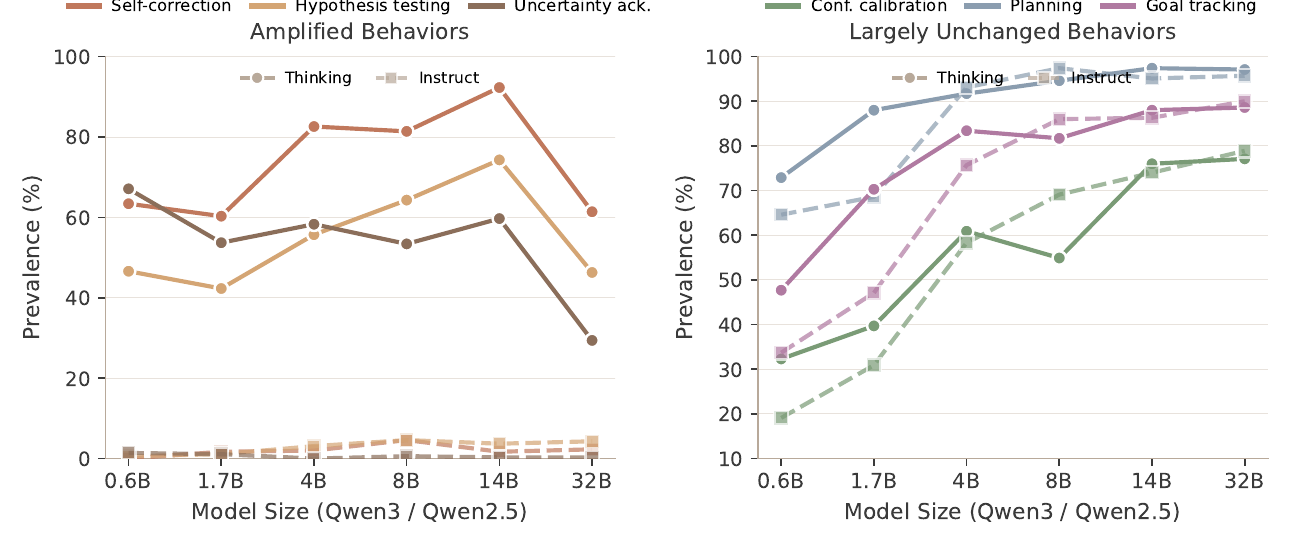}
\caption{Behavioral prevalence across model scales for Qwen3 on MATH-500. The gap in the three behaviors most amplified by thinking training persists from 0.6B to 32B parameters, while the largely unchanged behaviors converge as model size increases, with instruct models matching or exceeding thinking models at 32B.}
\label{fig:llm_scaling}
\end{figure*}

\begin{table*}[t]
\centering
\tiny
\setlength{\tabcolsep}{4pt}
\begin{tabular}{@{}llllp{0.35\linewidth}@{}}
\toprule
Model & Variant role & Family & Params & Checkpoint \\
\midrule
\multicolumn{5}{@{}l}{\textit{Vision-Language Models}} \\[2pt]
\quad Qwen3-VL-8B-Thinking
  & Think-oriented & Qwen & 8B
  & \texttt{Qwen/Qwen3-VL-8B-Thinking} \\
\quad Qwen3-VL-8B-Instruct
  & Non-thinking & Qwen & 8B
  & \texttt{Qwen/Qwen3-VL-8B-Instruct} \\
\quad Kimi-VL-A3B-Thinking
  & Think-oriented & Kimi & 3B$^{*}$
  & \texttt{moonshotai/Kimi-VL-A3B-Thinking-2506} \\
\quad Kimi-VL-A3B-Instruct
  & Non-thinking & Kimi & 3B$^{*}$
  & \texttt{moonshotai/Kimi-VL-A3B-Instruct} \\
\quad InternVL3.5-8B
  & Think-oriented & InternVL & 8B
  & \texttt{OpenGVLab/InternVL3\_5-8B} \\
\quad InternVL3.5-8B-Instruct
  & Non-thinking & InternVL & 8B
  & \texttt{OpenGVLab/InternVL3\_5-8B-Instruct} \\
\quad GLM-4.1V-9B-Thinking
  & think-oriented & GLM & 9B
  & \texttt{zai-org/GLM-4.1V-9B-Thinking} \\
\addlinespace[4pt]
\multicolumn{5}{@{}l}{\textit{Text-Only Language Models}} \\[2pt]
\quad Qwen3-4B-Thinking
  & Think-oriented & Qwen & 4B
  & \texttt{Qwen/Qwen3-4B-Thinking-2507} \\
\quad Qwen3-4B-Instruct
  & Non-thinking & Qwen & 4B
  & \texttt{Qwen/Qwen3-4B-Instruct-2507} \\
\quad OLMo-3-7B-Think
  & Think-oriented & OLMo & 7B
  & \texttt{allenai/Olmo-3-7B-Think} \\
\quad OLMo-3-7B-Instruct
  & Non-thinking & OLMo & 7B
  & \texttt{allenai/Olmo-3-7B-Instruct} \\
\quad DeepSeek-R1-Distill-Qwen-7B
  & think-oriented & DeepSeek/Qwen & 7B
  & \texttt{deepseek-ai/DeepSeek-R1-Distill-Qwen-7B} \\
\quad Qwen2.5-7B-Instruct
  & Additional comparison & Qwen & 7B
  & \texttt{Qwen/Qwen2.5-7B-Instruct} \\
\quad Nemotron-Nano-9B-v2
  & Reasoning variant & Nemotron & 9B
  & \texttt{nvidia/NVIDIA-Nemotron-Nano-9B-v2} \\
\quad Nemotron-Nano-9B-v2-Base
  & Non-thinking & Nemotron & 9B
  & \texttt{nvidia/NVIDIA-Nemotron-Nano-9B-v2-Base} \\
\bottomrule
\end{tabular}
\caption{%
Models evaluated.
The checkpoint column lists the exact public model identifiers used for inference; the variant-role column is descriptive and does not assert that any pair differs by a single isolated training stage.
Five released same-family, same-scale thinking/non-thinking endpoint pairs are Qwen3-VL, Kimi-VL, InternVL3.5, Qwen3-4B, and OLMo-3-7B.
GLM-4.1V and DeepSeek-R1-Distill-Qwen-7B are unmatched think-oriented auxiliary models; Qwen2.5-7B-Instruct is an additional comparison baseline.
Nemotron-Nano-9B-v2 and Nemotron-Nano-9B-v2-Base are included as a base-to-reasoning comparison rather than a clean instruct/thinking endpoint pair.
Analyses restricted to same-family thinking/non-thinking endpoint pairs are consistent with the overall results. $^{*}$Kimi-VL-A3B has 3B active parameters (MoE architecture).
}
\label{tab:models}
\end{table*}
\begin{table*}[!t]
\centering
\caption{%
Behavioral profile of thinking (Th) and instruct (In) models across six benchmarks.
Amplified behaviors (top) show large gaps between thinking and instruct models; non-amplified behaviors (middle) show small or reversed gaps.
Accuracy and recovery rate (bottom) show where behavior prevalence translates into task performance.
}
\label{tab:main_table_appendix}
\footnotesize
\setlength{\tabcolsep}{3.8pt}
\begin{tabular}{@{}l cc cc cc cc cc cc@{}}
\toprule
& \multicolumn{6}{c}{\textbf{VLM Benchmarks}} & \multicolumn{6}{c}{\textbf{LLM Benchmarks}} \\
\cmidrule(lr){2-7} \cmidrule(lr){8-13}
& \multicolumn{2}{c}{VisPuzz} & \multicolumn{2}{c}{MathVista} & \multicolumn{2}{c}{MMMU}
& \multicolumn{2}{c}{LogiQA2} & \multicolumn{2}{c}{MATH-500} & \multicolumn{2}{c}{MMLU-Pro} \\
\cmidrule(lr){2-3} \cmidrule(lr){4-5} \cmidrule(lr){6-7}
\cmidrule(lr){8-9} \cmidrule(lr){10-11} \cmidrule(lr){12-13}
& Th & In & Th & In & Th & In & Th & In & Th & In & Th & In \\
\midrule
\multicolumn{13}{@{}l}{\textit{Amplified behaviors (\%)}} \\[2pt]
\quad Self-correction      & 50.6 & 14.9 & 55.2 &  4.3 & 48.7 &  5.2 & 21.5 &  3.3 & 54.6 & 12.5 & 44.5 &  7.1 \\
\quad Hypothesis testing   & 42.6 & 17.6 & 21.6 &  3.7 & 51.5 & 11.3 & 25.6 &  6.4 & 42.8 & 12.8 & 34.4 &  8.2 \\
\quad Uncertainty ack.     & 85.1 & 27.5 & 51.6 &  7.2 & 68.9 & 10.6 & 25.0 &  3.9 & 43.5 &  8.1 & 52.8 & 10.2 \\
\addlinespace[4pt]
\multicolumn{13}{@{}l}{\textit{Largely unchanged behaviors (\%)}} \\[2pt]
\quad Self-awareness       & 27.1 & 21.6 & 57.3 & 30.6 & 47.0 & 27.0 & 33.1 & 26.0 & 52.6 & 36.8 & 37.8 & 34.1 \\
\quad Conf.\ calibration   & 18.3 & 24.5 & 58.8 & 42.3 & 43.9 & 35.1 & 47.4 & 46.5 & 66.1 & 67.9 & 40.9 & 46.7 \\
\quad Planning             & 63.0 & 67.0 & 63.7 & 38.6 & 74.5 & 39.7 & 74.3 & 50.2 & 91.2 & 92.3 & 86.9 & 83.8 \\
\quad Goal tracking        & 40.4 & 51.6 & 63.4 & 38.1 & 65.1 & 35.8 & 68.3 & 50.0 & 80.4 & 81.7 & 72.3 & 76.0 \\
\quad Knowledge align.     & 38.8 & 39.7 & 74.6 & 49.2 & 72.6 & 59.0 & 65.1 & 57.2 & 78.9 & 74.4 & 53.4 & 54.4 \\
\quad Evidence citation    & 42.0 & 55.0 & 73.3 & 47.8 & 56.4 & 48.8 & 87.0 & 73.1 & 86.1 & 91.2 & 71.6 & 80.3 \\
\midrule
Accuracy (\%)              & 36.5 & 29.7 & 77.5 & 66.6 & 59.2 & 52.4 & 54.1 & 58.4 & 81.4 & 74.3 & 51.4 & 51.5 \\
Recovery Rate (\%)         & 23.0 &  8.4 & 45.9 & 47.5 & 32.9 & 33.8 & 11.1 & 24.5 & 40.8 & 17.8 & 16.8 &  6.3 \\
\bottomrule
\end{tabular}
\vspace{4pt}
\end{table*}

\begin{table}[t]
\centering
\label{tab:lift}
\caption{%
Behavioral Lift for the nine cross-modal higher-order behaviors. Rows are grouped to show the amplified behaviors relative to the highest-Lift behaviors. Lift $=P(\text{correct} \mid b{=}\text{true}) - P(\text{correct} \mid b{=}\text{false})$.
Arrows (\markamplified) mark behaviors identified as amplified in the prevalence analysis
(${\geq}3\times$ thinking/instruct prevalence).
These behaviors cluster near the bottom of the Lift ranking.
}
\small
\setlength{\tabcolsep}{5pt}
\begin{tabular}{@{}l rr rr@{}}
\toprule
& \multicolumn{2}{c}{\textbf{VLM} ($N{=}7{,}000$)} & \multicolumn{2}{c}{\textbf{LLM} ($N{=}8{,}282$)} \\
\cmidrule(lr){2-3} \cmidrule(lr){4-5}
Behavior & Lift & Pres.\% & Lift & Pres.\% \\
\midrule
Conf.\ calibration         & $+$72.2 & 36.9 & $+$79.6 & 52.7 \\
Self-awareness              & $+$62.0 & 35.9 & $+$52.7 & 36.9 \\
Knowledge alignment         & $+$53.7 & 56.2 & $+$80.3 & 64.0 \\
Evidence citation           & $+$34.9 & 54.0 & $+$49.1 & 81.7 \\
Goal tracking               & $+$30.6 & 50.0 & $+$52.5 & 71.7 \\
Planning                    &  $+$6.6 & 59.4 & $+$26.3 & 80.1 \\
\addlinespace[3pt]
Self-correction\markamplified    & $+$20.1 & 32.9 & $+$12.4 & 24.2 \\
Hypothesis testing\markamplified &  $+$1.0 & 27.3 &  $+$1.0 & 21.9 \\
Uncertainty ack.\markamplified   & $-$16.1 & 46.3 & $-$13.9 & 24.2 \\
\bottomrule
\end{tabular}
\vspace{4pt}

\end{table}
\begin{table*}[t]
\centering
\small
\setlength{\tabcolsep}{4.2pt}
\caption{Conditional prevalence of cross-modal higher-order behaviors in correct versus incorrect LLM traces. Values report $P(b\mid\checkmark)$, $P(b\mid\times)$, and $\Delta=P(b\mid\checkmark)-P(b\mid\times)$. Confidence calibration and knowledge alignment are strongly enriched in correct traces, while uncertainty acknowledgment is more common in incorrect traces.}
\label{tab:conditional_prevalence}
\begin{tabular}{@{}lccc@{\hspace{6pt}}ccc@{\hspace{6pt}}ccc@{}}
\toprule
& \multicolumn{3}{c}{All} & \multicolumn{3}{c}{Thinking} & \multicolumn{3}{c}{Instruct} \\
\cmidrule(lr){2-4}\cmidrule(lr){5-7}\cmidrule(lr){8-10}
Behavior & $P(\checkmark)$ & $P(\times)$ & $\Delta$
         & $P(\checkmark)$ & $P(\times)$ & $\Delta$
         & $P(\checkmark)$ & $P(\times)$ & $\Delta$ \\
\midrule
confidence\_cali.    & 84.7\% & 0.6\%  & +84.1 & 82.0\% & 1.0\%  & +81.0 & 87.5\% & 0.2\%  & +87.3 \\
knowledge\_alig.     & 93.9\% & 15.4\% & +78.5 & 94.4\% & 18.5\% & +75.8 & 93.4\% & 12.3\% & +81.1 \\
self\_awareness      & 56.7\% & 4.6\%  & +52.1 & 62.2\% & 6.5\%  & +55.7 & 51.0\% & 2.8\%  & +48.2 \\
goal\_tracking       & 88.9\% & 43.7\% & +45.2 & 91.0\% & 45.1\% & +45.9 & 86.7\% & 42.3\% & +44.4 \\
evidence\_cita.      & 93.5\% & 62.3\% & +31.2 & 94.4\% & 60.4\% & +34.0 & 92.6\% & 64.3\% & +28.3 \\
planning\_pres.      & 86.9\% & 69.1\% & +17.8 & 90.7\% & 73.4\% & +17.3 & 82.9\% & 64.9\% & +18.1 \\
self\_correction     & 27.9\% & 18.2\% & +9.6  & 46.6\% & 29.7\% & +16.9 & 8.4\%  & 6.7\%  & +1.7  \\
hypothesis\_test.    & 22.2\% & 21.4\% & +0.7  & 34.5\% & 33.9\% & +0.5  & 9.4\%  & 8.9\%  & +0.5  \\
uncertainty\_ack.    & 20.1\% & 30.8\% & -10.8 & 34.9\% & 49.7\% & -14.9 & 4.7\%  & 11.9\% & -7.1  \\
\bottomrule
\end{tabular}
\end{table*}
\begin{table}[t]
\centering
\small
\caption{Behavior prevalence on post-hoc rationalization traces, non-post-hoc traces, lucky guesses, and sound reasoning traces for VLM models. Confidence calibration is almost absent when reasoning is reverse-engineered or lucky, while uncertainty acknowledgment is more common on post-hoc traces than on sound reasoning traces.}
\label{tab:posthoc_prevalence_vlm}
\begin{tabular}{@{}lrrrr@{}}
\toprule
Behavior & Post-hoc & No post-hoc & Lucky & Sound \\
\midrule
confidence\_calibration     & 3.4\%  & 76.4\% & 7.6\%  & 94.7\% \\
knowledge\_alignment        & 29.6\% & 87.2\% & 41.9\% & 95.5\% \\
self\_awareness             & 9.1\%  & 66.4\% & 14.4\% & 81.5\% \\
goal\_tracking              & 26.1\% & 76.7\% & 19.0\% & 81.5\% \\
evidence\_citation          & 28.7\% & 82.8\% & 24.1\% & 88.2\% \\
planning\_present           & 45.3\% & 73.5\% & 27.1\% & 75.3\% \\
self\_correction            & 30.5\% & 32.4\% & 37.6\% & 37.8\% \\
hypothesis\_testing         & 26.4\% & 26.0\% & 21.7\% & 28.7\% \\
uncertainty\_acknowledgment & 56.7\% & 29.0\% & 46.2\% & 30.8\% \\
\bottomrule
\end{tabular}
\end{table}

\begin{table}[t]
\centering
\small
\caption{Behavior prevalence on post-hoc rationalization traces, non-post-hoc traces, lucky guesses, and sound reasoning traces for LLM models. Confidence calibration is almost absent when reasoning is reverse-engineered or lucky, while uncertainty acknowledgment, hypothesis testing, and self-correction are more common on post-hoc traces than on sound reasoning traces.}
\label{tab:posthoc_prevalence_llm}
\begin{tabular}{@{}lrrrr@{}}
\toprule
Behavior & Post-hoc & No post-hoc & Lucky & Sound \\
\midrule
confidence\_calibration     & 1.6\%  & 72.5\% & 3.4\%  & 95.9\% \\
knowledge\_alignment        & 20.4\% & 80.9\% & 42.1\% & 99.9\% \\
self\_awareness             & 7.3\%  & 48.3\% & 8.3\%  & 62.8\% \\
goal\_tracking              & 34.7\% & 86.0\% & 30.2\% & 95.9\% \\
evidence\_citation          & 50.5\% & 93.8\% & 40.1\% & 99.8\% \\
planning\_present           & 62.0\% & 87.2\% & 44.9\% & 91.7\% \\
self\_correction            & 34.2\% & 20.3\% & 46.2\% & 24.0\% \\
hypothesis\_testing         & 32.2\% & 17.9\% & 32.0\% & 20.0\% \\
uncertainty\_acknowledgment & 47.1\% & 15.3\% & 44.9\% & 15.5\% \\
\bottomrule
\end{tabular}
\end{table}


\begin{table*}[t]
\centering
\caption{Detailed Behavioral Lift statistics for LLM benchmarks ($N{=}8{,}282$). 
Lift $= P(\checkmark|b) - P(\checkmark|\neg b)$.
Think-Lift and Inst-Lift show lift computed separately on thinking and instruct model pools.}
\label{tab:lift_detail_llm}
\footnotesize
\setlength{\tabcolsep}{4pt}
\begin{tabular}{@{}l rrr rr rr@{}}
\toprule
& & & & \multicolumn{2}{c}{Accuracy given $b$} & \multicolumn{2}{c}{Sample count} \\
\cmidrule(lr){5-6} \cmidrule(lr){7-8}
Behavior & Lift & Th-Lift & In-Lift & $P(\checkmark|b)$ & $P(\checkmark|\neg b)$ & $N_b$ & $N_{\neg b}$ \\
\midrule
Knowledge alignment      & $+$80.3 & $+$79.1 & $+$81.6 & 90.8 & 10.5 & 5299 & 2983 \\
Confidence calibration   & $+$79.6 & $+$76.2 & $+$83.2 & 99.6 & 20.0 & 4362 & 3920 \\
Context understanding    & $+$76.8 & $+$75.1 & $+$78.6 & 87.3 & 10.5 & 5542 & 2740 \\
Logical steps valid      & $+$69.9 & $+$68.7 & $+$71.3 & 88.8 & 18.9 & 5094 & 3188 \\
Self-awareness           & $+$52.7 & $+$54.0 & $+$52.1 & 95.2 & 42.5 & 3052 & 5230 \\
Goal tracking            & $+$52.5 & $+$55.6 & $+$49.7 & 76.8 & 24.3 & 5937 & 2345 \\
Evidence citation        & $+$49.1 & $+$53.2 & $+$44.9 & 70.9 & 21.8 & 6763 & 1519 \\
Reasoning present        & $+$29.1 & $+$42.1 & $+$24.5 & 63.6 & 34.5 & 7787 &  495 \\
Planning                 & $+$26.3 & $+$30.5 & $+$23.4 & 67.1 & 40.8 & 6636 & 1646 \\
Self-correction          & $+$12.4 & $+$16.5 &  $+$5.5 & 71.3 & 58.9 & 2003 & 6279 \\
Hypothesis testing       &  $+$1.0 &  $+$0.6 &  $+$1.4 & 62.7 & 61.7 & 1812 & 6470 \\
Uncertainty ack.         & $-$13.9 & $-$14.5 & $-$24.4 & 51.4 & 65.3 & 2002 & 6280 \\
\midrule
Shortcut                 & $-$49.1 & $-$48.3 & $-$50.0 & 23.1 & 72.2 & 1733 & 6549 \\
Post-hoc rational.       & $-$51.5 & $-$55.6 & $-$47.5 & 24.8 & 76.3 & 2317 & 5965 \\
Logical failure          & $-$71.1 & $-$69.9 & $-$72.3 & 18.6 & 89.6 & 3229 & 5053 \\
Factual error            & $-$71.8 & $-$68.4 & $-$75.0 &  3.7 & 75.6 & 1574 & 6708 \\
Knowledge gap            & $-$75.5 & $-$74.3 & $-$76.7 &  8.8 & 84.3 & 2452 & 5830 \\
Context misread          & $-$76.8 & $-$75.2 & $-$78.4 & 10.4 & 87.2 & 2727 & 5555 \\
\bottomrule
\end{tabular}
\end{table*}

\begin{table*}[t]
\centering
\caption{Detailed Behavioral Lift statistics for VLM benchmarks ($N{=}7{,}000$).
Lift $= P(\checkmark|b) - P(\checkmark|\neg b)$.
Think-Lift and Inst-Lift show lift computed separately on thinking and instruct model pools.}
\label{tab:lift_detail_vlm}
\footnotesize
\setlength{\tabcolsep}{4pt}
\begin{tabular}{@{}l rrr rr rr@{}}
\toprule
& & & & \multicolumn{2}{c}{Accuracy given $b$} & \multicolumn{2}{c}{Sample count} \\
\cmidrule(lr){5-6} \cmidrule(lr){7-8}
Behavior & Lift & Th-Lift & In-Lift & $P(\checkmark|b)$ & $P(\checkmark|\neg b)$ & $N_b$ & $N_{\neg b}$ \\
\midrule
Confidence calibration   & $+$72.2 & $+$69.9 & $+$74.8 & 98.8 & 26.7 & 2584 & 4416 \\
Logical steps valid      & $+$66.5 & $+$70.5 & $+$60.6 & 90.5 & 23.9 & 3089 & 3911 \\
Self-awareness           & $+$62.0 & $+$63.6 & $+$60.5 & 93.1 & 31.1 & 2511 & 4489 \\
Visual claims accurate   & $+$60.9 & $+$63.4 & $+$56.8 & 87.8 & 26.9 & 3032 & 3968 \\
Knowledge alignment      & $+$53.7 & $+$61.7 & $+$43.0 & 76.8 & 23.1 & 3933 & 3067 \\
Evidence citation        & $+$34.9 & $+$43.6 & $+$22.7 & 69.4 & 34.5 & 3777 & 3223 \\
Goal tracking            & $+$30.6 & $+$40.8 & $+$15.6 & 68.6 & 38.0 & 3498 & 3502 \\
Self-correction          & $+$20.1 & $+$24.5 &  $+$3.2 & 66.8 & 46.7 & 2302 & 4698 \\
Planning                 &  $+$6.6 & $+$12.7 &  $-$3.4 & 56.0 & 49.4 & 4156 & 2844 \\
Hypothesis testing       &  $+$1.0 &  $-$4.3 &  $+$5.3 & 54.0 & 53.0 & 1914 & 5086 \\
Visual refs present      &  $-$3.3 &  $-$3.1 &  $-$6.5 & 52.6 & 55.9 & 5579 & 1421 \\
Reasoning present        &  $-$8.8 & $+$21.5 & $-$16.7 & 52.5 & 61.3 & 6385 &  615 \\
Uncertainty ack.         & $-$16.1 & $-$30.2 & $-$24.8 & 44.7 & 60.7 & 3241 & 3759 \\
\midrule
Language bias            & $-$21.1 & $-$33.2 &  $-$6.9 & 35.0 & 56.1 &  940 & 6060 \\
Shortcut                 & $-$36.3 & $-$44.3 & $-$25.6 & 31.4 & 67.7 & 2784 & 4216 \\
Visual hallucination     & $-$45.7 & $-$47.7 & $-$45.3 & 17.6 & 63.3 & 1531 & 5469 \\
Post-hoc rational.       & $-$52.9 & $-$61.3 & $-$42.0 & 29.3 & 82.2 & 3824 & 3176 \\
Visual neglect           & $-$60.7 & $-$64.8 & $-$55.1 & 17.1 & 77.8 & 2826 & 4174 \\
Logical failure          & $-$67.3 & $-$70.9 & $-$61.8 & 23.6 & 90.9 & 3914 & 3086 \\
\bottomrule
\end{tabular}
\end{table*}
\begin{table}[t]
\centering
\small
\caption{Behavioral Lift for seven frontier models on GPQA-Diamond, computed from visible responses only and pooled across models. The same broad ranking appears as in the main analysis: knowledge alignment and confidence calibration are the strongest positive signals, hypothesis testing remains near zero, and uncertainty acknowledgment remains negative.}
\label{tab:frontier}
\begin{tabular}{lcccccc}
\toprule
\textbf{Behavior} & \textbf{Lift} & \textbf{$P(\checkmark \mid b)$} & \textbf{$P(\checkmark \mid \neg b)$} & \textbf{Present \%} & \textbf{$N_p$} & \textbf{$N_a$} \\
\midrule
Knowledge alignment         & +84.2 & 94.6 & 10.4 & 66.7 &  925 & 461 \\
Confidence calibration      & +77.9 & 99.9 & 22.0 & 57.3 &  794 & 592 \\
Goal tracking               & +67.7 & 79.2 & 11.6 & 81.3 & 1127 & 259 \\
Self-awareness              & +58.6 & 96.9 & 38.3 & 48.3 &  670 & 716 \\
Evidence citation           & +51.4 & 71.3 & 19.8 & 90.9 & 1260 & 126 \\
Planning present            & +20.2 & 67.1 & 46.9 & 97.7 & 1354 &  32 \\
Self-correction             & +14.1 & 76.3 & 62.2 & 31.3 &  434 & 952 \\
Hypothesis testing          & -2.5  & 65.1 & 67.6 & 38.8 &  538 & 848 \\
Uncertainty acknowledgment  & -23.2 & 51.5 & 74.6 & 34.6 &  480 & 906 \\
\bottomrule
\end{tabular}
\end{table}
\begin{table*}[t]
\centering
\small
\caption{Qualitative surface markers of annotated higher-order behaviors. For each behavior, we manually reviewed behavior-positive traces and summarized the most common recurring surface markers and broader discourse pattern. These summaries are qualitative and intended to illustrate how the behaviors tend to appear in visible reasoning traces.}
\label{tab:surface_markers}
\begin{tabular}{p{2.3cm}p{4.8cm}p{5.5cm}}
\toprule
\textbf{Behavior} & \textbf{Example surface markers} & \textbf{Qualitative pattern} \\
\midrule
Planning present
& \textit{``let me break this down,'' ``first\ldots then\ldots,'' ``let me define\ldots,'' ``step 1 / step 2,''} explicit headers
& Often marked by explicit upfront decomposition of the task into steps, variables, or constraints before the main reasoning begins. \\[0.5em]

Hypothesis testing
& \textit{``alternatively,'' ``what if,'' ``suppose,'' ``case 1 / case 2,'' ``if X\ldots if not X\ldots''}
& Often marked by explicit branching over alternative interpretations or scenarios, rather than generic linear continuation. \\[0.5em]

Self-correction
& \textit{``wait,'' ``actually,'' ``that’s not correct,'' ``let me rethink,'' ``let’s start over''}
& Often marked by explicit revision and backtracking after a contradiction, detected error, or failed line of reasoning. \\[0.5em]

Uncertainty acknowledgment
& \textit{``I’m not sure,'' ``I’m confused,'' ``maybe,'' ``am I missing something?''}
& Often marked by explicit hedging and self-doubt language, including admissions of ambiguity or difficulty. \\[0.5em]

Evidence citation
& \textit{``the passage states,'' ``given that,'' ``from condition 1,'' ``based on the fact that\ldots''}
& Often marked by explicit grounding of a claim in a specific fact, premise, or constraint from the prompt. \\[0.5em]

Confidence calibration
& \textit{``I’m not sure,'' ``maybe,'' ``let me double-check,'' ``must be,'' ``therefore, in all cases''}
& Often marked not by isolated hedging (which signals uncertainty acknowledgment), 
but by a shift in expressed certainty that tracks the strength of the reasoning — 
increasing confidence when steps succeed, decreasing when they falter. \\[0.5em]

Self-awareness
& \textit{``the passage does not mention,'' ``the text does not specify,'' ``cannot be determined,'' ``insufficient information''}
& Often marked by explicit recognition that the prompt lacks the information needed to justify a conclusion, rather than by generic hedging alone. \\[0.5em]

Goal tracking
& \textit{``I need to\ldots,'' ``the goal is\ldots,'' ``so far we know\ldots,'' ``the next step is\ldots,'' ``going back to the question\ldots''}
& Often marked by explicit monitoring of progress toward the main objective, rather than generic step-by-step sequencing alone. \\[0.5em]

Knowledge alignment
& \textit{``necessary condition,'' ``sufficient condition,'' ``correlation vs.\ causation,'' ``translate into logical expressions,''} domain-specific conceptual terms
& Often marked by converting the problem into the appropriate domain framework and reasoning within that framework, rather than by generic technical-sounding language. \\
\bottomrule
\end{tabular}
\end{table*}

\definecolor{emptycell}{gray}{0.94}

\begin{table*}[t]
\centering
\footnotesize
\setlength{\tabcolsep}{3.8pt}
\begin{tabular}{@{}l cc cc cc cc cc cc@{}}
\toprule
& \multicolumn{6}{c}{\textbf{VLM Benchmarks}} & \multicolumn{6}{c}{\textbf{LLM Benchmarks}} \\
\cmidrule(lr){2-7} \cmidrule(lr){8-13}
& \multicolumn{2}{c}{VisPuzz} & \multicolumn{2}{c}{MathVista} & \multicolumn{2}{c}{MMMU}
& \multicolumn{2}{c}{LogiQA2} & \multicolumn{2}{c}{MATH-500} & \multicolumn{2}{c}{MMLU-Pro} \\
\cmidrule(lr){2-3} \cmidrule(lr){4-5} \cmidrule(lr){6-7}
\cmidrule(lr){8-9} \cmidrule(lr){10-11} \cmidrule(lr){12-13}
& Th & In & Th & In & Th & In & Th & In & Th & In & Th & In \\
\midrule
\multicolumn{13}{@{}l}{\textit{Cross-modal failure modes (\%)}} \\[2pt]
\quad Logical failure          & 77.6 & 72.7 & 31.9 & 55.3 & 44.6 & 53.1 & 39.7 & 44.4 & 28.6 & 29.1 & 51.1 & 41.4 \\
\quad Shortcut                 & 52.7 & 43.1 & 23.2 & 49.3 & 27.7 & 45.9 & 20.3 & 34.2 & 16.4 & 19.0 & 18.2 & 18.3 \\
\quad Post-hoc rational.       & 79.1 & 61.2 & 35.1 & 48.7 & 47.9 & 51.7 & 33.1 & 34.8 & 22.9 & 14.2 & 40.0 & 23.2 \\
\quad Lucky guess              & 15.1 &  5.1 & 11.9 & 27.0 & 11.5 & 18.3 &  4.4 & 11.8 &  7.9 &  3.6 &  6.6 &  2.7 \\
\addlinespace[3pt]
\multicolumn{13}{@{}l}{\textit{VLM-specific (\%)}} \\[2pt]
\quad Visual hallucination     & 37.1 & 24.1 & 14.8 & 11.2 & 20.8 & 17.9 & \cellcolor{emptycell} & \cellcolor{emptycell} & \cellcolor{emptycell} & \cellcolor{emptycell} & \cellcolor{emptycell} & \cellcolor{emptycell} \\
\quad Visual neglect           & 55.3 & 57.0 & 20.0 & 36.6 & 30.6 & 43.5 & \cellcolor{emptycell} & \cellcolor{emptycell} & \cellcolor{emptycell} & \cellcolor{emptycell} & \cellcolor{emptycell} & \cellcolor{emptycell} \\
\quad Language bias            & 11.4 &  5.3 &  8.5 & 14.0 & 15.8 & 26.2 & \cellcolor{emptycell} & \cellcolor{emptycell} & \cellcolor{emptycell} & \cellcolor{emptycell} & \cellcolor{emptycell} & \cellcolor{emptycell} \\
\addlinespace[3pt]
\multicolumn{13}{@{}l}{\textit{LLM-specific (\%)}} \\[2pt]
\quad Factual error            & \cellcolor{emptycell} & \cellcolor{emptycell} & \cellcolor{emptycell} & \cellcolor{emptycell} & \cellcolor{emptycell} & \cellcolor{emptycell} &  7.1 &  9.8 & 16.7 & 24.0 & 26.1 & 29.7 \\
\quad Context misread          & \cellcolor{emptycell} & \cellcolor{emptycell} & \cellcolor{emptycell} & \cellcolor{emptycell} & \cellcolor{emptycell} & \cellcolor{emptycell} & 32.8 & 39.8 & 19.8 & 22.9 & 41.9 & 41.0 \\
\quad Knowledge gap            & \cellcolor{emptycell} & \cellcolor{emptycell} & \cellcolor{emptycell} & \cellcolor{emptycell} & \cellcolor{emptycell} & \cellcolor{emptycell} & 25.0 & 29.6 & 18.2 & 23.1 & 41.6 & 40.3 \\
\midrule
Recovery Rate (\%)             & 23.0 &  8.4 & 45.9 & 47.5 & 32.9 & 33.8 & 11.1 & 24.5 & 40.8 & 17.8 & 16.8 &  6.3 \\
Accuracy (\%)                  & 36.5 & 29.7 & 77.5 & 66.6 & 59.2 & 52.4 & 54.1 & 58.4 & 81.4 & 74.3 & 51.4 & 51.5 \\
\bottomrule
\end{tabular}
\caption{%
Failure mode rates, recovery rate, and accuracy across six benchmarks.
cross-modal failures (top) are directly comparable across modalities.
VLM-specific and LLM-specific failures (middle) apply only to their respective benchmarks; gray cells indicate non-applicable modality.
Recovery Rate $= P(\text{correct} \mid \text{any failure detected})$.
On reasoning-heavy benchmarks, thinking models recover at 2.3--2.7$\times$ the rate of instruct models.
On LogiQA2, instruct models recover better.
}
\label{tab:failures}
\end{table*}
\begin{table*}[t]
\centering
\small
\caption{Within-question paired analysis on MATH-500. For each of 250 questions, we generate 8 traces at temperature 0.6 and compute the per-question accuracy difference between traces where a behavior is present versus absent. Reported values are mean within-question $\Delta$ accuracy, with bootstrap 95\% confidence intervals and the number of questions for which both present and absent traces were observed.}
\label{tab:within_question_llm}
\begin{tabular}{lcccccc}
\toprule
& \multicolumn{3}{c}{\textbf{Qwen3-4B-Think}} & \multicolumn{3}{c}{\textbf{Qwen3-4B-Instruct}} \\
\cmidrule(lr){2-4} \cmidrule(lr){5-7}
\textbf{Behavior} & \textbf{Mean $\Delta$} & \textbf{95\% CI} & \textbf{$N$} & \textbf{Mean $\Delta$} & \textbf{95\% CI} & \textbf{$N$} \\
\midrule
Confidence calibration      & +0.305 & [+0.254, +0.363] & 189 & +0.518 & [+0.453, +0.582] & 145 \\
Self-awareness              & +0.187 & [+0.150, +0.226] & 225 & +0.158 & [+0.131, +0.189] & 227 \\
Knowledge alignment         & +0.596 & [+0.524, +0.673] & 110 & +0.728 & [+0.651, +0.794] &  91 \\
Evidence citation           & +0.299 & [+0.201, +0.388] &  78 & +0.268 & [+0.143, +0.393] &  54 \\
Goal tracking               & +0.405 & [+0.331, +0.481] & 112 & +0.434 & [+0.346, +0.516] &  99 \\
Planning present            & +0.120 & [+0.033, +0.208] &  69 & +0.138 & [+0.035, +0.241] &  72 \\
Self-correction             & +0.267 & [+0.203, +0.334] & 132 & +0.078 & [+0.035, +0.121] & 180 \\
Hypothesis testing          & +0.048 & [+0.017, +0.082] & 214 & +0.028 & [ -0.031, +0.085] & 126 \\
Uncertainty acknow.  & +0.006 & [-0.021, +0.034] & 222 & -0.094 & [ -0.168, -0.028 ] & 115 \\
\bottomrule
\end{tabular}
\end{table*}
\begin{table*}[t]
\centering
\small
\caption{Within-question paired analysis on MathVista for Qwen3-4B-VL-Think and Qwen3-4B-VL-Instruct. For each question, we generate 8 traces and compare mean accuracy between traces where a behavior is present versus absent on the same question. Reported values are mean within-question $\Delta$ accuracy, with bootstrap 95\% confidence intervals and the number of questions for which both present and absent traces were observed.}
\label{tab:within_question_vlm}
\begin{tabular}{lcccccc}
\toprule
& \multicolumn{3}{c}{\textbf{Qwen3-4B-VL-Think}} & \multicolumn{3}{c}{\textbf{Qwen3-4B-VL-Instruct}} \\
\cmidrule(lr){2-4} \cmidrule(lr){5-7}
\textbf{Behavior} & \textbf{Mean $\Delta$} & \textbf{95\% CI} & \textbf{$N$} & \textbf{Mean $\Delta$} & \textbf{95\% CI} & \textbf{$N$} \\
\midrule
Confidence calibration      & +0.186 & [+0.137, +0.240] & 115 & +0.342 & [+0.235, +0.452] &  62 \\
Self-awareness              & +0.136 & [+0.094, +0.180] & 178 & +0.131 & [+0.082, +0.190] & 130 \\
Knowledge alignment         & +0.166 & [+0.112, +0.216] & 134 & +0.109 & [+0.054, +0.164] &  96 \\
Evidence citation           & +0.149 & [+0.103, +0.200] & 149 & +0.107 & [+0.051, +0.168] &  95 \\
Goal tracking               & +0.080 & [+0.042, +0.121] & 172 & +0.057 & [+0.020, +0.101] & 108 \\
Planning present            & +0.002 & [-0.030, +0.037] & 188 & +0.028 & [-0.006, +0.066] & 102 \\
Self-correction             & +0.153 & [+0.103, +0.207] & 166 & +0.062 & [+0.015, +0.127] &  42 \\
Hypothesis testing          & -0.005 & [-0.038, +0.025] & 198 & +0.045 & [-0.004, +0.113] &  43 \\
Uncertainty acknow.  & -0.017 & [-0.054, +0.020] & 145 & -0.004 & [-0.061, +0.054] &  38 \\
\bottomrule
\end{tabular}
\end{table*}

\begin{table*}[t]
\centering
\caption{Scaling analysis: Qwen3-VL (thinking) vs.\ Qwen3-VL (instruct) on MathVista, 
2B to 32B parameters. Amplified behaviors show persistent gaps at all scales. Largely unchanged behaviors converge, with instruct matching or exceeding thinking at 32B.}
\label{tab:scaling_vlm}
\footnotesize
\setlength{\tabcolsep}{4pt}
\begin{tabular}{@{}l cc cc cc cc@{}}
\toprule
& \multicolumn{2}{c}{\textbf{2B}} & \multicolumn{2}{c}{\textbf{4B}} & \multicolumn{2}{c}{\textbf{8B}} & \multicolumn{2}{c}{\textbf{32B}} \\
\cmidrule(lr){2-3} \cmidrule(lr){4-5} \cmidrule(lr){6-7} \cmidrule(lr){8-9}
& Th & In & Th & In & Th & In & Th & In \\
\midrule
\quad Self-correction       & 64.0 &  4.7 & 72.7 & 10.3 & 69.3 & 11.3 & 61.3 & 11.3 \\
\quad Hypothesis testing    & 17.3 &  2.0 & 31.0 &  8.7 & 28.0 &  9.7 & 17.7 & 11.0 \\
\quad Uncertainty ack.      & 66.7 &  8.3 & 77.7 & 13.0 & 61.7 & 14.3 & 43.7 & 15.3 \\
\quad Self-awareness        & 45.7 & 25.3 & 50.3 & 46.0 & 64.7 & 39.7 & 66.0 & 63.3 \\
\addlinespace[3pt]
\quad Conf.\ calibration    & 40.7 & 32.0 & 43.3 & 56.3 & 61.0 & 51.3 & 71.7 & 78.0 \\
\quad Planning              & 55.3 & 37.3 & 62.7 & 43.7 & 67.7 & 45.7 & 78.7 & 59.0 \\
\quad Goal tracking         & 54.7 & 34.7 & 57.0 & 43.3 & 66.3 & 46.0 & 76.0 & 65.0 \\
\addlinespace[3pt]
\multicolumn{9}{@{}l}{\textit{Summary}} \\[2pt]
\quad Accuracy              & 72.0 & 53.7 & 71.7 & 74.0 & 77.0 & 71.0 & 83.3 & 81.7 \\
\quad Recovery Rate         & 48.1 & 34.4 & 48.5 & 45.8 & 38.2 & 44.5 & 43.7 & 31.6 \\
\addlinespace[3pt]
\multicolumn{9}{@{}l}{\textit{Behavioral Lift (pooled per size)}} \\[2pt]
\quad Self-correction       & \multicolumn{2}{c}{+30.0} & \multicolumn{2}{c}{+14.2} & \multicolumn{2}{c}{+10.3} & \multicolumn{2}{c}{+5.9} \\
\quad Conf.\ calibration    & \multicolumn{2}{c}{+57.7} & \multicolumn{2}{c}{+52.8} & \multicolumn{2}{c}{+59.3} & \multicolumn{2}{c}{+68.7} \\
\quad Uncertainty ack.      & \multicolumn{2}{c}{$-$4.5} & \multicolumn{2}{c}{$-$12.2} & \multicolumn{2}{c}{$-$16.8} & \multicolumn{2}{c}{$-$18.5} \\
\bottomrule
\end{tabular}
\end{table*}

\begin{table*}[t]
\centering
\caption{Scaling analysis: Qwen3 (thinking) vs.\ Qwen2.5 (instruct) on MATH-500, 
0.6B to 32B parameters. The same patterns hold as in the VLM scaling analysis: amplified behavior gaps persist across scale, largely unchanged behaviors converge, self-correction Lift diminishes while confidence calibration Lift remains high.}
\label{tab:scaling_llm}
\footnotesize
\setlength{\tabcolsep}{3.2pt}
\begin{tabular}{@{}l cc cc cc cc cc cc@{}}
\toprule
& \multicolumn{2}{c}{\textbf{0.6B}} & \multicolumn{2}{c}{\textbf{1.7B}} & \multicolumn{2}{c}{\textbf{4B}} & \multicolumn{2}{c}{\textbf{8B}} & \multicolumn{2}{c}{\textbf{14B}} & \multicolumn{2}{c}{\textbf{32B}} \\
\cmidrule(lr){2-3} \cmidrule(lr){4-5} \cmidrule(lr){6-7} \cmidrule(lr){8-9} \cmidrule(lr){10-11} \cmidrule(lr){12-13}
& Th & In & Th & In & Th & In & Th & In & Th & In & Th & In \\
\midrule
\quad Self-correction       & 63.4 &  0.0 & 60.3 &  1.7 & 82.6 &  2.0 & 81.4 &  4.6 & 92.3 &  1.7 & 61.4 &  2.3 \\
\quad Hypothesis testing    & 46.6 &  0.6 & 42.3 &  1.1 & 55.7 &  3.1 & 64.3 &  4.6 & 74.3 &  3.7 & 46.3 &  4.3 \\
\quad Uncertainty ack.      & 67.1 &  1.4 & 53.7 &  1.1 & 58.3 &  0.0 & 53.4 &  0.6 & 59.7 &  0.3 & 29.4 &  0.3 \\
\quad Self-awareness        & 22.0 &  8.3 & 22.0 & 12.0 & 45.7 & 27.1 & 43.4 & 28.0 & 69.1 & 26.3 & 54.9 & 33.1 \\
\addlinespace[3pt]
\quad Conf.\ calibration    & 32.3 & 19.1 & 39.7 & 30.9 & 60.9 & 58.3 & 54.9 & 69.1 & 76.0 & 74.0 & 77.1 & 78.9 \\
\quad Planning              & 72.9 & 64.6 & 88.0 & 68.6 & 91.7 & 93.1 & 94.6 & 97.4 & 97.4 & 95.1 & 97.1 & 95.7 \\
\quad Goal tracking         & 47.7 & 33.7 & 70.3 & 47.1 & 83.4 & 75.7 & 81.7 & 86.0 & 88.0 & 86.3 & 88.6 & 90.0 \\
\addlinespace[3pt]
\multicolumn{13}{@{}l}{\textit{Summary}} \\[2pt]
\quad Accuracy              & 52.6 & 20.3 & 71.4 & 34.6 & 90.6 & 61.4 & 85.1 & 71.4 & 92.3 & 78.6 & 89.4 & 81.4 \\
\quad Recovery Rate         & 28.4 &  1.8 & 47.0 &  5.8 & 71.8 &  8.2 & 59.1 &  3.9 & 60.9 & 15.7 & 48.6 & 12.5 \\
\addlinespace[3pt]
\multicolumn{13}{@{}l}{\textit{Behavioral Lift (pooled per size)}} \\[2pt]
\quad Self-correction       & \multicolumn{2}{c}{+39.7} & \multicolumn{2}{c}{+34.7} & \multicolumn{2}{c}{+35.7} & \multicolumn{2}{c}{+20.6} & \multicolumn{2}{c}{+16.0} & \multicolumn{2}{c}{+14.8} \\
\quad Conf.\ calibration    & \multicolumn{2}{c}{+84.8} & \multicolumn{2}{c}{+71.4} & \multicolumn{2}{c}{+59.4} & \multicolumn{2}{c}{+56.5} & \multicolumn{2}{c}{+57.5} & \multicolumn{2}{c}{+66.2} \\
\quad Uncertainty ack.      & \multicolumn{2}{c}{+19.4} & \multicolumn{2}{c}{+20.3} & \multicolumn{2}{c}{+17.3} & \multicolumn{2}{c}{+3.7} & \multicolumn{2}{c}{+2.4} & \multicolumn{2}{c}{$-$1.0} \\
\bottomrule
\end{tabular}
\end{table*}
\begin{table*}[t]
  \centering
  \scriptsize
  \caption{Sensitivity of LLM Behavioral Lift to simulated random annotation noise. The first subtable shows the clean baseline with no injected noise. At each nonzero noise level, we randomly flip a fixed fraction of behavior labels and recompute Lift over 1000 trials. Each noisy subtable reports the clean Lift, the mean noisy Lift, the 95\% range across trials, the mean rank, and the fraction of trials that preserve the original sign. Behaviors are kept in the clean baseline order for easier comparison across noise levels.}
  \label{tab:noise_llm_full}
  \captionsetup[subtable]{skip=0.8ex}
  \setlength{\tabcolsep}{4pt}

  \begin{subtable}[t]{\textwidth}
    \centering
    \caption{Clean baseline (0\% label flips)}
    \begin{tabular}{lccc}
      \toprule
      \textbf{Behavior} & \textbf{Lift} & \textbf{Rank} & \textbf{$N$} \\
      \midrule
      Knowledge alignment         & +0.803 & 1 & 8282 \\
      Confidence calibration      & +0.796 & 2 & 8282 \\
      Self-awareness              & +0.527 & 3 & 8282 \\
      Goal tracking               & +0.525 & 4 & 8282 \\
      Evidence citation           & +0.491 & 5 & 8282 \\
      Planning present            & +0.263 & 6 & 8282 \\
      Self-correction             & +0.124 & 7 & 8282 \\
      Hypothesis testing          & +0.010 & 8 & 8282 \\
      Uncertainty acknowledgment  & -0.139 & 9 & 8282 \\
      \bottomrule
    \end{tabular}
  \end{subtable}

  \vspace{1ex}

  \begin{subtable}[t]{\textwidth}
    \centering
    \caption{5\% label flips}
    \begin{tabular}{lccccc}
      \toprule
      \textbf{Behavior} & \textbf{Clean} & \textbf{Mean noisy} & \textbf{95\% range} & \textbf{Mean rank} & \textbf{Sign kept} \\
      \midrule
      Knowledge alignment         & +0.803 & +0.711 & [+0.705, +0.718] & 1.85 & 100.0\% \\
      Confidence calibration      & +0.796 & +0.716 & [+0.711, +0.721] & 1.15 & 100.0\% \\
      Self-awareness              & +0.527 & +0.468 & [+0.459, +0.477] & 3.02 & 100.0\% \\
      Goal tracking               & +0.525 & +0.453 & [+0.444, +0.463] & 3.98 & 100.0\% \\
      Evidence citation           & +0.491 & +0.392 & [+0.380, +0.404] & 5.00 & 100.0\% \\
      Planning present            & +0.263 & +0.213 & [+0.201, +0.225] & 6.00 & 100.0\% \\
      Self-correction             & +0.124 & +0.104 & [+0.092, +0.116] & 7.00 & 100.0\% \\
      Hypothesis testing          & +0.010 & +0.008 & [-0.005, +0.021] & 8.00 & 89.7\% \\
      Uncertainty acknowledgment  & -0.139 & -0.117 & [-0.128, -0.105] & 9.00 & 100.0\% \\
      \bottomrule
    \end{tabular}
  \end{subtable}

  \vspace{1ex}

  \begin{subtable}[t]{\textwidth}
    \centering
    \caption{10\% label flips}
    \begin{tabular}{lccccc}
      \toprule
      \textbf{Behavior} & \textbf{Clean} & \textbf{Mean noisy} & \textbf{95\% range} & \textbf{Mean rank} & \textbf{Sign kept} \\
      \midrule
      Knowledge alignment         & +0.803 & +0.623 & [+0.616, +0.631] & 1.99 & 100.0\% \\
      Confidence calibration      & +0.796 & +0.636 & [+0.629, +0.643] & 1.01 & 100.0\% \\
      Self-awareness              & +0.527 & +0.411 & [+0.400, +0.422] & 3.00 & 100.0\% \\
      Goal tracking               & +0.525 & +0.388 & [+0.375, +0.400] & 4.00 & 100.0\% \\
      Evidence citation           & +0.491 & +0.317 & [+0.301, +0.332] & 5.00 & 100.0\% \\
      Planning present            & +0.263 & +0.175 & [+0.159, +0.190] & 6.00 & 100.0\% \\
      Self-correction             & +0.124 & +0.088 & [+0.072, +0.103] & 7.00 & 100.0\% \\
      Hypothesis testing          & +0.010 & +0.008 & [-0.008, +0.023] & 8.00 & 81.6\% \\
      Uncertainty acknowledgment  & -0.139 & -0.098 & [-0.112, -0.082] & 9.00 & 100.0\% \\
      \bottomrule
    \end{tabular}
  \end{subtable}

  \vspace{1ex}

  \begin{subtable}[t]{\textwidth}
    \centering
    \caption{15\% label flips}
    \begin{tabular}{lccccc}
      \toprule
      \textbf{Behavior} & \textbf{Clean} & \textbf{Mean noisy} & \textbf{95\% range} & \textbf{Mean rank} & \textbf{Sign kept} \\
      \midrule
      Knowledge alignment         & +0.803 & +0.539 & [+0.529, +0.549] & 2.00 & 100.0\% \\
      Confidence calibration      & +0.796 & +0.556 & [+0.548, +0.565] & 1.00 & 100.0\% \\
      Self-awareness              & +0.527 & +0.356 & [+0.344, +0.369] & 3.00 & 100.0\% \\
      Goal tracking               & +0.525 & +0.329 & [+0.314, +0.343] & 4.00 & 100.0\% \\
      Evidence citation           & +0.491 & +0.256 & [+0.240, +0.273] & 5.00 & 100.0\% \\
      Planning present            & +0.263 & +0.143 & [+0.126, +0.160] & 6.00 & 100.0\% \\
      Self-correction             & +0.124 & +0.074 & [+0.056, +0.091] & 7.00 & 100.0\% \\
      Hypothesis testing          & +0.010 & +0.006 & [-0.013, +0.023] & 8.00 & 72.6\% \\
      Uncertainty acknowledgment  & -0.139 & -0.082 & [-0.098, -0.064] & 9.00 & 100.0\% \\
      \bottomrule
    \end{tabular}
  \end{subtable}

  \vspace{1ex}

  \begin{subtable}[t]{\textwidth}
    \centering
    \caption{20\% label flips}
    \begin{tabular}{lccccc}
      \toprule
      \textbf{Behavior} & \textbf{Clean} & \textbf{Mean noisy} & \textbf{95\% range} & \textbf{Mean rank} & \textbf{Sign kept} \\
      \midrule
      Knowledge alignment         & +0.803 & +0.457 & [+0.447, +0.468] & 2.00 & 100.0\% \\
      Confidence calibration      & +0.796 & +0.477 & [+0.467, +0.487] & 1.00 & 100.0\% \\
      Self-awareness              & +0.527 & +0.302 & [+0.288, +0.316] & 3.01 & 100.0\% \\
      Goal tracking               & +0.525 & +0.275 & [+0.259, +0.290] & 3.99 & 100.0\% \\
      Evidence citation           & +0.491 & +0.206 & [+0.189, +0.225] & 5.00 & 100.0\% \\
      Planning present            & +0.263 & +0.116 & [+0.098, +0.134] & 6.00 & 100.0\% \\
      Self-correction             & +0.124 & +0.060 & [+0.041, +0.079] & 7.00 & 100.0\% \\
      Hypothesis testing          & +0.010 & +0.005 & [-0.014, +0.023] & 8.00 & 71.0\% \\
      Uncertainty acknowledgment  & -0.139 & -0.068 & [-0.085, -0.050] & 9.00 & 100.0\% \\
      \bottomrule
    \end{tabular}
  \end{subtable}
\end{table*}

\begin{table*}[t]
  \centering
  \scriptsize
  \caption{Sensitivity of VLM Behavioral Lift to simulated random annotation noise. The first subtable shows the clean baseline with no injected noise. At each nonzero noise level, we randomly flip a fixed fraction of behavior labels and recompute Lift over 1000 trials. Each noisy subtable reports the clean Lift, the mean noisy Lift, the 95\% range across trials, the mean rank, and the fraction of trials that preserve the original sign. Behaviors are kept in the clean baseline order for easier comparison across noise levels.}
  \label{tab:noise_vlm_full}
  \captionsetup[subtable]{skip=0.8ex}
  \setlength{\tabcolsep}{4pt}

  \begin{subtable}[t]{\textwidth}
    \centering
    \caption{Clean baseline (0\% label flips)}
    \begin{tabular}{lccc}
      \toprule
      \textbf{Behavior} & \textbf{Lift} & \textbf{Rank} & \textbf{$N$} \\
      \midrule
      Confidence calibration      & +0.722 & 1 & 7000 \\
      Self-awareness              & +0.620 & 2 & 7000 \\
      Knowledge alignment         & +0.537 & 3 & 7000 \\
      Evidence citation           & +0.349 & 4 & 7000 \\
      Goal tracking               & +0.306 & 5 & 7000 \\
      Self-correction             & +0.201 & 6 & 7000 \\
      Planning present            & +0.066 & 7 & 7000 \\
      Hypothesis testing          & +0.010 & 8 & 7000 \\
      Uncertainty acknowledgment  & -0.161 & 9 & 7000 \\
      \bottomrule
    \end{tabular}
  \end{subtable}

  \vspace{1ex}

  \begin{subtable}[t]{\textwidth}
    \centering
    \caption{5\% label flips}
    \begin{tabular}{lccccc}
      \toprule
      \textbf{Behavior} & \textbf{Clean} & \textbf{Mean noisy} & \textbf{95\% range} & \textbf{Mean rank} & \textbf{Sign kept} \\
      \midrule
      Confidence calibration      & +0.722 & +0.641 & [+0.633, +0.649] & 1.00 & 100.0\% \\
      Self-awareness              & +0.620 & +0.549 & [+0.540, +0.558] & 2.00 & 100.0\% \\
      Knowledge alignment         & +0.537 & +0.482 & [+0.473, +0.491] & 3.00 & 100.0\% \\
      Evidence citation           & +0.349 & +0.314 & [+0.304, +0.323] & 4.00 & 100.0\% \\
      Goal tracking               & +0.306 & +0.275 & [+0.266, +0.285] & 5.00 & 100.0\% \\
      Self-correction             & +0.201 & +0.177 & [+0.166, +0.188] & 6.00 & 100.0\% \\
      Planning present            & +0.066 & +0.059 & [+0.048, +0.069] & 7.00 & 100.0\% \\
      Hypothesis testing          & +0.010 & +0.009 & [-0.003, +0.021] & 8.00 & 91.9\% \\
      Uncertainty acknowledgment  & -0.161 & -0.144 & [-0.155, -0.133] & 9.00 & 100.0\% \\
      \bottomrule
    \end{tabular}
  \end{subtable}

  \vspace{1ex}

  \begin{subtable}[t]{\textwidth}
    \centering
    \caption{10\% label flips}
    \begin{tabular}{lccccc}
      \toprule
      \textbf{Behavior} & \textbf{Clean} & \textbf{Mean noisy} & \textbf{95\% range} & \textbf{Mean rank} & \textbf{Sign kept} \\
      \midrule
      Confidence calibration      & +0.722 & +0.563 & [+0.553, +0.573] & 1.00 & 100.0\% \\
      Self-awareness              & +0.620 & +0.481 & [+0.470, +0.494] & 2.00 & 100.0\% \\
      Knowledge alignment         & +0.537 & +0.427 & [+0.416, +0.439] & 3.00 & 100.0\% \\
      Evidence citation           & +0.349 & +0.279 & [+0.266, +0.293] & 4.00 & 100.0\% \\
      Goal tracking               & +0.306 & +0.244 & [+0.232, +0.257] & 5.00 & 100.0\% \\
      Self-correction             & +0.201 & +0.154 & [+0.139, +0.168] & 6.00 & 100.0\% \\
      Planning present            & +0.066 & +0.053 & [+0.038, +0.066] & 7.00 & 100.0\% \\
      Hypothesis testing          & +0.010 & +0.007 & [-0.009, +0.023] & 8.00 & 80.2\% \\
      Uncertainty acknowledgment  & -0.161 & -0.128 & [-0.142, -0.115] & 9.00 & 100.0\% \\
      \bottomrule
    \end{tabular}
  \end{subtable}

  \vspace{1ex}

  \begin{subtable}[t]{\textwidth}
    \centering
    \caption{15\% label flips}
    \begin{tabular}{lccccc}
      \toprule
      \textbf{Behavior} & \textbf{Clean} & \textbf{Mean noisy} & \textbf{95\% range} & \textbf{Mean rank} & \textbf{Sign kept} \\
      \midrule
      Confidence calibration      & +0.722 & +0.487 & [+0.474, +0.499] & 1.00 & 100.0\% \\
      Self-awareness              & +0.620 & +0.416 & [+0.402, +0.431] & 2.00 & 100.0\% \\
      Knowledge alignment         & +0.537 & +0.372 & [+0.359, +0.386] & 3.00 & 100.0\% \\
      Evidence citation           & +0.349 & +0.244 & [+0.228, +0.260] & 4.01 & 100.0\% \\
      Goal tracking               & +0.306 & +0.214 & [+0.198, +0.230] & 4.99 & 100.0\% \\
      Self-correction             & +0.201 & +0.132 & [+0.114, +0.149] & 6.00 & 100.0\% \\
      Planning present            & +0.066 & +0.046 & [+0.029, +0.063] & 7.00 & 100.0\% \\
      Hypothesis testing          & +0.010 & +0.006 & [-0.013, +0.026] & 8.00 & 74.9\% \\
      Uncertainty acknowledgment  & -0.161 & -0.112 & [-0.128, -0.096] & 9.00 & 100.0\% \\
      \bottomrule
    \end{tabular}
  \end{subtable}

  \vspace{1ex}

  \begin{subtable}[t]{\textwidth}
    \centering
    \caption{20\% label flips}
    \begin{tabular}{lccccc}
      \toprule
      \textbf{Behavior} & \textbf{Clean} & \textbf{Mean noisy} & \textbf{95\% range} & \textbf{Mean rank} & \textbf{Sign kept} \\
      \midrule
      Confidence calibration      & +0.722 & +0.414 & [+0.399, +0.427] & 1.00 & 100.0\% \\
      Self-awareness              & +0.620 & +0.353 & [+0.337, +0.368] & 2.00 & 100.0\% \\
      Knowledge alignment         & +0.537 & +0.319 & [+0.303, +0.334] & 3.00 & 100.0\% \\
      Evidence citation           & +0.349 & +0.208 & [+0.191, +0.225] & 4.03 & 100.0\% \\
      Goal tracking               & +0.306 & +0.184 & [+0.166, +0.202] & 4.97 & 100.0\% \\
      Self-correction             & +0.201 & +0.111 & [+0.093, +0.131] & 6.00 & 100.0\% \\
      Planning present            & +0.066 & +0.039 & [+0.019, +0.057] & 7.01 & 100.0\% \\
      Hypothesis testing          & +0.010 & +0.005 & [-0.017, +0.026] & 7.99 & 67.7\% \\
      Uncertainty acknowledgment  & -0.161 & -0.096 & [-0.114, -0.077] & 9.00 & 100.0\% \\
      \bottomrule
    \end{tabular}
  \end{subtable}
\end{table*}

\begin{table}[t]
\centering
\scriptsize
\caption{Temporal position of behaviors relative to answer commitment in thinking-model traces. All four behaviors typically appear before the answer, suggesting this pattern reflects the general structure of reasoning traces rather than a property unique to calibration.}
\label{tab:temporal}
\setlength{\tabcolsep}{5pt}
\begin{tabular}{lrrrr}
\toprule
Behavior & $N$ & Mean beh.\ pos. & Before answer & Mean gap \\
\midrule
Hypothesis testing         & 266 & 26.7\% & 97.0\% & 65.0 pp \\
Self-correction           & 300 & 34.0\% & 99.0\% & 60.0 pp \\
Uncertainty acknowledgment & 253 & 35.2\% & 95.3\% & 56.2 pp \\
Confidence calibration    & 300 & 34.8\% & 91.0\% & 47.8 pp \\
\bottomrule
\end{tabular}
\end{table}
\begin{table*}[t]
\centering
\scriptsize
\caption{LLM annotation taxonomy: reasoning quality, metacognitive behaviors, and reasoning types.}
\label{tab:llm_taxonomy_1}
\captionsetup[subtable]{skip=0.5ex}
\setlength{\tabcolsep}{3pt}
\renewcommand{\arraystretch}{1.08}

\begin{subtable}[t]{\textwidth}
\centering
\caption{Group 1: Reasoning quality}
\begin{tabularx}{\textwidth}{>{\raggedright\arraybackslash}p{2.5cm} X X X}
\toprule
\textbf{Field} & \textbf{Definition} & \textbf{True if} & \textbf{False if} \\
\midrule
reasoning\_present & Is there actual step-by-step reasoning rather than a direct answer? & Shows intermediate steps, explanations, or deductions & Jumps directly to the answer with no reasoning shown \\[0.3em]
logical\_steps\_valid & Do the reasoning steps logically follow from each other? & Each step follows logically from previous steps & Contains non-sequiturs, invalid inferences, circular reasoning, or logical gaps \\[0.3em]
reaches\_correct\_ \\ conclusion & Does the CoT reasoning lead to the correct answer? & Final answer matches ground truth & Final answer does not match ground truth \\[0.3em]
context\_understanding & Does it correctly understand and use information from the question or context? & Correctly interprets the question, extracts relevant information, and uses correct values & Misreads the question, misinterprets information, or uses wrong values \\
\bottomrule
\end{tabularx}
\end{subtable}

\vspace{0.8ex}

\begin{subtable}[t]{\textwidth}
\centering
\caption{Group 2: Metacognitive behaviors}
\begin{tabularx}{\textwidth}{>{\raggedright\arraybackslash}p{2.5cm} X X X}
\toprule
\textbf{Field} & \textbf{Definition} & \textbf{True if} & \textbf{False if} \\
\midrule
planning\_present & Does it break the problem into sub-steps or make a plan upfront? & Explicit problem decomposition or plan before solving & Solves directly without structure \\[0.3em]
hypothesis\_testing & Does it propose multiple possibilities and test them? & Considers alternatives or explicitly checks different possibilities & Commits to the first interpretation without considering alternatives \\[0.3em]
self\_correction & Does it revise or correct earlier claims? & Explicitly revises or corrects an earlier statement & Never revisits or corrects earlier statements \\[0.3em]
uncertainty\_\\acknowledgment & Does it explicitly note ambiguity or uncertainty? & Explicitly acknowledges ambiguity or uncertainty & States everything with full certainty regardless of ambiguity \\[0.3em]
evidence\_citation & Does it reference specific parts of the question or context to support claims? & Grounds claims in specific information from the prompt or context & Makes claims without connecting them to specific evidence \\[0.3em]
confidence\_calibration & Does the model's confidence match the certainty of its reasoning? & Uncertain when reasoning is weak; confident when reasoning is strong & Overconfident with weak reasoning, or underconfident with strong reasoning \\[0.3em]
self\_awareness & Does it recognize the limits of what can be determined from given information? & Notes that some conclusion cannot be determined from the available information & Claims certainty about things that cannot be determined \\[0.3em]
goal\_tracking & Does it maintain and reference sub-goals in multi-step problems? & Tracks progress toward intermediate goals & Loses track of the original goal or solves the wrong sub-problem \\[0.3em]
knowledge\_alignment & Does it invoke appropriate domain knowledge to solve the problem? & Applies relevant mathematical, scientific, or logical concepts & Manipulates symbols without domain understanding, or applies the wrong knowledge \\
\bottomrule
\end{tabularx}
\end{subtable}

\vspace{0.8ex}

\begin{subtable}[t]{\textwidth}
\centering
\caption{Group 3: Reasoning types}
\begin{tabularx}{\textwidth}{>{\raggedright\arraybackslash}p{2.5cm} X X X}
\toprule
\textbf{Field} & \textbf{Definition} & \textbf{True if} & \textbf{False if} \\
\midrule
mathematical\_reasoning & Involves calculations, equations, formulas, or numerical operations & Uses arithmetic, algebra, geometry, statistics, or proofs & No mathematical operations are present \\[0.3em]
logical\_reasoning & Involves deductive or inductive logic, formal reasoning & Uses if-then reasoning, deduction, syllogisms, or truth-table style logic & No formal logical operations are present \\[0.3em]
causal\_reasoning & Analyzes cause-effect relationships & Explains mechanisms, causes, or effects & No causal analysis \\[0.3em]
analogical\_reasoning & Makes comparisons or draws analogies & Uses analogy or cross-domain comparison & No analogies or comparisons \\[0.3em]
procedural\_reasoning & Follows or describes step-by-step processes or algorithms & Executes sequential operations or procedures & No procedural steps \\[0.3em]
factual\_recall & Relies primarily on retrieving factual knowledge & Uses definitions, historical facts, or established knowledge without derivation & No factual retrieval is needed \\[0.3em]
conceptual\_reasoning & Reasons about abstract concepts and their relationships & Relates concepts or reasons about abstract ideas & No abstract conceptual work \\
\bottomrule
\end{tabularx}
\end{subtable}
\end{table*}

\begin{table*}[t]
\centering
\scriptsize
\caption{LLM annotation taxonomy: failure modes and summary metrics.}
\label{tab:llm_taxonomy_2}
\captionsetup[subtable]{skip=0.5ex}
\setlength{\tabcolsep}{3pt}
\renewcommand{\arraystretch}{1.08}

\begin{subtable}[t]{\textwidth}
\centering
\caption{Group 4: Failure modes}
\begin{tabularx}{\textwidth}{>{\raggedright\arraybackslash}p{2.5cm} X X X}
\toprule
\textbf{Field} & \textbf{Definition} & \textbf{True if (failure occurred)} & \textbf{False if (no failure)} \\
\midrule
factual\_error & Contains incorrect factual claims, knowledge errors, or calculation mistakes & Includes wrong facts, wrong definitions, wrong formulas, or math errors & All facts and calculations are correct \\[0.3em]
logical\_failure & Reasoning steps do not logically connect or contain errors & Contains invalid inferences, contradictions, or non-sequiturs & All logical steps are valid \\[0.3em]
context\_misread & Misunderstands or misinterprets the question or given information & Answers the wrong question, misreads values, or misinterprets constraints & Correctly understands the question and context \\[0.3em]
knowledge\_gap & Lacks necessary domain knowledge to solve the problem & Does not know required concepts, formulas, or facts & Has the necessary domain knowledge \\[0.3em]
post\_hoc\_rationalization & Appears to give the answer first, then make up reasoning & Reasoning seems reverse-engineered and does not support the conclusion & Reasoning genuinely leads to the conclusion \\[0.3em]
shortcut & Skips important logical or computational steps & Jumps to the conclusion without necessary intermediate reasoning or calculation & All necessary steps are shown \\[0.3em]
lucky\_guess & Reaches the correct answer but the reasoning is wrong or irrelevant & Correct answer with invalid or irrelevant reasoning path & Correct answer with valid supporting reasoning \\
\bottomrule
\end{tabularx}
\end{subtable}

\vspace{0.8ex}

\begin{subtable}[t]{\textwidth}
\centering
\caption{Group 5: Summary metrics}
\begin{tabularx}{\textwidth}{>{\raggedright\arraybackslash}p{2.6cm} >{\raggedright\arraybackslash}p{3.4cm} X}
\toprule
\textbf{Field} & \textbf{Values} & \textbf{Definition} \\
\midrule
complexity\_score & 1--5 & 1 = single-step inference; 2 = sequential 2--3 steps; 3 = multi-step with dependencies; 4 = hierarchical decomposition with sub-problems; 5 = complex reasoning with hypothesis testing and self-correction \\[0.3em]
reasoning\_direction & forward / backward / mixed & forward = starts from given information and works toward the answer; backward = starts from answer options and works backward to justify; mixed = combines both approaches \\[0.3em]
efficiency & concise / verbose / insufficient & concise = appropriate length with necessary steps; verbose = unnecessarily long or repetitive; insufficient = too short and missing key steps \\[0.3em]
overall\_quality & strong / flawed / poor & strong = sound reasoning, accurate logic, and coherent chain to the correct conclusion; flawed = reasoning present with notable issues; poor = major failures, severe errors, or no real reasoning \\[0.3em]
would\_make\_good\_judge & true / false & Whether the response would be trustworthy for judging other models' reasoning \\
\bottomrule
\end{tabularx}
\end{subtable}
\end{table*}
\begin{table*}[t]
\centering
\scriptsize
\caption{VLM annotation taxonomy: visual grounding, reasoning quality, and advanced/metacognitive behaviors.}
\label{tab:vlm_taxonomy_1}
\captionsetup[subtable]{skip=0.5ex}
\setlength{\tabcolsep}{3pt}
\renewcommand{\arraystretch}{1.08}

\begin{subtable}[t]{\textwidth}
\centering
\caption{Group 1: Visual grounding}
\begin{tabularx}{\textwidth}{>{\raggedright\arraybackslash}p{2.5cm} X X X}
\toprule
\textbf{Field} & \textbf{Definition} & \textbf{True if} & \textbf{False if} \\
\midrule
visual\_references\_\\present & Does the CoT explicitly mention specific visual elements? & Mentions concrete visual details such as objects, colors, counts, or positions & Generic reasoning with no visual specifics \\[0.3em]
visual\_claims\_accurate & Are all visual descriptions factually correct? & All mentioned visual details match the image & Hallucinated objects, wrong colors, incorrect counts, or wrong spatial relations \\[0.3em]
visual\_input\_necessary & Could this question be answered without seeing the image? & Requires specific visual details to answer & Could be answered from question text alone or general knowledge \\
\bottomrule
\end{tabularx}
\end{subtable}

\vspace{0.8ex}

\begin{subtable}[t]{\textwidth}
\centering
\caption{Group 2: Reasoning quality}
\begin{tabularx}{\textwidth}{>{\raggedright\arraybackslash}p{2.5cm} X X X}
\toprule
\textbf{Field} & \textbf{Definition} & \textbf{True if} & \textbf{False if} \\
\midrule
reasoning\_present & Is there actual step-by-step reasoning rather than a direct answer? & Shows intermediate steps, explanations, or deductions & Jumps directly to the answer with no reasoning shown \\[0.3em]
logical\_steps\_valid & Do the reasoning steps logically follow from each other? & Steps are coherent and valid & Contains non-sequiturs, invalid inferences, circularity, or logical gaps \\[0.3em]
reaches\_correct\_\\conclusion & Does the CoT lead to the correct answer? & Final answer matches ground truth & Final answer is wrong \\
\bottomrule
\end{tabularx}
\end{subtable}

\vspace{0.8ex}

\begin{subtable}[t]{\textwidth}
\centering
\caption{Group 3: Advanced and metacognitive behaviors}
\begin{tabularx}{\textwidth}{>{\raggedright\arraybackslash}p{2.5cm} X X X}
\toprule
\textbf{Field} & \textbf{Definition} & \textbf{True if} & \textbf{False if} \\
\midrule
planning\_present & Does it break the problem into sub-steps or make a plan upfront? & Explicit decomposition or plan before solving & Solves directly without structure \\[0.3em]
hypothesis\_testing & Does it propose multiple possibilities and test them? & Considers alternatives or checks competing interpretations & Commits to the first interpretation \\[0.3em]
self\_correction & Does it revise or correct earlier claims? & Explicitly revises or retracts an earlier step & Never revisits earlier claims \\[0.3em]
uncertainty\_\\acknowledgment & Does it explicitly note ambiguity or uncertainty? & Notes uncertainty or ambiguity & States everything with full certainty \\[0.3em]
evidence\_citation & Does it reference specific visual evidence to support claims? & Grounds claims in specific image details & Makes claims without visual support \\[0.3em]
confidence\_calibration & Does confidence match reasoning strength? & Uncertain when reasoning is weak; confident when reasoning is strong & Overconfident with weak reasoning, or underconfident with strong reasoning \\[0.3em]
self\_awareness & Does it recognize the limits of what can be determined from the image? & Notes that some conclusion cannot be determined from available visual information & Claims certainty beyond what the image supports \\[0.3em]
goal\_tracking & Does it maintain and reference sub-goals in multi-step problems? & Tracks progress toward intermediate goals & Loses track of the original objective \\[0.3em]
knowledge\_alignment & Does it invoke appropriate domain schemas to interpret visual content? & Applies relevant domain concepts to solve the problem & Only describes surface features or applies the wrong domain knowledge \\
\bottomrule
\end{tabularx}
\end{subtable}
\end{table*}

\begin{table*}[t]
\centering
\scriptsize
\caption{VLM annotation taxonomy: reasoning types, failure modes, and summary metrics.}
\label{tab:vlm_taxonomy_2}
\captionsetup[subtable]{skip=0.5ex}
\setlength{\tabcolsep}{3pt}
\renewcommand{\arraystretch}{1.08}

\begin{subtable}[t]{\textwidth}
\centering
\caption{Group 4: Reasoning types}
\begin{tabularx}{\textwidth}{>{\raggedright\arraybackslash}p{2.5cm} X X X}
\toprule
\textbf{Field} & \textbf{Definition} & \textbf{True if} & \textbf{False if} \\
\midrule
spatial\_reasoning & Analyzes position, layout, or relative location & Reasons about left/right, above/below, containment, distance, or orientation & No spatial analysis \\[0.3em]
counting\_reasoning & Involves quantitative analysis or enumeration & Counts objects, compares quantities, or estimates numerically & No counting or quantity comparison \\[0.3em]
compositional\_reasoning & Combines multiple attributes or objects & Integrates several visual attributes or elements & Relies on a single attribute only \\[0.3em]
causal\_reasoning & Analyzes cause-effect relationships & Explains mechanisms or effects & No causal analysis \\[0.3em]
mathematical\_reasoning & Involves calculations, equations, charts, or graphs & Uses arithmetic, geometry, or quantitative interpretation & No mathematical operations \\[0.3em]
temporal\_reasoning & Involves sequence, time, or temporal relations & Reasons about before/after, order, stages, or prediction over time & No temporal analysis \\
\bottomrule
\end{tabularx}
\end{subtable}

\vspace{0.8ex}

\begin{subtable}[t]{\textwidth}
\centering
\caption{Group 5: Failure modes}
\begin{tabularx}{\textwidth}{>{\raggedright\arraybackslash}p{2.5cm} X X X}
\toprule
\textbf{Field} & \textbf{Definition} & \textbf{True if} & \textbf{False if} \\
\midrule
visual\_hallucination & Claims objects or attributes not present in the image & Mentions nonexistent objects, colors, text, or features & All visual claims are accurate \\[0.3em]
visual\_neglect & Misses critical visual details that are present & Overlooks key information needed for the answer & Attends to relevant visual information \\[0.3em]
logical\_failure & Reasoning steps do not connect or contain errors & Invalid inferences, contradictions, or non-sequiturs & All logical steps are valid \\[0.3em]
language\_bias & Over-relies on text priors instead of visual content & Answer follows stereotypes or textual priors rather than image evidence & Reasoning is grounded in the image \\[0.3em]
post\_hoc\_rationalization & Appears to give the answer first, then invent reasoning & Reasoning seems reverse-engineered and does not support the conclusion & Reasoning genuinely leads to the conclusion \\[0.3em]
shortcut & Skips important reasoning steps or bypasses visual analysis & Jumps to the conclusion without needed intermediate steps & Complete reasoning chain is present \\[0.3em]
lucky\_guess & Reaches the correct answer with wrong or irrelevant reasoning & Correct answer but invalid supporting reasoning & Correct answer with valid supporting reasoning \\
\bottomrule
\end{tabularx}
\end{subtable}

\vspace{0.8ex}

\begin{subtable}[t]{\textwidth}
\centering
\caption{Group 6: Summary metrics}
\begin{tabularx}{\textwidth}{>{\raggedright\arraybackslash}p{2.6cm} >{\raggedright\arraybackslash}p{3.4cm} X}
\toprule
\textbf{Field} & \textbf{Values} & \textbf{Definition} \\
\midrule
complexity\_score & 1--5 & 1 = single-step inference; 2 = sequential 2--3 steps; 3 = multi-step with dependencies; 4 = hierarchical with sub-problems; 5 = complex reasoning with hypothesis testing and corrections \\[0.3em]
reasoning\_direction & forward / backward / mixed & forward = starts from given information and works toward the answer; backward = starts from answer options and works backward; mixed = combines both \\[0.3em]
efficiency & concise / verbose / insufficient & concise = appropriate length with necessary steps; verbose = unnecessarily long or repetitive; insufficient = too short and missing key steps \\[0.3em]
overall\_quality & strong / flawed / poor & strong = sound reasoning and accurate grounding; flawed = reasoning present but with notable issues; poor = major failures or no real reasoning \\[0.3em]
would\_make\_good\_judge & true / false & Whether this model's reasoning would be trustworthy for judging other models' responses \\
\bottomrule
\end{tabularx}
\end{subtable}
\end{table*}

\definecolor{cardterracotta}{HTML}{C0785C}
\definecolor{cardbg}{HTML}{FDFCFA}
\definecolor{cardborder}{HTML}{D8CFC6}
\definecolor{cardtext}{HTML}{3A3A3A}
\definecolor{cardmuted}{HTML}{8A8A8A}
\definecolor{groupbg}{HTML}{F3EEEA}

\newtcolorbox{promptcard}[1]{
  enhanced,
  colback=cardbg,
  colframe=cardborder,
  coltext=cardtext,
  boxrule=0.6pt,
  arc=2pt,
  left=8pt, right=8pt, top=6pt, bottom=6pt,
  fonttitle=\sffamily\bfseries\color{cardterracotta},
  title={#1},
  coltitle=cardterracotta,
  attach boxed title to top left={yshift=-2mm, xshift=4mm},
  boxed title style={colback=white, colframe=cardborder, boxrule=0.4pt, arc=1.5pt},
}

\newcommand{\promptsection}[1]{%
  \vspace{6pt}%
  \noindent\colorbox{groupbg}{\parbox{\dimexpr\linewidth-2\fboxsep}{%
    \smallskip\sffamily\textcolor{cardterracotta}{\textbf{#1}}\smallskip%
  }}%
  \vspace{4pt}\par%
}

\clearpage
\phantomsection
\label{app:llm_prompt}
\begin{promptcard}{LLM Behavioral Annotation Prompt}

\small

You are evaluating the chain-of-thought reasoning quality of a text-only language model. Analyze the model's reasoning process using the taxonomy below. Be strict and objective.

\promptsection{Group 1: Reasoning Quality}

\texttt{reasoning\_present}: Is there actual step-by-step reasoning rather than a direct answer?
\texttt{logical\_steps\_valid}: Do the reasoning steps logically follow from each other?
\texttt{reaches\_correct\_conclusion}: Does the CoT reasoning lead to the correct answer?
\texttt{context\_understanding}: Does it correctly understand and use information from the question or context?

\promptsection{Group 2: Shared Higher-Order Behaviors}

\texttt{planning\_present}: Does it break the problem into sub-steps or make a plan upfront?
\texttt{hypothesis\_testing}: Does it propose multiple possibilities and test them?
\texttt{self\_correction}: Does it revise or correct earlier claims?
\texttt{uncertainty\_\\acknowledgment}: Does it explicitly note ambiguity or uncertainty?
\texttt{evidence\_citation}: Does it reference specific parts of the question or context to support claims?
\texttt{confidence\_calibration}: Does the model's confidence match the certainty of its reasoning?
\texttt{self\_awareness}: Does it recognize the limits of what can be determined from given information?
\texttt{goal\_tracking}: Does it maintain and reference sub-goals in multi-step problems?
\texttt{knowledge\_alignment}: Does it invoke appropriate domain knowledge to solve the problem?

\promptsection{Group 3: Reasoning Types}

\texttt{mathematical\_reasoning},
\texttt{logical\_reasoning},
\texttt{causal\_reasoning},
\texttt{analogical\_reasoning},
\texttt{procedural\_reasoning},
\texttt{factual\_recall},
\texttt{conceptual\_reasoning}.
Multiple can be true.

\promptsection{Group 4: Failure Modes}

\texttt{factual\_error}: Incorrect factual claims or calculation mistakes.
\texttt{logical\_failure}: Reasoning steps don't logically connect.
\texttt{context\_misread}: Misunderstands the question or given information.
\texttt{knowledge\_gap}: Lacks necessary domain knowledge.
\texttt{post\_hoc\_rationalization}: Reasoning appears reverse-engineered from the answer.
\texttt{shortcut}: Skips important logical or computational steps.
\texttt{lucky\_guess}: Reaches correct answer but reasoning is wrong.
For all failure modes, \texttt{true} means the failure occurred.

\promptsection{Group 5: Summary Metrics}

\texttt{complexity\_score} (1--5), \texttt{reasoning\_direction} (forward/backward/mixed), \texttt{efficiency} (concise/verbose/insufficient), \texttt{overall\_quality} (strong/flawed/poor), \texttt{would\_make\_good\_judge} (true/false).

\vspace{4pt}
{\color{cardmuted}\rule{\linewidth}{0.3pt}}
\vspace{2pt}

{\small\color{cardmuted} Evaluate each metric independently. Be strict: only mark true if clearly demonstrated. For failure modes, false means the failure did not occur. Respond with valid JSON only.}

\end{promptcard}

\clearpage

\begin{promptcard}{LLM Annotation Output Format}

\small
\ttfamily
\raggedright

\{\\
\quad "evaluation": \{\\
\quad\quad "reasoning\_quality": \{\\
\quad\quad\quad "reasoning\_present": true|false,\\
\quad\quad\quad "logical\_steps\_valid": true|false,\\
\quad\quad\quad "reaches\_correct\_conclusion": true|false,\\
\quad\quad\quad "context\_understanding": true|false\\
\quad\quad \},\\[4pt]
\quad\quad "advanced\_and\_metacognitive": \{\\
\quad\quad\quad "planning\_present": true|false,\\
\quad\quad\quad "hypothesis\_testing": true|false,\\
\quad\quad\quad "self\_correction": true|false,\\
\quad\quad\quad "uncertainty\_acknowledgment": true|false,\\
\quad\quad\quad "evidence\_citation": true|false,\\
\quad\quad\quad "confidence\_calibration": true|false,\\
\quad\quad\quad "self\_awareness": true|false,\\
\quad\quad\quad "goal\_tracking": true|false,\\
\quad\quad\quad "knowledge\_alignment": true|false\\
\quad\quad \},\\[4pt]
\quad\quad "reasoning\_types": \{\\
\quad\quad\quad "mathematical\_reasoning": true|false,\\
\quad\quad\quad "logical\_reasoning": true|false,\\
\quad\quad\quad "causal\_reasoning": true|false,\\
\quad\quad\quad "analogical\_reasoning": true|false,\\
\quad\quad\quad "procedural\_reasoning": true|false,\\
\quad\quad\quad "factual\_recall": true|false,\\
\quad\quad\quad "conceptual\_reasoning": true|false\\
\quad\quad \},\\[4pt]
\quad\quad "failure\_modes": \{\\
\quad\quad\quad "factual\_error": true|false,\\
\quad\quad\quad "logical\_failure": true|false,\\
\quad\quad\quad "context\_misread": true|false,\\
\quad\quad\quad "knowledge\_gap": true|false,\\
\quad\quad\quad "post\_hoc\_rationalization": true|false,\\
\quad\quad\quad "shortcut": true|false,\\
\quad\quad\quad "lucky\_guess": true|false\\
\quad\quad \},\\[4pt]
\quad\quad "summary\_metrics": \{\\
\quad\quad\quad "complexity\_score": 1--5,\\
\quad\quad\quad "reasoning\_direction": "forward"|"backward"|"mixed",\\
\quad\quad\quad "efficiency": "concise"|"verbose"|"insufficient",\\
\quad\quad\quad "overall\_quality": "strong"|"flawed"|"poor",\\
\quad\quad\quad "would\_make\_good\_judge": true|false\\
\quad\quad \}\\
\quad \},\\[4pt]
\quad "evaluator\_notes": "Brief justification for key ratings"\\
\}

\end{promptcard}

\clearpage

\phantomsection
\label{app:vlm_prompt}
\begin{promptcard}{VLM Behavioral Annotation Prompt}
\small

You are evaluating the chain-of-thought reasoning quality of a vision-language model. Analyze the model's reasoning process using the taxonomy below. Be strict and objective.

\promptsection{Group 1: Visual Grounding}

\texttt{visual\_references\_present}: Does the CoT explicitly mention specific visual elements?
\texttt{visual\_claims\_accurate}: Are all visual descriptions factually correct?
\texttt{visual\_input\_necessary}: Could this question be answered without seeing the image?

\promptsection{Group 2: Reasoning Quality}

\texttt{reasoning\_present}: Is there actual step-by-step reasoning rather than a direct answer?
\texttt{logical\_steps\_valid}: Do the reasoning steps logically follow from each other?
\texttt{reaches\_correct\_conclusion}: Does the CoT reasoning lead to the correct answer?

\promptsection{Group 3: Shared Higher-Order Behaviors}

\texttt{planning\_present}: Does it break the problem into sub-steps or make a plan upfront?
\texttt{hypothesis\_testing}: Does it propose multiple possibilities and test them?
\texttt{self\_correction}: Does it revise or correct earlier claims?
\texttt{uncertainty\_acknowledgment}: Does it explicitly note ambiguity or uncertainty?
\texttt{evidence\_citation}: Does it reference specific visual evidence to support claims?
\texttt{confidence\_calibration}: Does the model's confidence match the certainty of its reasoning?
\texttt{self\_awareness}: Does it recognize the limits of what can be determined from the image?
\texttt{goal\_tracking}: Does it maintain and reference sub-goals in multi-step problems?
\texttt{knowledge\_alignment}: Does it invoke appropriate domain schemas to interpret visual content?

\promptsection{Group 4: Reasoning Types}

\texttt{spatial\_reasoning},
\texttt{counting\_reasoning},
\texttt{compositional\_reasoning},
\texttt{causal\_reasoning},
\texttt{mathematical\_reasoning},
\texttt{temporal\_reasoning}.
Multiple can be true.

\promptsection{Group 5: Failure Modes}

\texttt{visual\_hallucination}: Claims objects or attributes not present in the image.
\texttt{visual\_neglect}: Misses critical visual details that are present.
\texttt{logical\_failure}: Reasoning steps don't logically connect.
\texttt{language\_bias}: Over-relies on text priors, ignores actual visual content.
\texttt{post\_hoc\_rationalization}: Reasoning appears reverse-engineered from the answer.
\texttt{shortcut}: Skips important reasoning steps or bypasses visual analysis.
\texttt{lucky\_guess}: Reaches correct answer but reasoning is wrong.
For all failure modes, \texttt{true} means the failure occurred.

\promptsection{Group 6: Summary Metrics}

\texttt{complexity\_score} (1--5), \texttt{reasoning\_direction} (forward/backward/mixed), \texttt{efficiency} (concise/verbose/insufficient), \texttt{overall\_quality} (strong/flawed/poor), \texttt{would\_make\_good\_judge} (true/false).

\vspace{4pt}
{\color{cardmuted}\rule{\linewidth}{0.3pt}}
\vspace{2pt}

{\small\color{cardmuted} Evaluate each metric independently. Be strict: only mark true if clearly demonstrated. For failure modes, false means the failure did not occur. Respond with valid JSON only.}

\end{promptcard}

\clearpage

\begin{promptcard}{VLM Annotation Output Format}

\small
\ttfamily
\raggedright

\{\\
\quad "evaluation": \{\\
\quad\quad "visual\_grounding": \{\\
\quad\quad\quad "visual\_references\_present": true|false,\\
\quad\quad\quad "visual\_claims\_accurate": true|false,\\
\quad\quad\quad "visual\_input\_necessary": true|false\\
\quad\quad \},\\[4pt]
\quad\quad "reasoning\_quality": \{\\
\quad\quad\quad "reasoning\_present": true|false,\\
\quad\quad\quad "logical\_steps\_valid": true|false,\\
\quad\quad\quad "reaches\_correct\_conclusion": true|false\\
\quad\quad \},\\[4pt]
\quad\quad "advanced\_and\_metacognitive": \{\\
\quad\quad\quad "planning\_present": true|false,\\
\quad\quad\quad "hypothesis\_testing": true|false,\\
\quad\quad\quad "self\_correction": true|false,\\
\quad\quad\quad "uncertainty\_acknowledgment": true|false,\\
\quad\quad\quad "evidence\_citation": true|false,\\
\quad\quad\quad "confidence\_calibration": true|false,\\
\quad\quad\quad "self\_awareness": true|false,\\
\quad\quad\quad "goal\_tracking": true|false,\\
\quad\quad\quad "knowledge\_alignment": true|false\\
\quad\quad \},\\[4pt]
\quad\quad "reasoning\_types": \{\\
\quad\quad\quad "spatial\_reasoning": true|false,\\
\quad\quad\quad "counting\_reasoning": true|false,\\
\quad\quad\quad "compositional\_reasoning": true|false,\\
\quad\quad\quad "causal\_reasoning": true|false,\\
\quad\quad\quad "mathematical\_reasoning": true|false,\\
\quad\quad\quad "temporal\_reasoning": true|false\\
\quad\quad \},\\[4pt]
\quad\quad "failure\_modes": \{\\
\quad\quad\quad "visual\_hallucination": true|false,\\
\quad\quad\quad "visual\_neglect": true|false,\\
\quad\quad\quad "logical\_failure": true|false,\\
\quad\quad\quad "language\_bias": true|false,\\
\quad\quad\quad "post\_hoc\_rationalization": true|false,\\
\quad\quad\quad "shortcut": true|false,\\
\quad\quad\quad "lucky\_guess": true|false\\
\quad\quad \},\\[4pt]
\quad\quad "summary\_metrics": \{\\
\quad\quad\quad "complexity\_score": 1--5,\\
\quad\quad\quad "reasoning\_direction": "forward"|"backward"|"mixed",\\
\quad\quad\quad "efficiency": "concise"|"verbose"|"insufficient",\\
\quad\quad\quad "overall\_quality": "strong"|"flawed"|"poor",\\
\quad\quad\quad "would\_make\_good\_judge": true|false\\
\quad\quad \}\\
\quad \},\\[4pt]
\quad "evaluator\_notes": "Brief justification for key ratings"\\
\}

\end{promptcard}
\newtcolorbox{samplecard}[2]{
  enhanced,
  colback=cardbg,
  colframe=cardborder,
  coltext=cardtext,
  boxrule=0.6pt,
  arc=2pt,
  left=8pt, right=8pt, top=6pt, bottom=6pt,
  fonttitle=\sffamily\color{cardterracotta},
  title={\textbf{#1} \hfill \normalsize\textsf{\textcolor{cardgreen}{#2}}},
  coltitle=cardterracotta,
  attach boxed title to top left={yshift=-2mm, xshift=4mm},
  boxed title style={colback=white, colframe=cardborder, boxrule=0.4pt, arc=1.5pt,
    right=4pt, left=4pt},
}

\newcommand{\samplefooter}[3]{%
  \vspace{4pt}%
  {\color{cardmuted}\rule{\linewidth}{0.3pt}}%
  \vspace{2pt}%
  \par{\small\sffamily\color{cardmuted}%
    \textbf{Model:} #1 \hfill \textbf{Benchmark:} #2}%

}

\label{confidence_samples}
\begin{samplecard}{Confidence Calibration}{\checkmark\; Correct}
\label{card:first}

\small

\textsf{\textcolor{cardmuted}{\textbf{Question:}}}
If $2^8=4^x$, what is the value of $x$?

\vspace{6pt}
\textsf{\textcolor{cardmuted}{\textbf{Model trace:}}}

\vspace{2pt}
First, I remember that $4$ is a power of $2$. Specifically, $4=2^2$.
So
\[
4^x=(2^2)^x=2^{2x}.
\]
Setting $8=2x$ gives $x=4$.

Let me check that. If $x=4$, then $4^4=256$, and $2^8=256$.
\colorbox{groupbg}{\textcolor{cardterracotta}{Yep, that works.}}

Wait, let me make sure I didn't make a mistake. Let's verify step by step:
\[
4^4 = 4 \times 4 \times 4 \times 4 = 256.
\]
So both sides are $256$ when $x=4$.
\colorbox{groupbg}{\textcolor{cardterracotta}{That checks out.}}

Another way to think about it is to take logarithms:
\[
\log_2(2^8)=\log_2(4^x).
\]
The left side is $8$, and the right side is
\[
x\log_2(4)=2x.
\]
So again $8=2x$, hence $x=4$.

\colorbox{groupbg}{\textcolor{cardterracotta}{I think that's solid.}}
The key was recognizing that $4$ is a power of $2$, then confirming the same answer through direct substitution and logarithms.
\colorbox{groupbg}{\textcolor{cardterracotta}{No steps seem to be missing here.}}

\textbf{Final answer: $4$} \hfill \textit{(Groundtruth: \textbf{$4$})}

\vspace{4pt}
{\color{cardmuted}\rule{\linewidth}{0.3pt}}
\vspace{2pt}

{\small\sffamily
\colorbox{groupbg}{\textbf{Note:}}The model's certainty grows with converging support: it derives $x=4$, verifies by substitution, and confirms through an equivalent logarithmic formulation before committing to the answer.}

\vspace{4pt}

\samplefooter{Qwen3-4B-Think}{MATH-500}{77}

\end{samplecard}

\begin{samplecard}{Confidence Calibration}{\checkmark\; Correct}

\small

\textsf{\textcolor{cardmuted}{\textbf{Question:}}}
If a snack-size tin of peaches has $40$ calories and is $2\%$ of a person's
daily caloric requirement, how many calories fulfill a person's daily caloric requirement?

\vspace{6pt}
\textsf{\textcolor{cardmuted}{\textbf{Model trace:}}}

\vspace{2pt}
Let the daily caloric requirement be $x$. Since $2\%$ of $x$ equals $40$,
\[
0.02x = 40.
\]
Dividing by $0.02$ gives
\[
x = \frac{40}{0.02} = 2000.
\]

The snack has $40$ calories and that is supposed to be $2\%$ of the daily need.
To recover the full amount, divide the part by the percentage:
\[
40 \div 0.02 = 2000.
\]

Maybe I can convert $0.02$ into a fraction.
Since
\[
0.02 = \frac{2}{100} = \frac{1}{50},
\]
dividing by $\frac{1}{50}$ is the same as multiplying by $50$:
\[
40 \times 50 = 2000.
\]
\colorbox{groupbg}{\textcolor{cardterracotta}{That seems right.}}

Wait, let me check another way.
If $2\%$ is $40$, then $1\%$ is $20$.
So $100\%$ is
\[
20 \times 100 = 2000.
\]
\colorbox{groupbg}{\textcolor{cardterracotta}{Yep, that matches.}}
\colorbox{groupbg}{\textcolor{cardterracotta}{So both methods give me $2000$. That makes me more confident.}}

\textbf{Final answer: $2000$} \hfill \textit{(Groundtruth: \textbf{$2000$})}

\vspace{4pt}
{\color{cardmuted}\rule{\linewidth}{0.3pt}}
\vspace{2pt}

{\small\sffamily
\colorbox{groupbg}{\textbf{Note:}}An especially clear case: the model solves the percentage equation directly, rederives the answer via a fraction conversion, and confirms it with a $1\% \rightarrow 100\%$ argument, explicitly noting that the agreement increases its confidence.}

\vspace{4pt}

\samplefooter{OLMo3-7B-Think}{MATH-500}{38}

\end{samplecard}

\begin{samplecard}{Confidence Calibration}{\checkmark\; Correct}

\small

\textsf{\textcolor{cardmuted}{\textbf{Question:}}}
If $n \equiv 2 \pmod{7}$, find the remainder when
\[
(n+2)(n+4)(n+6)
\]
is divided by $7$.

\vspace{6pt}
\textsf{\textcolor{cardmuted}{\textbf{Model trace:}}}

\vspace{2pt}
Let $n=7k+2$. Then
\[
n+2 \equiv 4 \pmod{7}, \qquad
n+4 \equiv 6 \pmod{7}, \qquad
n+6 \equiv 1 \pmod{7}.
\]
So
\[
(n+2)(n+4)(n+6) \equiv 4 \cdot 6 \cdot 1 = 24 \equiv 3 \pmod{7}.
\]

\colorbox{groupbg}{\textcolor{cardterracotta}{The remainder is $3$.}}

But let me test with a specific value to make sure.
Take $n=2$:
\[
(2+2)(2+4)(2+6)=4\cdot6\cdot8=192.
\]
Since
\[
192 = 7\cdot27 + 3,
\]
the remainder is $3$.
\colorbox{groupbg}{\textcolor{cardterracotta}{That checks out.}}

Another example: take $n=9$, which also satisfies $n\equiv2\pmod7$.
Then
\[
(9+2)(9+4)(9+6)=11\cdot13\cdot15.
\]
Modulo $7$, these are again $4$, $6$, and $1$, so the product has remainder $3$.
Computing directly also gives remainder $3$.
\colorbox{groupbg}{\textcolor{cardterracotta}{Correct again.}}

\colorbox{groupbg}{\textcolor{cardterracotta}{So the solution is correct.}}

\textbf{Final answer: $3$} \hfill \textit{(Groundtruth: \textbf{$3$})}

\vspace{4pt}
{\color{cardmuted}\rule{\linewidth}{0.3pt}}
\vspace{2pt}

{\small\sffamily
\colorbox{groupbg}{\textbf{Note:}} The model derives the remainder symbolically, then checks two concrete substitutions that independently recover the same result; only after both checks agree does it affirm the answer.}

\vspace{4pt}

\samplefooter{Qwen3-4B-Think}{MATH-500}{35}

\end{samplecard}


\definecolor{uncbg}{HTML}{E8EDF2}      
\definecolor{unctext}{HTML}{5B7D99}    

\begin{samplecard}{Uncertainty Acknowledgment + Confidence Calibration}{\checkmark\; Correct}

\small

\textsf{\textcolor{cardmuted}{\textbf{Question:}}}
Convert the point $(0,3)$ in rectangular coordinates to polar coordinates. Enter your answer in the form $(r,\theta)$, where $r > 0$ and $0 \le \theta < 2\pi$.

\vspace{6pt}
\textsf{\textcolor{cardmuted}{\textbf{Model trace:}}}

\vspace{2pt}
We know that $r = \sqrt{x^2 + y^2}$ and $\theta = \tan^{-1}(y/x)$ for $x > 0$. But $x = 0$ in this case, so we can't use the formula for $\theta$ directly.

Instead, we can look at the point $(0,3)$ on the coordinate plane. This point is on the positive $y$-axis. Therefore, $\theta = \pi/2$. [\ldots]

\colorbox{uncbg}{\textcolor{unctext}{But I'm confused}} why the solution says ``we can't use the formula for $\theta$ directly.'' Because the formula $\tan^{-1}(y/x)$ is undefined when $x = 0$, so we can't use it directly. But they still got the right answer.

\colorbox{uncbg}{\textcolor{unctext}{Wait, maybe the solution is correct. Let me think again.}} Another way to think about polar coordinates: $(r, \theta)$ corresponds to the point $(r\cos\theta, r\sin\theta)$. So we want $r\cos\theta = 0$ and $r\sin\theta = 3$. [\ldots]

Now, $\sin\theta = 3/r > 0$ because $r > 0$ and $3 > 0$, so $\theta = \pi/2$. \colorbox{groupbg}{\textcolor{cardterracotta}{Yes, that makes sense.}} So $\theta = \pi/2$ is the correct angle. So the polar coordinates are $(3, \pi/2)$. \colorbox{groupbg}{\textcolor{cardterracotta}{I think the solution is correct.}}

\vspace{4pt}
{\color{cardmuted}\rule{\linewidth}{0.3pt}}
\vspace{2pt}

{\small\sffamily
\colorbox{uncbg}{\textcolor{unctext}{\textbf{~Note:~}}} The model voices confusion about the undefined arctangent formula, then resolves it by rederiving $\theta$ from $r \cos \theta = 0$ and $r \sin \theta = 3$; confidence rises only once the geometric and algebraic views agree.}

\vspace{4pt}

{\small\sffamily\color{cardmuted}%
\textbf{Model:} Qwen3-4B-Think \hfill \textbf{Benchmark:} MATH-500 \hfill \textbf{} }

\end{samplecard}


\begin{samplecard}{Uncertainty Acknowledgment + Confidence Calibration}{\checkmark\; Correct}

\small

\textsf{\textcolor{cardmuted}{\textbf{Question:}}}
According to the above definition of the sucker effect, which of the following options belongs to the straw effect?

\vspace{6pt}
\textsf{\textcolor{cardmuted}{\textbf{Model trace:}}}

\vspace{2pt}
[\ldots] Wait, the answer given is A, but the question asks for ``straw effect'' while the passage talks about ``sucker effect''. \colorbox{uncbg}{\textcolor{unctext}{This is a bit confusing.}}

\colorbox{groupbg}{\textcolor{cardterracotta}{Let me check the Chinese terms to see if there's a translation mix-up.}} In Chinese, ``sucker effect'' is often translated as ``free rider effect''. ``Straw effect'' is not a standard term in this context. \colorbox{uncbg}{\textcolor{unctext}{It's possible that the question meant to ask for the}}
\colorbox{uncbg}{\textcolor{unctext}{``sucker effect'' but wrote ``straw effect'' by mistake.}}

Given the passage defines the ``sucker effect'' [\ldots] Option A describes exactly this [\ldots]

\colorbox{groupbg}{\textcolor{cardterracotta}{Given the context, it's likely that the intended answer is A}}, as it fits the sucker effect described.

\vspace{4pt}
{\color{cardmuted}\rule{\linewidth}{0.3pt}}
\vspace{2pt}

{\small\sffamily
\colorbox{uncbg}{\textcolor{unctext}{\textbf{~Note:~}}}The model flags a likely translation error in the question, states that uncertainty explicitly, and commits to A only after checking that the option fits the definition the passage actually provides.}

\vspace{4pt}

{\small\sffamily\color{cardmuted}%
\textbf{Model:} Qwen3-4B-Think \hfill \textbf{Benchmark:} LogiQA2 \hfill \textbf{Doc ID:} 163}

\end{samplecard}
\definecolor{kalbg}{HTML}{EDE8F2}      
\definecolor{kaltext}{HTML}{8A6A9E}    

\begin{samplecard}{Knowledge Alignment}{\checkmark\; Correct}
\small
\textsf{\textcolor{cardmuted}{\textbf{Question:}}}
Simplify $\dfrac{\sec x}{\sin x} - \dfrac{\sin x}{\cos x}$.

\vspace{6pt}
\textsf{\textcolor{cardmuted}{\textbf{Model trace:}}}
\vspace{2pt}

To simplify the expression, \colorbox{kalbg}{\textcolor{kaltext}{we first recall that $\sec x = \frac{1}{\cos x}$.}} Rewriting each term:
\[
\frac{1}{\cos x \sin x} - \frac{\sin^2 x}{\cos x \sin x} = \frac{1 - \sin^2 x}{\cos x \sin x}.
\]
\colorbox{kalbg}{\textcolor{kaltext}{Now, recall the Pythagorean identity: $\sin^2 x + \cos^2 x = 1$, which implies $1 - \sin^2 x = \cos^2 x$.}} Substituting:
\[
\frac{\cos^2 x}{\cos x \sin x} = \frac{\cos x}{\sin x} = \cot x.
\]

\textbf{Final answer: $\cot x$} \hfill \textit{(Groundtruth: $\cot x$)}

\samplefooter{Qwen3-4B-Think}{MATH-500}{74}
\end{samplecard}

\begin{samplecard}{Knowledge Alignment}{\checkmark\; Correct}
\small
\textsf{\textcolor{cardmuted}{\textbf{Question:}}}
All the water lilies bloom at night, and this lotus blossoms during the day, so this lotus is not a water lily. Which of the following is very similar to the formal structure of the above reasoning?

\vspace{6pt}
\textsf{\textcolor{cardmuted}{\textbf{Model trace:}}}
\vspace{2pt}

The reasoning in the passage is: All water lilies bloom at night, this lotus blossoms during the day, so this lotus is not a water lily. Let's break down the structure:
$A$ = water lilies,\; $B$ = bloom at night,\; $C$ = this lotus.

\[
\text{All } A \text{ are } B \qquad C \text{ is not } B \qquad \therefore C \text{ is not } A
\]

\colorbox{kalbg}{\textcolor{kaltext}{This is a valid syllogism known as ``Denying the Consequent'' or \textit{Modus Tollens} in logic.}}
[\ldots] Option B: All goats have beards, and this sheep doesn't have beards, so this sheep is not a goat. \colorbox{kalbg}{\textcolor{kaltext}{This matches the structure of the passage.}} The other options commit \colorbox{kalbg}{\textcolor{kaltext}{the fallacy of affirming the consequent}}, which is a distinct and invalid form of reasoning.

\textbf{Final answer: B} \hfill \textit{(Groundtruth: \textbf{B})}

\vspace{4pt}
{\color{cardmuted}\rule{\linewidth}{0.3pt}}
\vspace{2pt}

{\small\sffamily
\colorbox{kalbg}{\textcolor{kaltext}{\textbf{~Note:~}}}The trace names Modus Tollens and Affirming the Consequent unprompted, applying an internalized logical framework instead of matching the surface wording of the options.}

\vspace{4pt}

\samplefooter{Qwen3-4B-Think}{LogiQA2}{65}
\end{samplecard}
\begin{samplecard}{Uncertainty Acknowledgment}{\texttimes\; Incorrect}
\small
\textsf{\textcolor{cardmuted}{\textbf{Question:}}}
A figure skater is facing north when she begins to spin to her right. She spins 2250 degrees. Which direction (north, south, east or west) is she facing when she finishes her spin?
\vspace{6pt}
\textsf{\textcolor{cardmuted}{\textbf{Model trace:}}}
\vspace{2pt}
She starts facing north. Spinning to her right means she turns clockwise. A full circle is 360 degrees. So, 2250 divided by 360 equals 6 with a remainder of 90. [\ldots]
Starting from north, turning 90 degrees to the right leads to west. \colorbox{uncbg}{\textcolor{unctext}{Wait, I think I might be confused.}} Let me recheck. If you start facing north and turn clockwise 90 degrees, you face east. \colorbox{uncbg}{\textcolor{unctext}{Wait, no! Wait, I think I messed up the direction of the turn.}} [\ldots] 90 degrees right (clockwise) from north is east. But the solution says west. \colorbox{uncbg}{\textcolor{unctext}{Hmm, that's a problem. Wait, now I'm confused. Which is it?}} [\ldots]
\textbf{Final answer: west} \hfill \textit{(Groundtruth: \textbf{east})}
\samplefooter{Qwen3-4B-Think}{MATH-500}{97}
\end{samplecard}

\begin{samplecard}{Uncertainty Acknowledgment}{\texttimes\; Incorrect}
\small
\textsf{\textcolor{cardmuted}{\textbf{Question:}}}
Given $f(x) = \frac{\sqrt{x-1}}{x-2}$, what is the smallest possible integer value for $x$ such that $f(x)$ has a real number value?
\vspace{6pt}
\textsf{\textcolor{cardmuted}{\textbf{Model trace:}}}
\vspace{2pt}
To ensure $f(x)$ is real, we need $x - 1 \ge 0$ and $x - 2 \neq 0$, giving $x \ge 1$ and $x \neq 2$. The smallest integer satisfying this is $x = 1$. \colorbox{uncbg}{\textcolor{unctext}{But the correct answer is $x = 3$. Why is $x = 1$ not acceptable?}} [\ldots]
When $x = 1$, the denominator is $1 - 2 = -1$, which is nonzero, so $f(1)$ is defined. \colorbox{uncbg}{\textcolor{unctext}{So why is the answer 3? Hmm, maybe there's a mistake in the problem statement or the solution?}} [\ldots] \colorbox{uncbg}{\textcolor{unctext}{Maybe the solution is wrong? Or maybe there's a misunderstanding.}} Wait, maybe the original problem was different [\ldots]
\textbf{Final answer: 3} \hfill \textit{(Groundtruth: \textbf{1})}
\samplefooter{Qwen3-4B-Think}{MATH-500}{91}
\end{samplecard}
\definecolor{sawbg}{HTML}{F2EAD8}      
\definecolor{sawtext}{HTML}{8A5E1A}    

\begin{samplecard}{Self-Awareness}{\checkmark\; Correct}
\small
\textsf{\textcolor{cardmuted}{\textbf{Question:}}}
People generally think that left-handed people are more likely to have operational
accidents than right-handed people, which is a misunderstanding. In fact, most
household accidents --- from fire burns to cut fingers --- involve right-handed
people. Which of the following best summarizes the loophole in the above argument?

\vspace{6pt}
\textsf{\textcolor{cardmuted}{\textbf{Model trace:}}}
\vspace{2pt}

[\ldots] \colorbox{sawbg}{\textcolor{sawtext}{The argument doesn't consider how many left-handed}}
\colorbox{sawbg}{\textcolor{sawtext}{people there are compared to right-handed people.}}
The flaw is that \colorbox{sawbg}{\textcolor{sawtext}{it compares absolute numbers without}}
\colorbox{sawbg}{\textcolor{sawtext}{considering the base rate of each group.}}
For example, suppose 90\% of the population is right-handed. If 50\% of accidents
involve right-handed people, that is actually a \emph{lower} accident rate than
for left-handed people. \colorbox{sawbg}{\textcolor{sawtext}{Even if the absolute count is higher,}}
\colorbox{sawbg}{\textcolor{sawtext}{the proportional rate could still favor left-handers.}}
Option D correctly identifies this missing baseline as the core loophole.

\textbf{Final answer: D} \hfill \textit{(Groundtruth: \textbf{D})}

\vspace{4pt}
{\color{cardmuted}\rule{\linewidth}{0.3pt}}
\vspace{2pt}

{\small\sffamily
\colorbox{sawbg}{\textcolor{sawtext}{\textbf{~Note:~}}} The trace recognizes the limits of what the passage can establish: it flags the missing base-rate data and constructs a concrete counterexample (90\% right-handed) to expose the gap.}

\vspace{4pt}

\samplefooter{Qwen3-4B-Think}{LogiQA2}{40}
\end{samplecard}
\definecolor{ecibg}{HTML}{E2EDE3}      
\definecolor{ecitext}{HTML}{3A6B40}    

\begin{samplecard}{Evidence Citation}{\checkmark\; Correct}
\small
\textsf{\textcolor{cardmuted}{\textbf{Question:}}}
Behavior refers to what people say and do, with more than one measurement
scale, which can be observed, described and recorded [\ldots] It can be
public or hidden. Accordingly, which of the following is \emph{not} behavior?

\vspace{6pt}
\textsf{\textcolor{cardmuted}{\textbf{Model trace:}}}
\vspace{2pt}

Let's analyze each option [\ldots] Option C: Xiao Jun is secretly angry
with his mother in the house. This is a hidden emotion that cannot be
directly observed or recorded.
\colorbox{ecibg}{\textcolor{ecitext}{The passage states that behavior}}
\colorbox{ecibg}{\textcolor{ecitext}{can be public or hidden, but the key point is that}}
\colorbox{ecibg}{\textcolor{ecitext}{behavior must be something that can be observed,}}
\colorbox{ecibg}{\textcolor{ecitext}{described and recorded.}}
Secretly being angry does not meet this criterion because
\colorbox{ecibg}{\textcolor{ecitext}{it is an internal state, not an observable action.}}
Therefore, it is not behavior.

\textbf{Final answer: C} \hfill \textit{(Groundtruth: \textbf{C})}

\vspace{4pt}
{\color{cardmuted}\rule{\linewidth}{0.3pt}}
\vspace{2pt}

{\small\sffamily
\colorbox{ecibg}{\textcolor{ecitext}{\textbf{~Note: ~}}}The trace anchors its reasoning to specific clauses of the definition, applying the stricter observability criterion to defeat the trap set by the 'public or hidden' clause.}

\vspace{4pt}

\samplefooter{Qwen3-4B-Think}{LogiQA2}{21}
\end{samplecard}
\definecolor{htbg}{HTML}{EDE8F2}      
\definecolor{httext}{HTML}{7A4F8A}    

\label{hypothesis_samples}
\begin{samplecard}{Hypothesis Testing}{\checkmark\; Correct}
\small
\textsf{\textcolor{cardmuted}{\textbf{Question:}}}
Let $x$, $y$, and $z$ be positive real numbers such that $xyz = 2$.
Find the minimum value of $x^4 + 4y^2 + 4z^4$.

\vspace{6pt}
\textsf{\textcolor{cardmuted}{\textbf{Model trace:}}}
\vspace{2pt}

We can write the expression as $x^4 + 4y^2 + 4z^4$ and apply AM-GM [\ldots]
\colorbox{htbg}{\textcolor{httext}{Alternative approach: apply AM-GM to $x^4$, $4$, $4y^2$, $4z^4$.}}\\
\colorbox{htbg}{\textcolor{httext}{But this does not give the minimum because $4$ is constant}}\\
\colorbox{htbg}{\textcolor{httext}{and we don't know if the minimum occurs when $x^4 = 4$.}} \\
\colorbox{htbg}{\textcolor{httext}{Alternative approach: apply AM-GM to $x^4$, $x^4$, $x^4$,}}
\colorbox{htbg}{\textcolor{httext}{$y^2$, $y^2$, $z^4$, $z^4$, $z^4$.}}
Let's check the product [\ldots]
\colorbox{htbg}{\textcolor{httext}{This is not constant, so it doesn't work.}}
\colorbox{htbg}{\textcolor{httext}{Alternative approach: apply AM-GM to $x^4$, $4y^2$, $4z^4$.}}
The equality condition would require $x^4 = 4y^2 = 4z^4$ [\ldots]
\colorbox{htbg}{\textcolor{httext}{which is not satisfied. Therefore, this approach is incorrect.}}
\colorbox{htbg}{\textcolor{httext}{Alternative approach: fix $y$ and $z$ such that $yz = 2/x$,}}
\colorbox{htbg}{\textcolor{httext}{then minimize over $x$ [\ldots]}}

\textbf{Final answer: 16} \hfill \textit{(Groundtruth: \textbf{16})}

\vspace{4pt}
{\color{cardmuted}\rule{\linewidth}{0.3pt}}
\vspace{2pt}

{\small\sffamily
\colorbox{htbg}{\textcolor{httext}{\textbf{~Note: ~}}} The model cycles through four distinct AM-GM configurations, explicitly ruling each out before pivoting to the substitution that yields the minimum.}

\vspace{4pt}

\samplefooter{Qwen3-4B-Think }{MATH-500}{147}
\end{samplecard}
\definecolor{scbg}{HTML}{F6E8E3}      
\definecolor{sctext}{HTML}{9A5B47}    

\label{self_correction_sum_sample}
\begin{samplecard}{Self-Correction}{\checkmark\; Correct}
\small
\textsf{\textcolor{cardmuted}{\textbf{Question:}}}
Compute
\[
1 - 2 + 3 - 4 + 5 - \dots + 99 - 100.
\]

\vspace{6pt}
\textsf{\textcolor{cardmuted}{\textbf{Model trace:}}}
\vspace{2pt}

First, group the sequence as
\[
(1-2) + (3-4) + \dots + (99-100).
\]
Each pair sums to $-1$, and there are $50$ pairs, so the total is $-50$.

Alternatively, write
\[
1 + (-2+3) + (-4+5) + \dots + (-98+99) - 100.
\]
Each inner pair sums to $1$, and there are $49$ such pairs. Then
\[
49 - 100 = -51.
\]

\colorbox{scbg}{\textcolor{sctext}{Wait, now I'm confused. Which one is correct?}}\\
\colorbox{scbg}{\textcolor{sctext}{But earlier, when I tried grouping differently, I got $-51$. Maybe I made a mistake there.}}\\
\colorbox{scbg}{\textcolor{sctext}{Wait, let's recount that alternative approach.}}\\
\colorbox{scbg}{\textcolor{sctext}{There are $49$ inner pairs, so the total is $1 + 49 - 100 = -50$.}}\\
\colorbox{scbg}{\textcolor{sctext}{Earlier, when I thought I got $-51$, I must have miscounted the grouped terms.}}

\textbf{Final answer: $-50$} \hfill \textit{(Groundtruth: \textbf{$-50$})}

\vspace{4pt}
{\color{cardmuted}\rule{\linewidth}{0.3pt}}
\vspace{2pt}

{\small\sffamily
\colorbox{scbg}{\textcolor{sctext}{\textbf{~Note:~}}} After its two derivations disagree (\textbf{$-50$} vs. \textbf{$-51$}), the model recounts the terms in the alternate grouping, locates the miscount, and repairs the calculation; the correction changes the final answer.}

\vspace{4pt}

\samplefooter{Qwen3-4B-Think }{MATH-500}{N/A}
\end{samplecard}

\label{self_correction_functional_sample}
\begin{samplecard}{Self-Correction}{\checkmark\; Correct}
\label{card:last}
\small
\textsf{\textcolor{cardmuted}{\textbf{Question:}}}
The function $f$ satisfies
\[
f(x) + f(y) = f(x+y) - xy - 1
\]
for all real numbers $x$ and $y$. If $f(1)=1$, find all integers $n$ such that $f(n)=n$.

\vspace{6pt}
\textsf{\textcolor{cardmuted}{\textbf{Model trace:}}}
\vspace{2pt}

The model first proposes
\[
f(x)=g(x)+\frac{1}{2}x^2+x+c,
\]
then concludes that $c=\frac{1}{2}$ and therefore that there are no integer solutions.

\colorbox{scbg}{\textcolor{sctext}{But wait, let me check if I made any mistakes.}}\\
\colorbox{scbg}{\textcolor{sctext}{Hmm, maybe I should try a different approach.}}

It then assumes
\[
f(x)=ax^2+bx+c
\]
and compares coefficients in the functional equation:
\[
0 = 2a - 1 \Rightarrow a=\frac{1}{2}, \qquad 2c = c - 1 \Rightarrow c=-1.
\]

\colorbox{scbg}{\textcolor{sctext}{Wait, this is different from the previous solution. Where did I go wrong in the first approach?}}\\
\colorbox{scbg}{\textcolor{sctext}{In the first approach, I incorrectly set $c=\frac{1}{2}$. Actually, $c=-1$.}}

The corrected form is
\[
f(x)=\frac{1}{2}x^2+\frac{3}{2}x-1.
\]
Setting $f(n)=n$ gives
\[
\frac{1}{2}n^2+\frac{3}{2}n-1=n
\]
and hence
\[
(n+2)(n-1)=0.
\]

\textbf{Final answer: $-2, 1$} \hfill \textit{(Groundtruth: \textbf{$-2,1$})}

\vspace{4pt}
{\color{cardmuted}\rule{\linewidth}{0.3pt}}
\vspace{2pt}

{\small\sffamily
\colorbox{scbg}{\textcolor{sctext}{\textbf{~Note:~}}} The model reaches a wrong solution set, rechecks the algebra, identifies the mistaken constant term ($c = \tfrac{1}{2}$ vs.\ $c = -1$), and updates the final answer.}

\vspace{4pt}

\samplefooter{Qwen3-4B-Think }{MATH-500}{N/A}
\end{samplecard}

\end{document}